\documentclass{article}

\usepackage{microtype}
\usepackage{graphicx}
\usepackage{booktabs} 

\usepackage{hyperref}

\usepackage[accepted]{icml2020}

\usepackage{tikz}

\usepackage[font=footnotesize]{caption}

\usepackage{amsmath, amsthm, amssymb, amsfonts}

\usepackage{multirow}
\usepackage{multicol}

\usepackage{url}
\PassOptionsToPackage{hyphens}{url} 

\usepackage{csquotes}

\usepackage{pifont}
\usepackage{textcomp}
\usepackage{wasysym}

\usepackage[normalem]{ulem}

\usepackage[acronym,nohypertypes={acronym,notation}]{glossaries}

\newcommand{\eg}{e.g.}
\newcommand{\ie}{i.e.}
\newcommand{\etal}{et.\,al.}

\newcommand{\toolstyling}[1]{#1}
\newcommand{\styleG}{\toolstyling{StyleGAN}}
\newcommand{\bigG}{\toolstyling{BigGAN}}
\newcommand{\proG}{\toolstyling{ProGAN}}
\newcommand{\sndG}{\toolstyling{SN-DCGAN}}
\newcommand{\dcG}{\toolstyling{DCGAN}}
\newcommand{\cycleG}{\toolstyling{CycleGAN}}
\newcommand{\cramerG}{\toolstyling{CramerGAN}}
\newcommand{\mmdG}{\toolstyling{MMDGAN}}

\newcommand{\genm}{G} 
 
\newcommand{\dctm}{\mathcal{D}} 
\newcommand{\dctcom}[1]{\dctm{}(#1)}

\newcommand{\E}[1]{\mathbb{E}[#1]}

\usepackage{enumitem}

\usepackage{subcaption}
\icmltitlerunning{Leveraging Frequency Analysis for Deep Fake Image Recognition}

\begin{document}

\twocolumn[
\icmltitle{Leveraging Frequency Analysis for Deep Fake Image Recognition}

\icmlsetsymbol{equal}{*}

\begin{icmlauthorlist}
\icmlauthor{Joel Frank}{hgi}
\icmlauthor{Thorsten Eisenhofer}{hgi}
\icmlauthor{Lea Sch\"onherr}{hgi}
\icmlauthor{Asja Fischer}{hgi}
\icmlauthor{Dorothea Kolossa}{hgi}
\icmlauthor{Thorsten Holz}{hgi}
\end{icmlauthorlist}

\icmlaffiliation{hgi}{Ruhr-University Bochum, Horst G\"ortz Institute for IT-Security, Bochum, Germany}

\icmlcorrespondingauthor{Joel Frank}{joel.frank@rub.de}

\icmlkeywords{Machine Learning, ICML, GAN, Generative Adversarial Network, Discrete Cosine Transformation, DCT, Frequency Domain, Neural Network Architecture}

\vskip 0.3in
]

\printAffiliationsAndNotice{}  

\begin{abstract}
Deep neural networks can generate images that are astonishingly realistic, so much so that it is often hard for humans to distinguish them from actual photos.
These achievements have been largely made possible by Generative Adversarial Networks (GANs).
While \emph{deep fake} images have been thoroughly investigated in the image domain---a classical approach from the area of image forensics---an analysis in the \emph{frequency domain} has been missing so far.
In this paper, we address this shortcoming and our results reveal that in frequency space, GAN-generated images exhibit severe artifacts that can be easily identified.
We perform a comprehensive analysis, showing that these artifacts are consistent across different neural network architectures, data sets, and resolutions.
In a further investigation, we demonstrate that these artifacts are caused by upsampling operations found in all current GAN architectures, indicating a structural and fundamental problem in the way images are generated via GANs.
Based on this analysis, we demonstrate how the frequency representation can be used to identify deep fake images in an automated way, surpassing state-of-the-art methods.
\end{abstract} \section{Introduction}
\label{chap:intro}

GANs produce sample outputs---images, audio signals, or complete video sequences---that are astonishingly effective at fooling humans into believing their veracity~\cite{fried2019text, karras2019style, kumar2019melgan, song2020everybody}.
The difficulty of distinguishing these 
so-called \emph{deep fakes} from real media is for example demonstrated at \textit{whichfaceisreal.com}~\cite{which2019website}, a website on which a user can see two different images: one from the Flicker-Faces-HQ data set and one generated by \styleG{}~\cite{karras2019style}. The task is to decide which of these two images is real.
Even though humans generally do better than random guessing, players' performance reportedly peaked at around 75\% accuracy~\cite{simonite2019whichface}.

At a time where fake news have become a practical problem and Internet information campaigns might have influenced democratic processes~\cite{thompson2017memewarfare}, developing automated detection methods is a crucial task.
A worrying reminder is the example of Gabon's president Ali Bongo:
In late 2018, the president fell ill, not appearing in public for months.
As the public grew weary, the government released a video of the president, only to be immediately labeled as a deep fake. 
Albeit never to be confirmed as such, one week later the military launched an unsuccessful coup, citing the video as part of the motivation~\cite{hao2019deepfake}.

Previous research on detecting GAN-generated images has either utilized large, complex convolutional networks directly trained in the image domain~\cite{mo2018fake, yu2019attributing, tariq2019gan} or used hand-crafted features from the frequency domain~\cite{marra2019gans, valle2018tequilagan}.
In contrast, we provide in this paper a comprehensive analysis of the frequency spectrum across multiple different GAN architectures and data sets.
The surprising finding is that \emph{all} GAN architectures exhibit severe artifacts in the frequency domain.
Speculating that these artifacts stem from upsampling operations, we experiment with different upsampling techniques and identify patterns which are consistent with our earlier observations.

Based on these insights, we demonstrate that the frequency domain can be utilized for (i) efficiently separating real from fake images, as well as, (ii) identifying by which specific GAN a sample was generated.
In the first case, we demonstrate that the artifacts are so severe that a linear separation of the data is possible in the frequency space.
In the second case, we demonstrate that we can achieve higher accuracy, while simultaneously utilizing significantly less complex models, than state-of-the-art approaches~\cite{yu2019attributing} (using roughly 1.9\% of their parameters).
Additionally, we demonstrate that classifiers trained in the frequency domain are more robust against common image perturbations (\eg, blurring or cropping).
The code to reproduce our experiments and plots as well as all pre-trained models are available online at \href{https://github.com/RUB-SysSec/GANDCTAnalysis}{github.com/RUB-SysSec/GANDCTAnalysis}. 

In summary, our key contributions are as follows:
\begin{itemize}[topsep=0pt, itemsep=0pt, partopsep=4pt, parsep=4pt]
    \item We perform a comprehensive frequency-domain analysis of images generated by various popular GANs, revealing severe artifacts common across different neural network architectures, data sets, and resolutions.
    \item We show in several experiments that these artifacts arise from upsampling operations employed in all current GAN architectures, hinting towards a structural problem on how generative neural networks that map from a low-dimensional latent space to a higher-dimensional input space are constructed.
    \item We demonstrate the effectiveness of employing frequency representations for detecting GAN-generated deep fake images by an empirical comparison against state-of-the-art approaches. More specifically, we show that frequency-representation-based classifiers yield higher accuracy, while simultaneously needing significantly fewer parameters. 
    Additionally, these classifiers are more robust to common image perturbations.
\end{itemize} \section{Related Work}
\label{chap:related}

In the following, we present an overview of related work and discuss the connections to our approach.

\paragraph{Generative Adversarial Networks} 
GANs~\cite{goodfellow2014generative} have essentially established a new sub-field of modern machine learning research.
As generative models, they aim at estimating the probability density distribution underlying the training data.
Instead of employing standard approaches like likelihood maximization for training, they are based on the idea of defining a game between two competing models (usually neural networks): a generator and a classifier (also called discriminator).
The generator is tasked with producing samples that look like training data, while the discriminator attempts to distinguish real from fake (i.\,e., generated) samples.
These tasks can be translated into a min-max problem: a joint objective which is minimized w.r.t.~the parameters of the generator and maximized w.r.t.~the parameters of the discriminator. 

\paragraph{Image Synthesis}

While there exists a huge variety of models for image generation~\citep[e.g. see][]{van2017neural, razavi2019generating}, we will focus on images generated by GANs.
The earliest breakthrough in generating images with GANs was the switch to Convolutional Neural Network (CNN)~\cite{radford2015unsupervised}. While this might seem trivial today, it allowed GANs to outperform similar image synthesis methods at the time.
In follow-up work, GAN research yielded a steady stream of innovations, which pushed the state-of-the-art further: training with labeled data~\cite{mirza2014conditional, salimans2016improved}, utilizing the Wasserstein distance~\cite{arjovsky2017wasserstein, gulrajani2017improved, petzka2017regularization}, spectral normalization~\cite{miyato2018spectral}, progressive growing~\cite{karras2018progressive} or style mixing~\cite{karras2019style}, and employing very large models~\cite{brock2018large}, just to name a few examples.

\paragraph{Image Forensics}

Traditional image forensics uses the natural statistics of images to detect tampered media~\cite{fridrich2009digital, lyu2013natural}.
A promising approach is steganalysis~\cite{lukavs2006digital, fridrich2009digital, bestagini2013local}, where high-frequency residuals are used to detect manipulations.
These traditional methods have recently been expanded by CNN-based methods~\cite{bayar2016deep, bappy2017exploiting, cozzolino2017recasting, zhou2018learning}, which learn a more complex feature representation, improving the state-of-the-art for tampered media detection.

Prior work has utilized these findings for identifying GAN-generated images:
\citet{marra2018detection} provide a comparison of different steganalysis and CNN-based methods, 
several approaches use CNNs in the image domain~\cite{mo2018fake, yu2019attributing, tariq2019gan}, others use statistics in the image domain~\cite{mccloskey2018detecting, nataraj2019detecting}.
Another group of systems employs handcrafted features from the frequency domain, namely, steganalysis-based features~\cite{marra2019gans} and spectral centroids~\cite{valle2018tequilagan}.
In contrast, our method explores the entire frequency spectrum and we link our detection capabilities to fundamental shortcomings in the construction of modern generative neural networks.

Concurrently and independently to our research, both~\citet{wang2019cnn} and~\citet{durall2020watch} made similar observations:
\citeauthor{wang2019cnn} demonstrate that a deep fake classifier trained with careful data augmentation on the images of only one specific CNN generator is able to generalize to unseen architectures, data sets, and training methods.
They suggest that their findings hint at systematic flaws in today's CNN-generated images, preventing them from achieving realistic image synthesis.
\citeauthor{durall2020watch} show that current CNN-based generative models fail to reproduce spectral distributions.
They utilize the discrete Fourier Transform to analyze generated images and propose a spectral regularization term to tackle these issues. 
\begin{figure*}[t!]
    \centering
    \begin{minipage}{0.23\textwidth}
        \includegraphics[width=.99\textwidth]{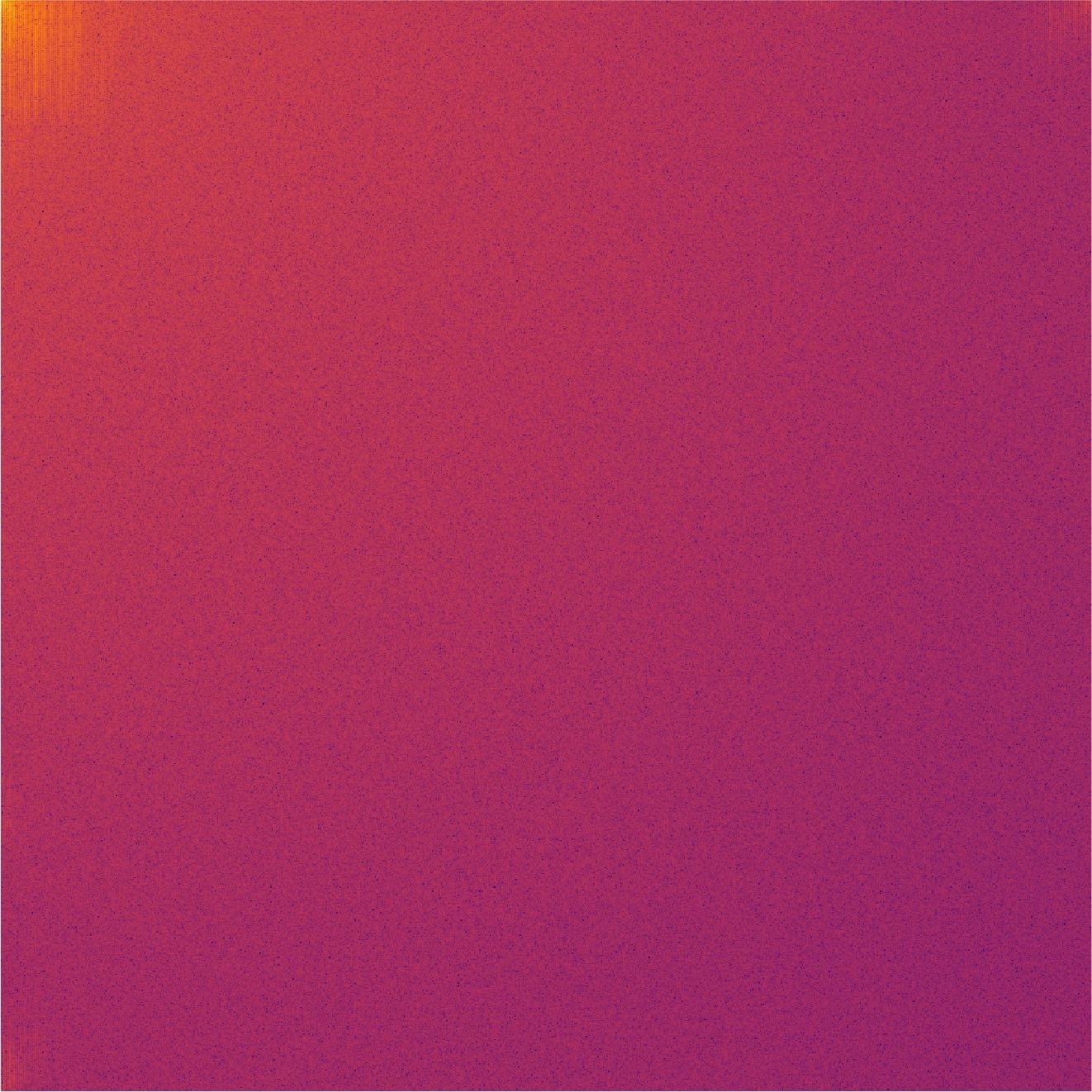}
        \caption*{FFHQ Spectrum}
    \end{minipage}
    \begin{minipage}{0.23\textwidth}
        \includegraphics[width=.99\textwidth]{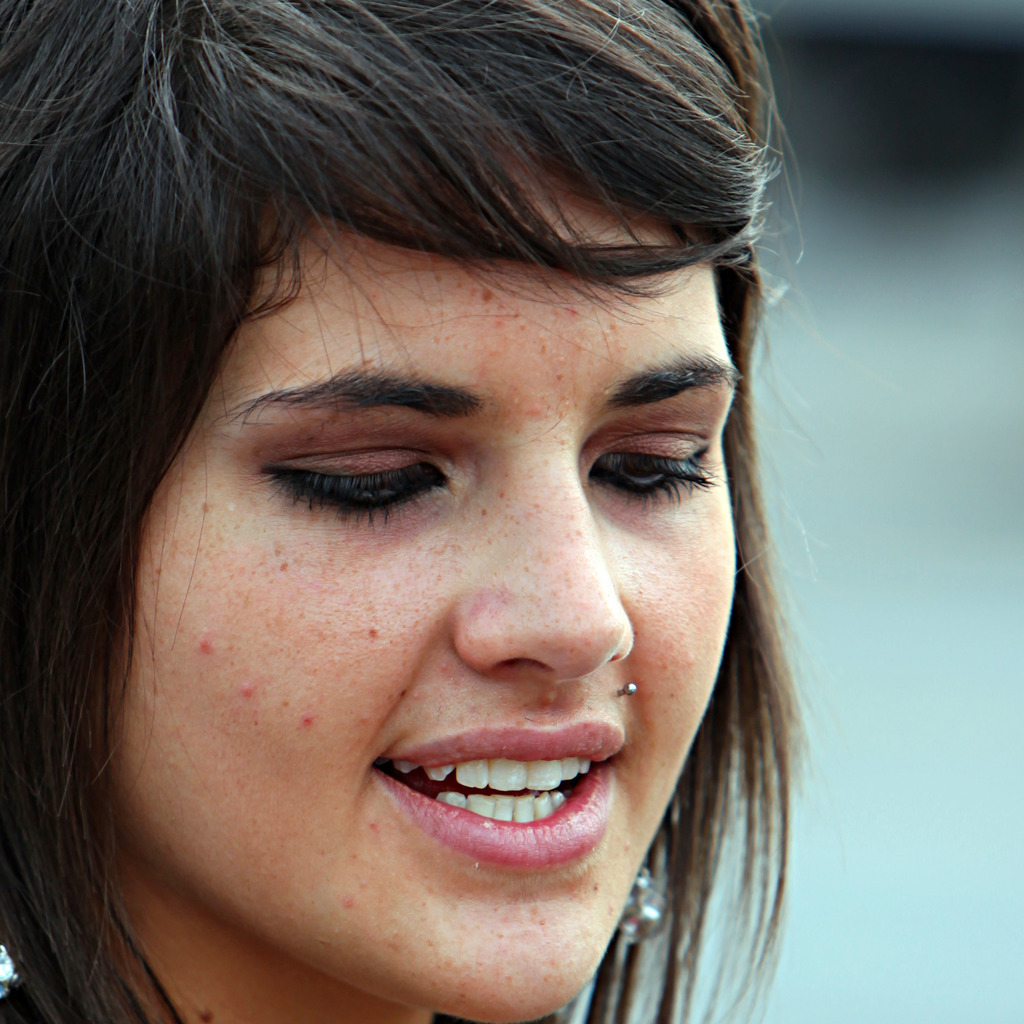}
        \caption*{FFHQ}
    \end{minipage}
    \begin{minipage}{0.23\textwidth}
        \includegraphics[width=.99\textwidth]{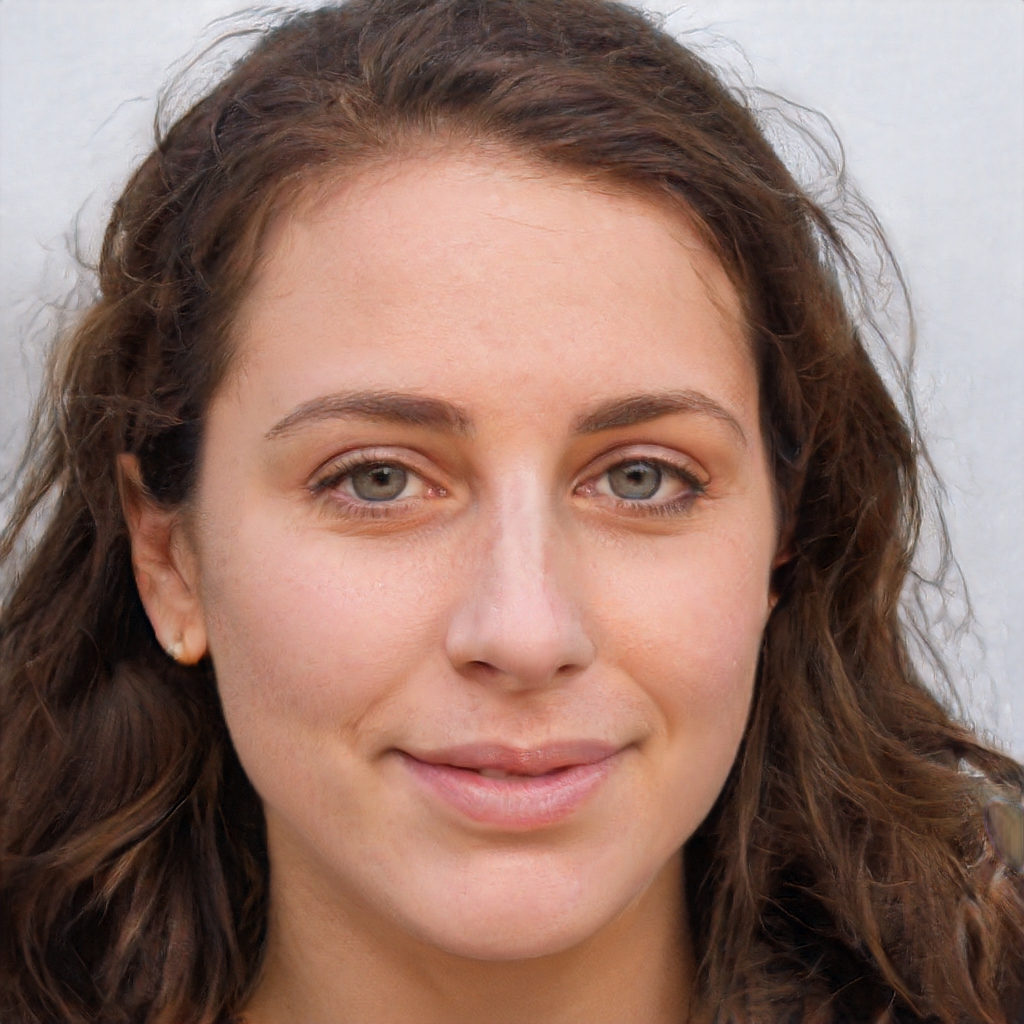}
        \caption*{\styleG{}}
    \end{minipage}
    \begin{minipage}{0.23\textwidth}
        \includegraphics[width=.99\textwidth]{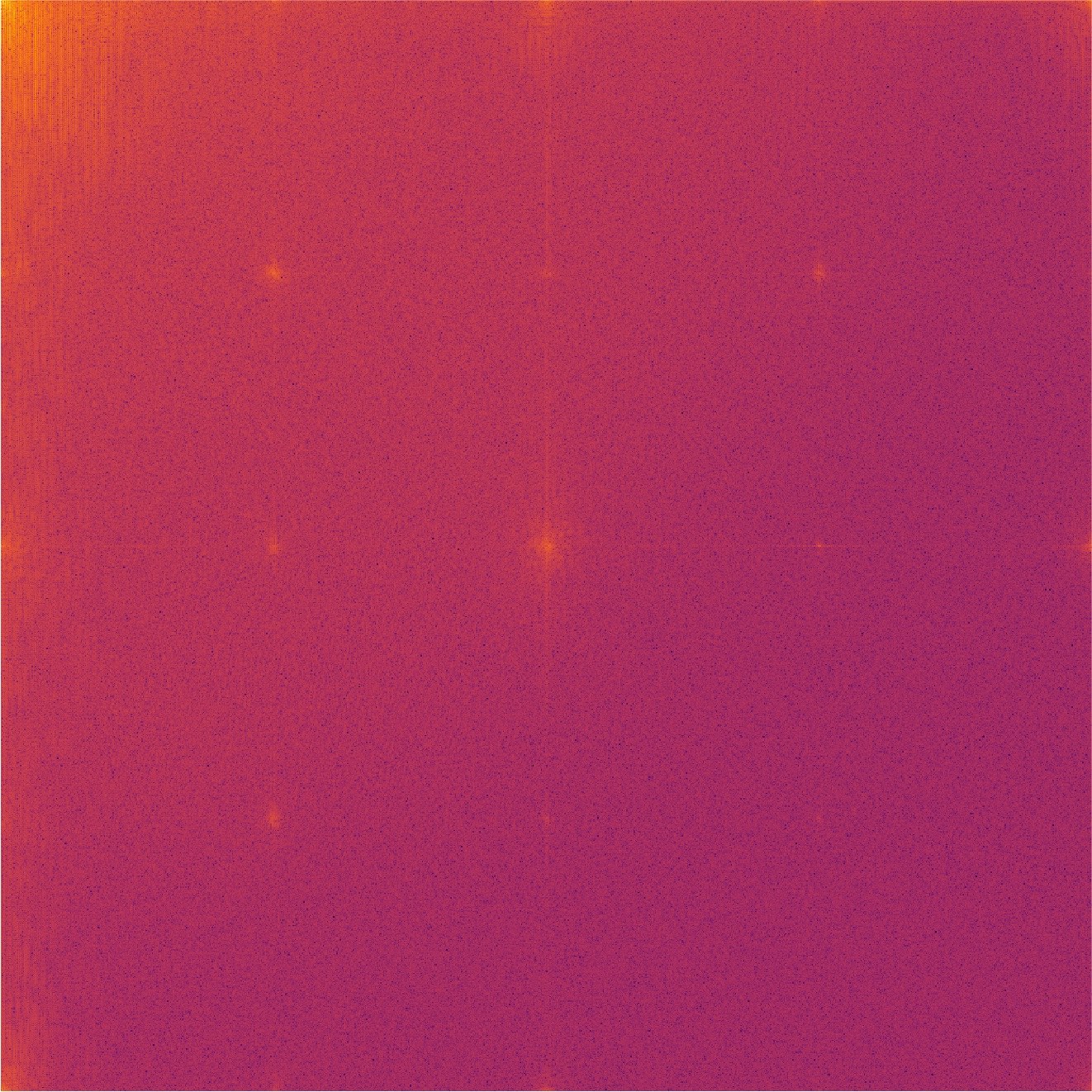}
        \caption*{\styleG{} Spectrum}
    \end{minipage}
    \begin{minipage}{0.040\textwidth}
        \includegraphics[width=.99\textwidth]{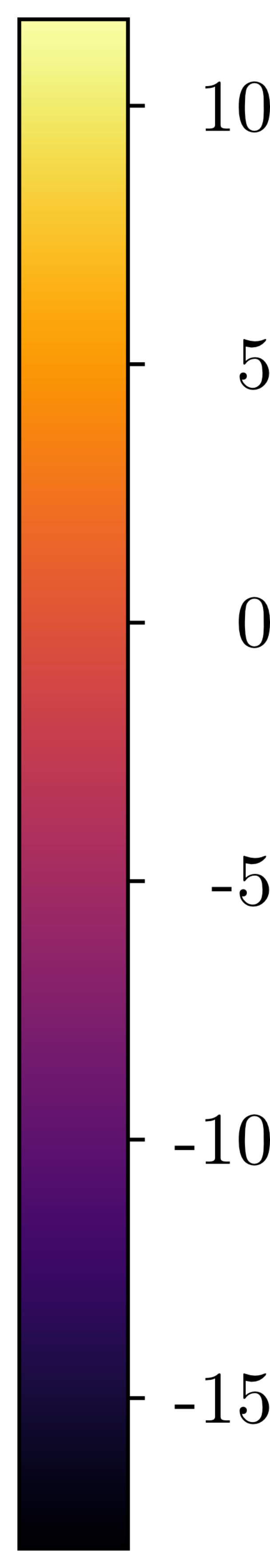}
        \caption*{}
    \end{minipage}
    \caption{
        \textbf{A side-by-side comparison of real and generated faces in image and frequency domain.}
        The left side shows an example and the mean DCT spectrum of the FFHQ data set.
        The right side shows an example and the mean DCT
        spectrum of a data set sampled from \styleG{} trained on FFHQ.
        We plot the mean by averaging over 10,000 images.
    }
    \label{fig:stats:flickr}
\end{figure*}

\section{Frequency Artifacts}
\label{chap:stats}

While early GAN-generated images were easily distinguishable from real images, newer generations fool even human observers~\cite{simonite2019whichface}.
To facilitate the development of automated methods for recognizing fake images, we take inspiration from traditional image forensics~\cite{lyu2013natural} and examine GAN-generated images in the frequency domain.

\subsection{Preliminaries}
\label{chap:stats:prelim}

We transform images into the frequency domain using the discrete cosine transform (DCT).
The DCT expresses, much like the discrete Fourier transform (DFT), a finite sequence of data points as a sum of cosine functions oscillating at different frequencies.
The DCT is commonly used in image processing due to its excellent energy compaction properties and its separability, which allows for efficient implementations. Together with a circular convolution-multiplication relationship~\cite{Chen1977}, it enables fast filtering.
We use the type-II 2D-DCT, which is, for example, also used in JPEG compression~\cite{fridrich2009digital}.

More formally, let an input image be given by the matrix\footnote{For simplicity, we omit the color channels and treat images as matrices, not as tensors.} $I \in  \mathbb{R}^{N_1 \times N_2}$, where the entries (specifying the pixel values) are denoted by $I_{x,y}$, and its DCT-transformed representation by the matrix $D \in \mathbb{R}^{N_1 \times N_2}$. 
The 2D-DCT is given by a function $\dctm{} : \mathbb{R}^{N_1 \times N_2} \to \mathbb{R}^{N_1 \times N_2}$ that maps an image $I=\{I_{x,y}\}$ to its frequency representation $D=\{D_{k_x, k_y}\}$, with
\begin{equation*} 
   \resizebox{\linewidth}{!}{
$D_{k_x, k_y} = w(k_x) w(k_y) \displaystyle\sum_{x=0}^{N_1 - 1} \displaystyle\sum_{y=0}^{N_2 - 1} I_{x, y}  \; \text{cos} \bigg[ \frac{\pi}{N_1} \bigg( x + \frac{1}{2} \bigg)  k_x \bigg] \; \text{cos} \bigg[ \frac{\pi}{N_2} \bigg(y + \frac{1}{2} \bigg) k_y \bigg], $
   }
\end{equation*}
for $\forall k_x = 0, 1, 2, \dots, N_1 - 1$ and $ \forall k_y = 0, 1, 2, \dots, N_2 - 1 $, and where  $w(0) = \sqrt{\frac{1}{4N}}$ and $w(k) = \sqrt{\frac{1}{2N}}$ for $k>0$.

When we plot the DCT spectrum, we depict the DCT coefficients as a heatmap.
Intuitively, the magnitude of each coefficient is a measure of how much the corresponding spatial frequency contributed to the overall image.
The horizontal direction corresponds to frequencies in the $x$ direction, while the vertical direction corresponds to frequencies in the $y$ direction.
In practice, we compute the 2D-DCT as a product of two 1D-DCTs, \ie, for images we first compute a DCT along the columns and then a DCT along the rows.
This results in the top left corner of the heatmap corresponding to low frequencies ($k_x$ and $k_y$ close to zero), while the right bottom corner corresponds to high frequencies ($k_x$ and $k_y$ close to $N_1 - 1$ and $N_2 - 1$, respectively). 
Due to the energy compaction property of the DCT, the coefficients drop very quickly in magnitude when moving to high frequencies, thus, we log-scale the coefficients before plotting.
All plots are computed for gray-scale images, produced using standard gray-scale transformations (\ie, a weighted average over the color channels).
We also computed statistics for each color channel separately, which are consistent with the findings for gray-scale images. These can be found in the supplementary material, where we also provide plots of the absolute difference between the spectra of real and fake images.

\subsection{Investigating Generated Images in the Frequency Domain}
\label{chap:stats:experiments}

\begin{figure*}[t!]
    \centering
    \begin{minipage}{0.18\textwidth}
        \includegraphics[width=.99\textwidth]{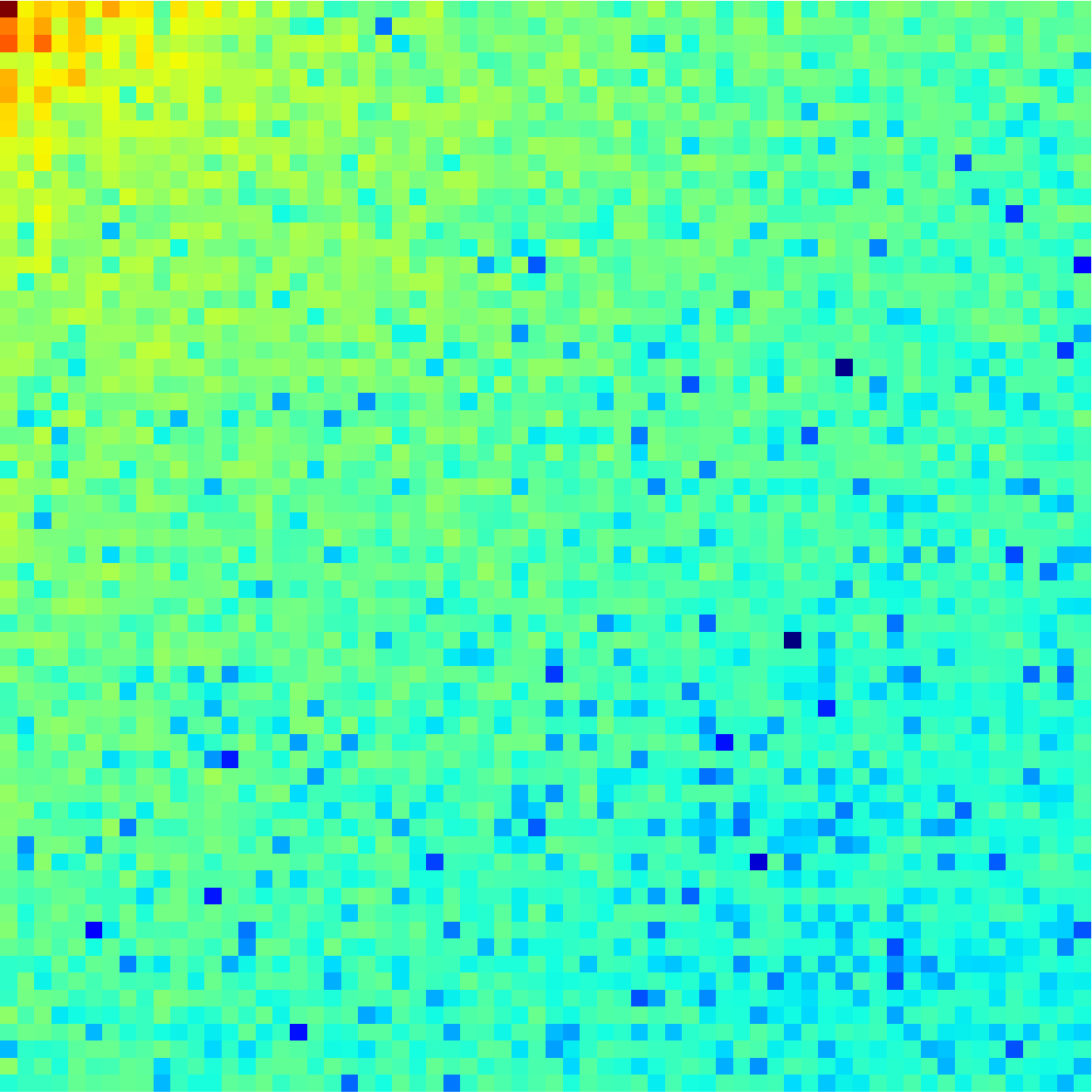}
        \caption*{Stanford dogs}
    \end{minipage}
    \begin{minipage}{0.18\textwidth}
        \includegraphics[width=.99\textwidth]{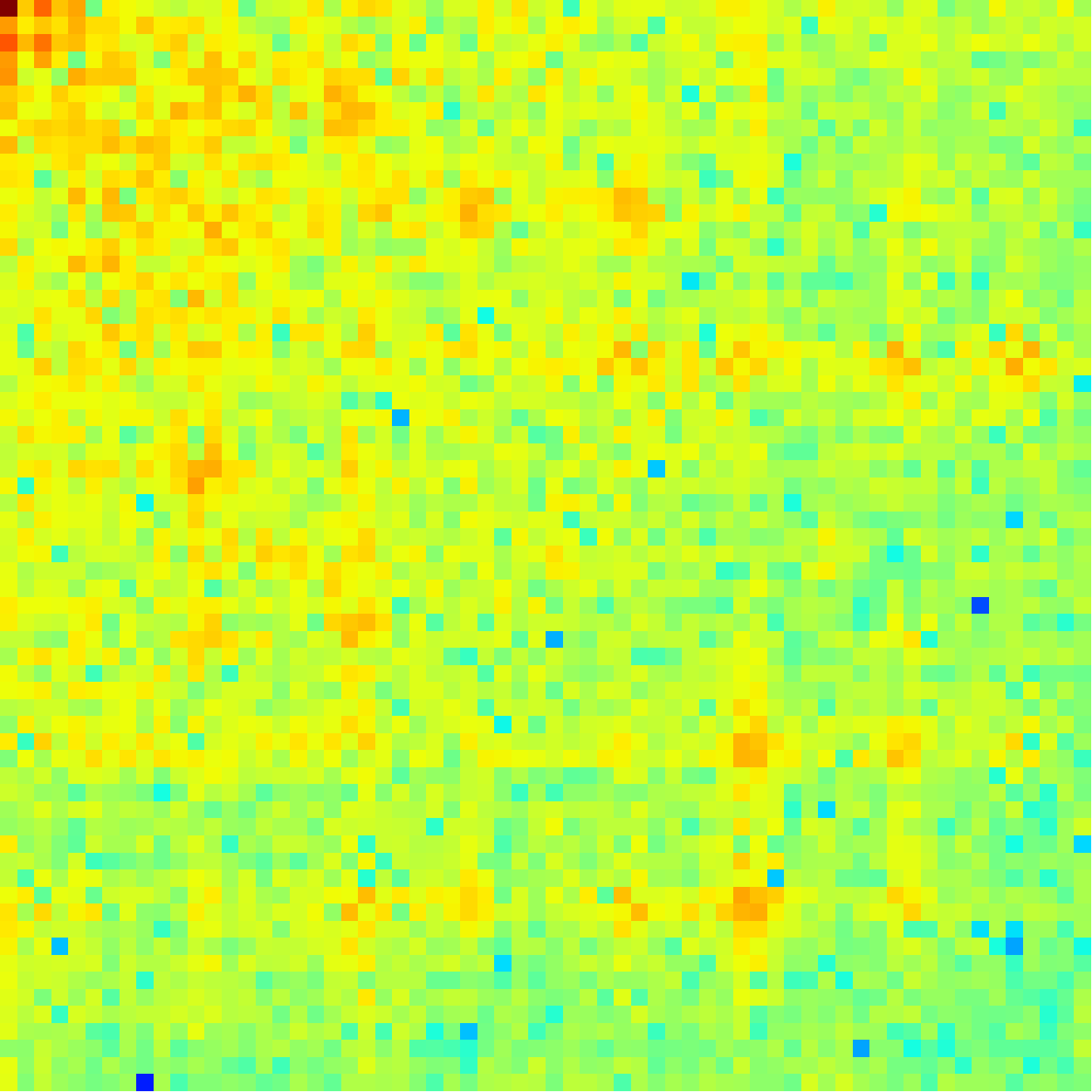}
        \caption*{\bigG{}}
    \end{minipage}
    \begin{minipage}{0.18\textwidth}
        \includegraphics[width=.99\textwidth]{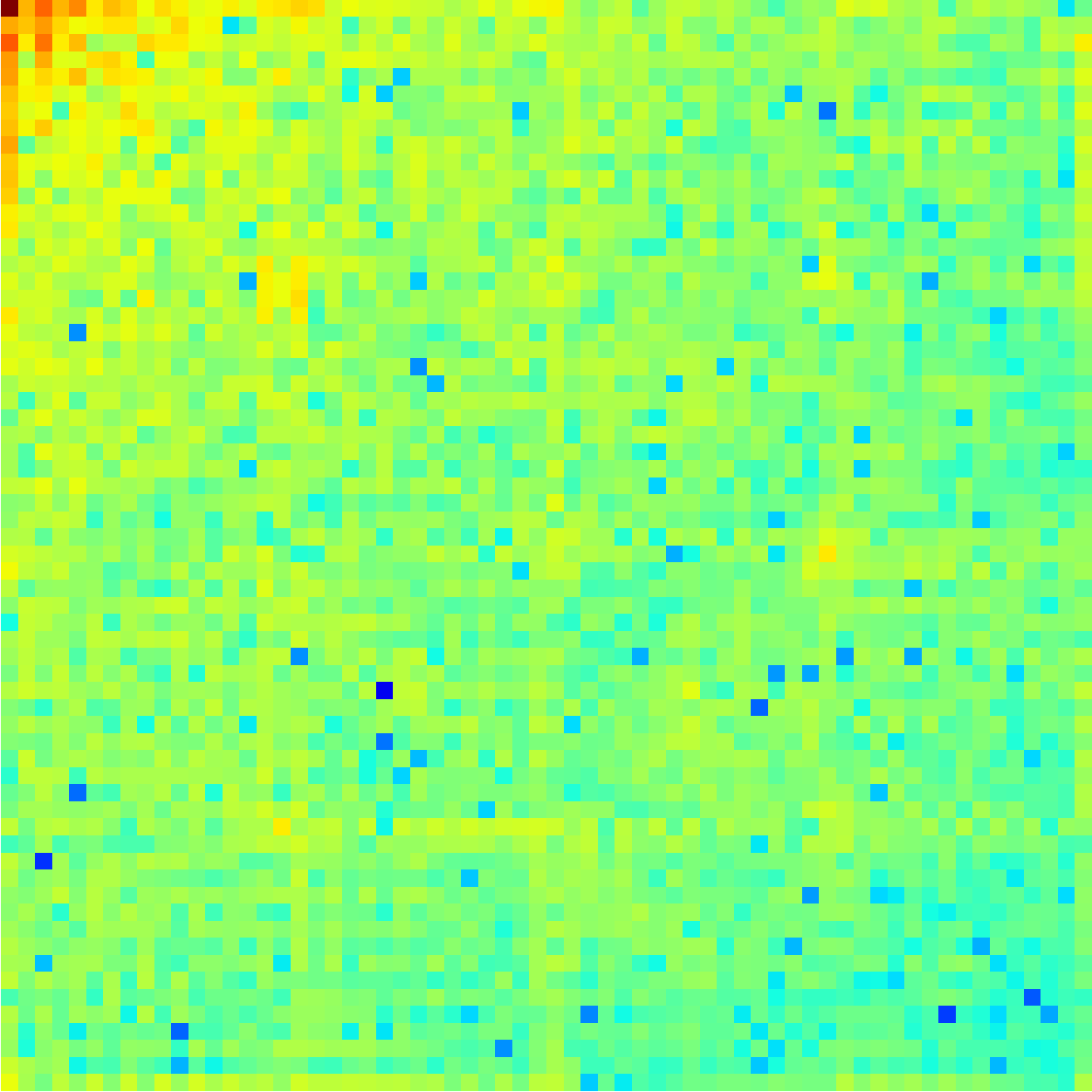}
        \caption*{\proG{}}
    \end{minipage}
    \begin{minipage}{0.18\textwidth}
        \includegraphics[width=.99\textwidth]{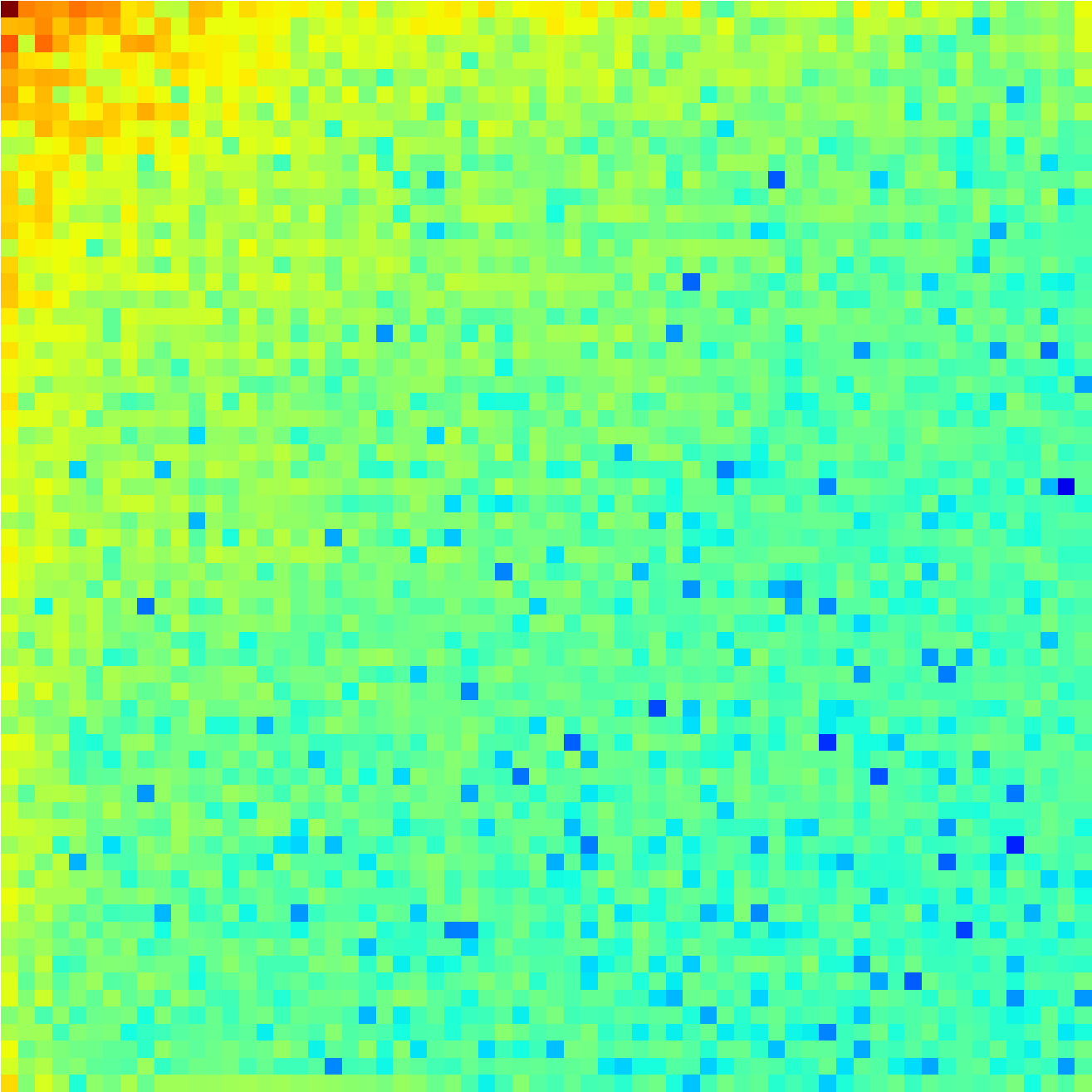}
        \caption*{\styleG{}}
    \end{minipage}
    \begin{minipage}{0.18\textwidth}
        \includegraphics[width=.99\textwidth]{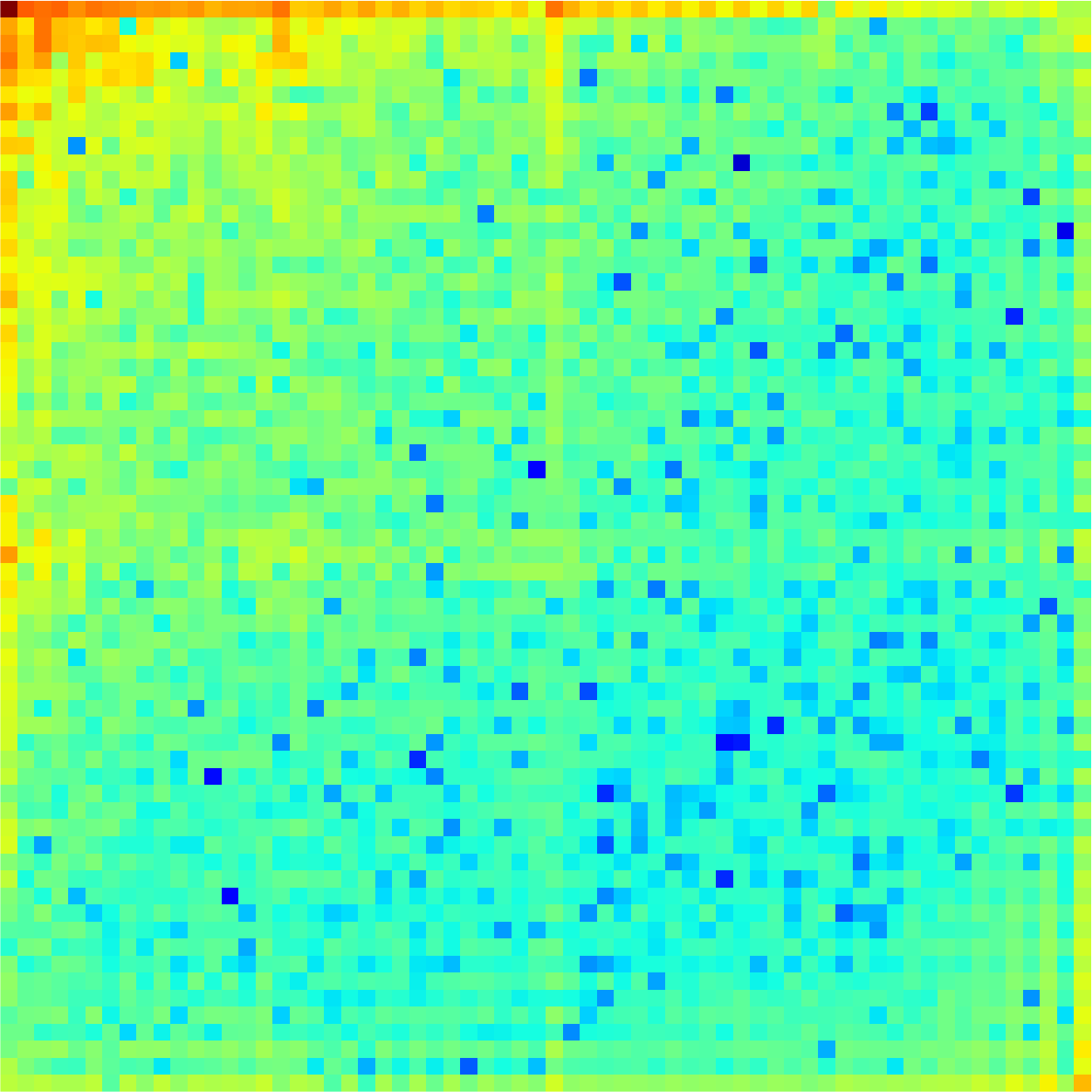}
        \caption*{\sndG{}}
    \end{minipage}
    \begin{minipage}{0.04\textwidth}
        \includegraphics[width=.99\textwidth]{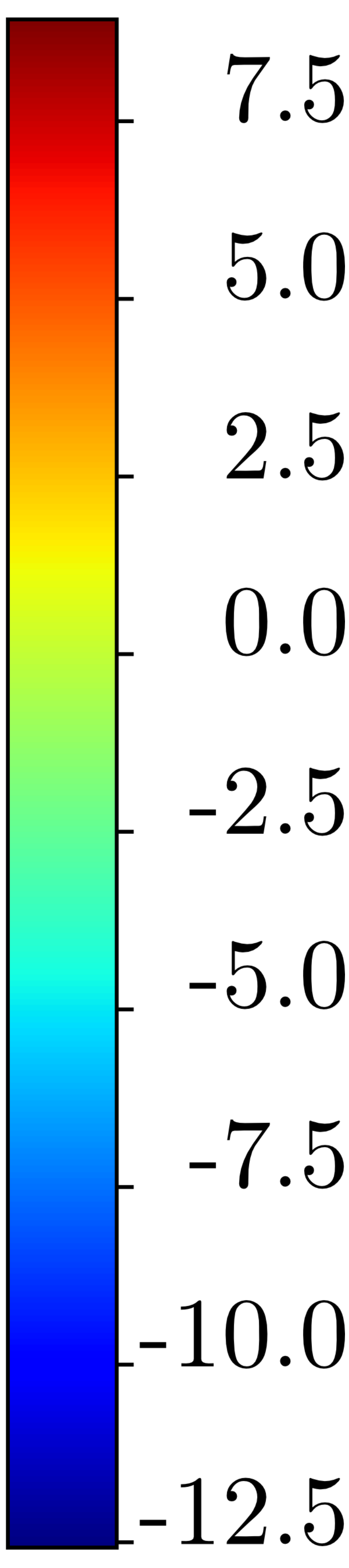}
        \caption*{}
    \end{minipage}
    \caption{
        \textbf{The spectra of images generated by different neural networks trained on the Stanford dog data set.}
        The left-most heatmap depicts the mean spectrum of the Stanford dog data set. 
        The rest depicts the mean spectra of images generated by different GANs.
        We plot the mean of the DCT spectra by averaging over 10,000 images.
    }
    \label{fig:stats:dogs}
\end{figure*}

We start with examining our introductory example, \ie, images from the website \textit{whichfaceisreal.com}.
The images are either from the Flickr-Faces-HQ (FFHQ) data set 
or from a set generated by \styleG{}~\cite{karras2019style}. 
In Figure~\ref{fig:stats:flickr}, we visualize the frequency statistics of the data set by plotting the means of the corresponding spectra over the sets.
As a reference, we include a sample from each data set.
In the image domain, both samples look similar, however, in the frequency domain, one can easily spot multiple clearly visible artifacts for the generated images.

The spectrum of the FFHQ images represents a regular spectrum of DCT coefficients.
Multiple studies \citep[e.g. see][]{burton1987color, tolhurst1992amplitude, field1987relations, field1999wavelets} have observed that the average power spectra of natural images tend to follow a $\frac{1}{f^\alpha}$ curve, where $f$ is the frequency along a given axis and $\alpha \approx 2$~\citep[see Figure 2 (a) from][]{torralba2003statistics}.
The low frequencies (in the upper left corner of the heatmap) contribute the most to the image, and the contribution is gradually decreasing as we approach the higher frequencies (lower right corner).
Intuitively, if a group of neighboring pixels contains similar values, \ie, they form an isochromatic area in the image, one can approximate those with a sum of low frequency functions. 
However, if there is a sudden change in the values, \eg, corresponding to an edge in the images, one has to use higher frequency functions to achieve a good approximation.
Therefore, since most pixels in images are correlated to each other, \ie, colors mostly change gradually, large parts of the image can be approximated well by using low-frequency functions.

The images generate by \styleG{}, however, exhibit artifacts throughout the spectrum.
In comparison to the spectra of natural images, \styleG{}-generated images contain strong high frequencies components (visible as high values in the lower right corner), as well as generally higher magnitudes throughout the spectrum.
Especially notably is the grid-like pattern scattered throughout,
also clearly noticeable in the top right (highest frequencies in $x$ direction) and lower left corner (highest frequencies in $y$ direction).

To analyze if this pattern in the spectral domain is a common occurrence 
for different GAN types and implementations, or simply a fault specific to the \styleG{} instance we studied, we selected four different architectures, namely \bigG{}~\cite{brock2018large}, \proG{}~\cite{karras2018progressive}, \styleG{}~\cite{karras2019style}, and \sndG~\cite{miyato2018spectral}. Note that all of them were under the top-ten entries in the recent Kaggle competition on generating images of dogs~\cite{kaggle2019comp}.
In this competition, the participants were required to upload 10,000 samples from their respective GAN instance.
For each architecture, we downloaded and analyzed the corresponding samples and the training data~\cite{standford2011dog}.

We show the mean of the spectra of the samples of each GAN and the training data in Figure~\ref{fig:stats:dogs}.
As discussed, statistics of natural images have been found to follow particular regularities. 
Like natural images, the DCT coefficients of the GAN-generated images decrease rapidly towards high frequencies. 
However, the spectral images often show a grid-like pattern.
\styleG{} seems to better approximate spectra of natural images than the other GANs, but still contains high coefficients along the upper and left side of their spectra.
These findings indicate a structural problem of the way GANs generate images that we explore further. \begin{figure*}
    \centering
    \begin{minipage}{0.23\textwidth}
        \includegraphics[width=\textwidth]{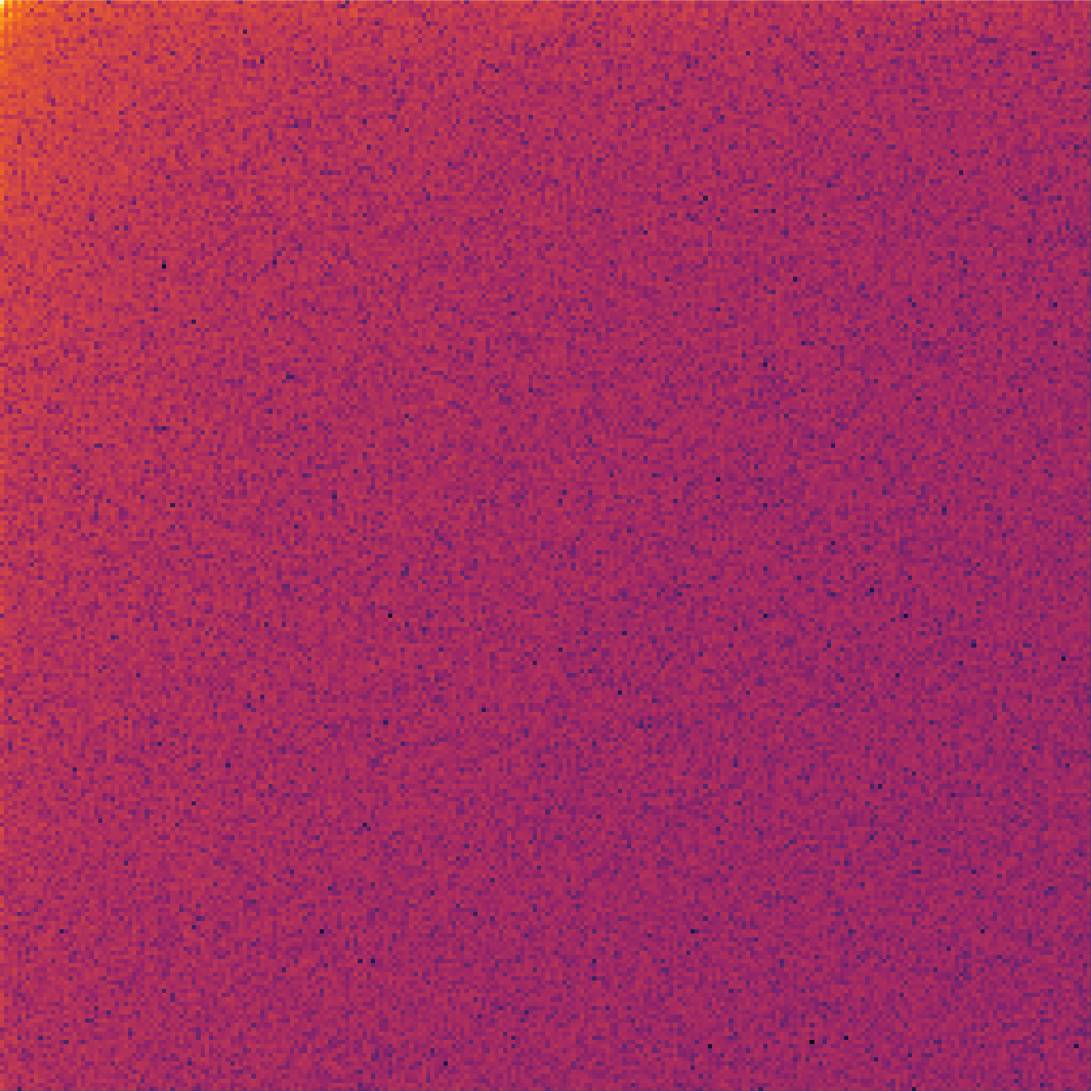}
        \caption*{LSUN bedrooms}
    \end{minipage}
    \begin{minipage}{0.23\textwidth}
        \includegraphics[width=\textwidth]{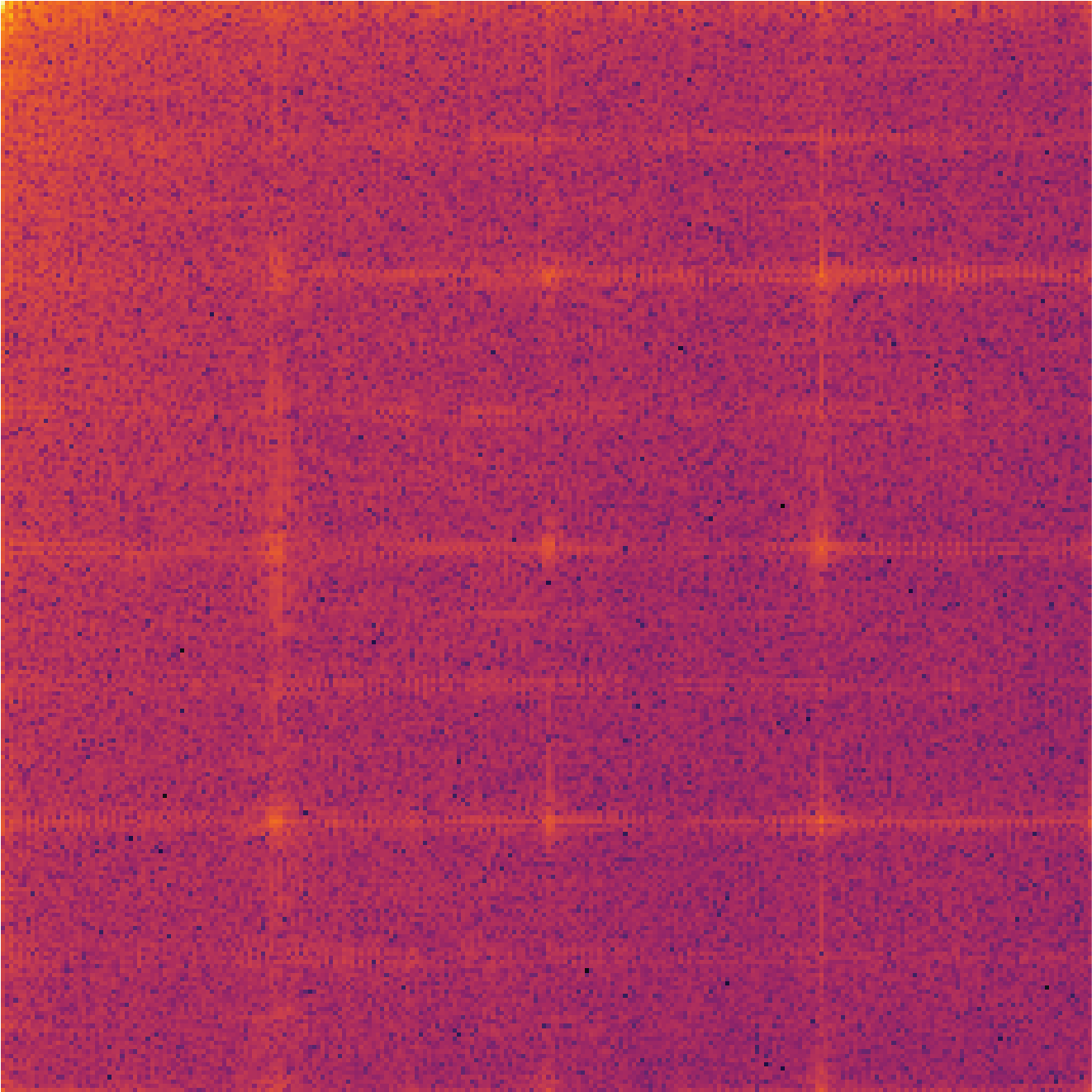}
        \caption*{Nearest Neighbor}
    \end{minipage}
    \begin{minipage}{0.23\textwidth}
        \includegraphics[width=\textwidth]{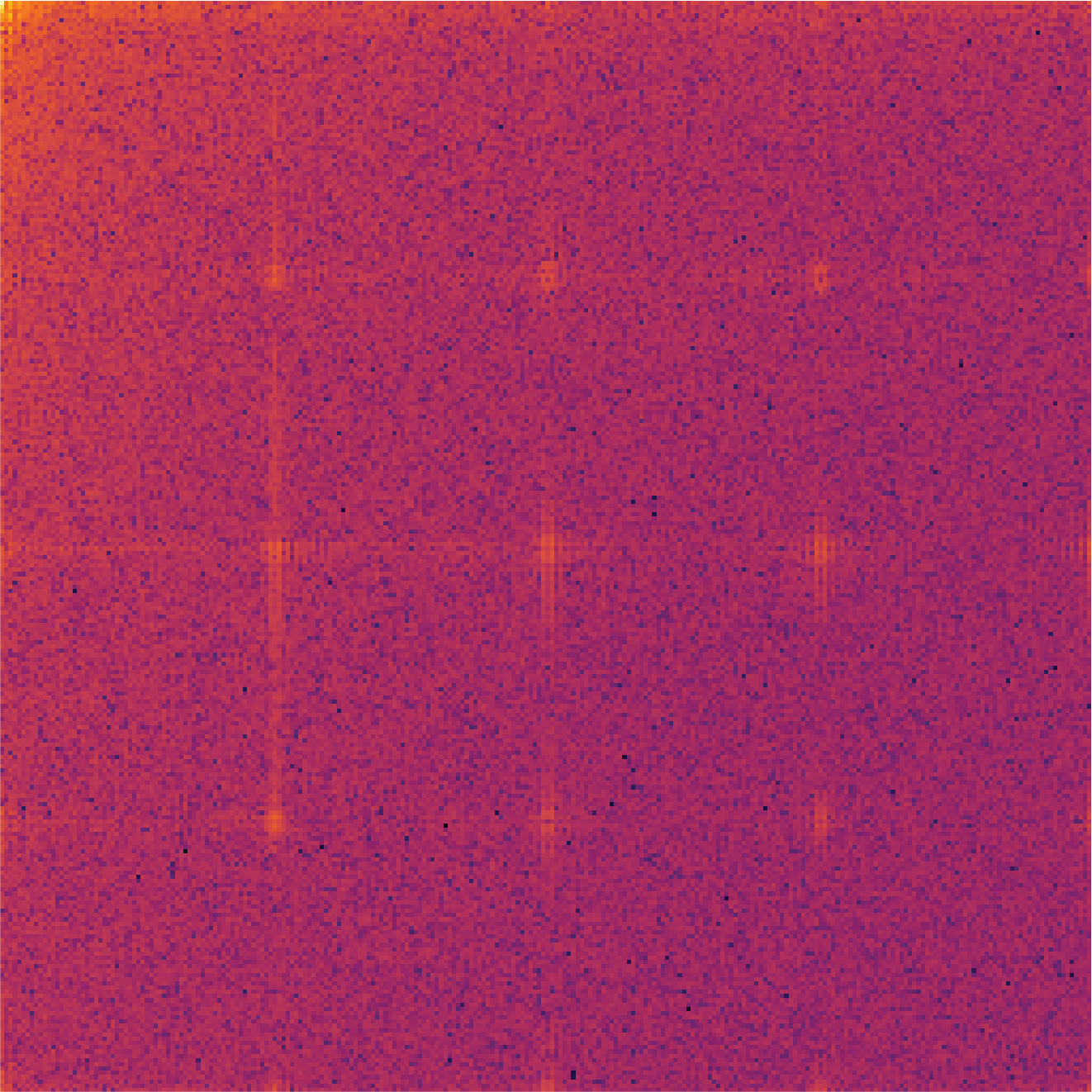}
        \caption*{Bilinear}
    \end{minipage}
    \begin{minipage}{0.23\textwidth}
        \includegraphics[width=\textwidth]{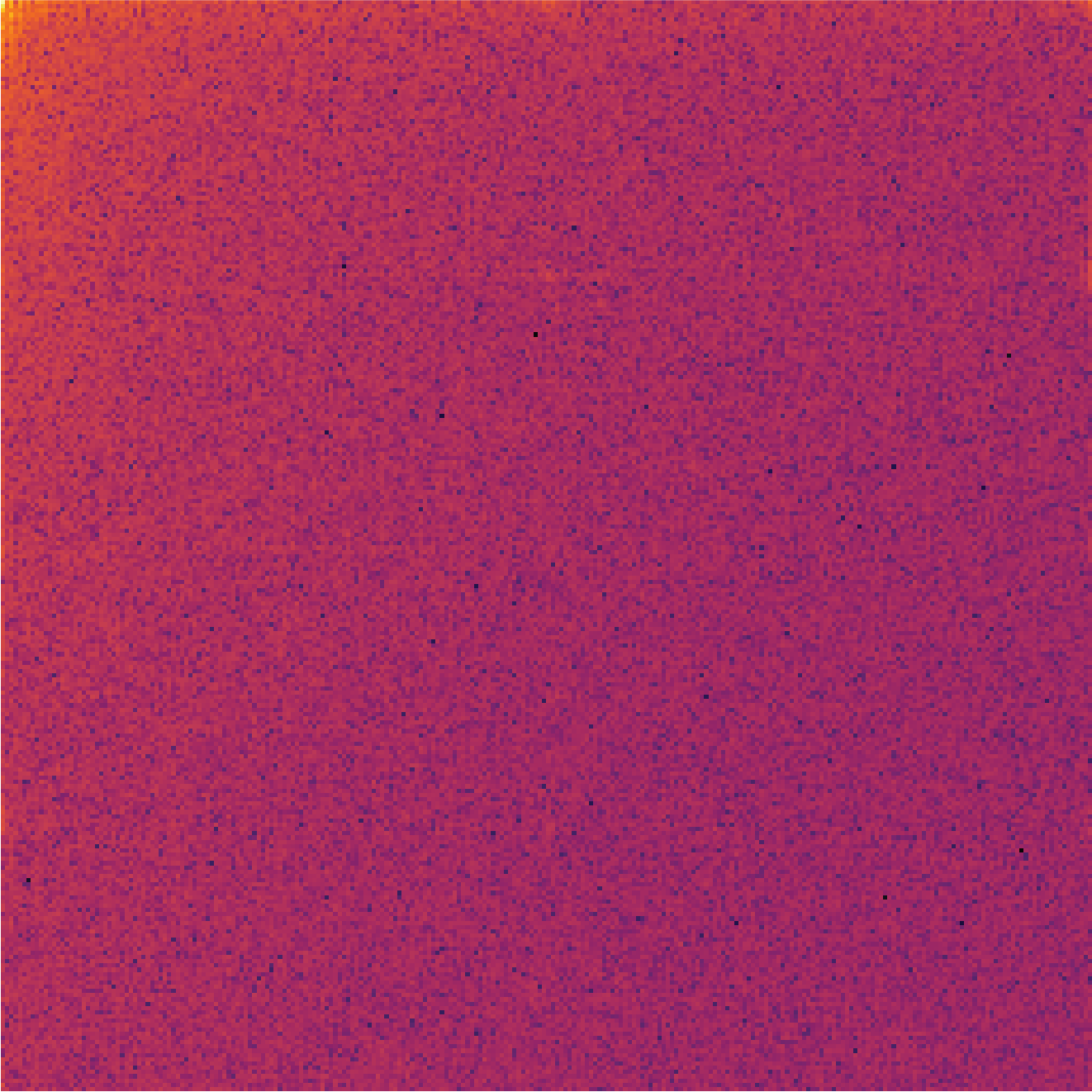}
        \caption*{Binomial}
    \end{minipage}
        \begin{minipage}{0.04\textwidth}
        \includegraphics[width=\textwidth]{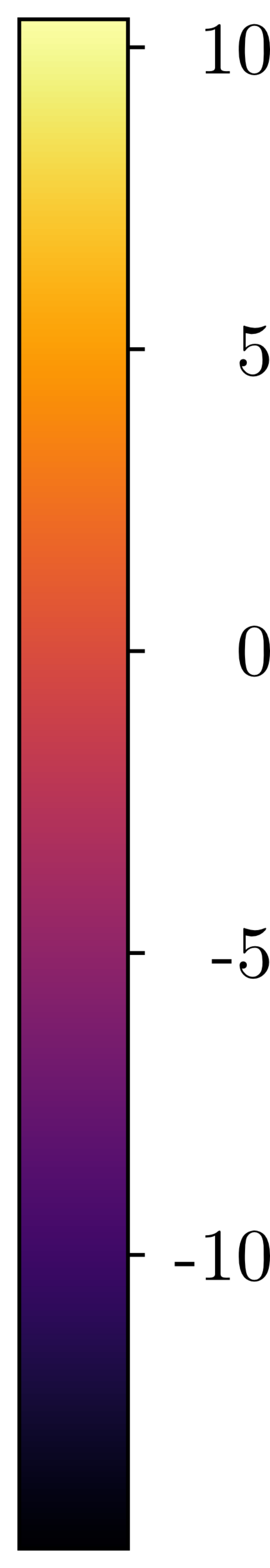}
        \caption*{}
    \end{minipage}
    \caption{
        \textbf{The frequency spectrum resulting from different upsampling techniques.}
        We plot the mean of the DCT spectrum.
        We estimate $\E{\dctcom{I}}$ by averaging over 10,000 images sampled from the corresponding network or the training data.
    }
    \label{fig:upsample:comparison}
\end{figure*}

\subsection{Upsampling}
\label{chap:upsampling}

We hypothesize that the artifacts found for GAN-generated images in the frequency domain  
stem from their employed 
upsampling operations. We make this more concrete in the following:
The generator of a GAN forms a mapping from a low-dimensional latent space to the higher-dimensional data space.
In practice, the dimensionality of the latent space is much lower than the dimensionality of the data space, \eg, the generator of the \styleG{}-instance which generated the images presented in Figure~\ref{fig:stats:flickr} defines a mapping $\genm: \mathbb{R}^{100} \to \mathbb{R}^{1024 \times 1024}$.
In typical GAN architectures, the latent vector gets successively upsampled until it reaches the final output dimension. 

Previous work has already linked upsampling operations to causing grid-like patterns in the image domain~\cite{odena2016deconvolution}.
Recognizing this, the architecture of both the generator-network and the discriminator-network shifted from using strided transposed convolution (e.g., employed in \dcG{}~\cite{radford2015unsupervised}, \cramerG{}~\cite{bellemare2017cramer}, \cycleG{}~\cite{zhu2017unpaired},  \mmdG{}~\cite{binkowski2018demystifying}, and \sndG{}~\cite{miyato2018spectral}) to using traditional upsampling methods---like nearest neighbor or bilinear upsampling~\cite{gonzalez2004digital}---followed by a convolutional layer (e.g., employed in \proG{}~\cite{karras2018progressive}, \bigG{}~\cite{brock2018large}, and \styleG{}~\cite{karras2019style}).
While these changes addressed the problem in the image domain, our results show that the artifacts are still detectable in the frequency domain.

Moreover, both upsampling and downsampling operations have recently been linked to compromising shift invariance in neural networks, \ie, they cause classifier predictions to vary dramatically due to a simple one-pixel shift in the input image~\cite{azulay2018deep}.
Recently, \citet{zhang2019shiftinvar} proposed to use low-pass filtering after convolution and pooling layers to mitigate some of these effects.

We investigate how different upsampling strategies affect the DCT spectrum.
Typically, we want to double the dimensionality of an image.
When doing so, we have to fill in the blanks.
However, this is known to cause artifacts.
Thus, a range of different techniques have been invented throughout the past years to minimize these detrimental effects~\cite{gonzalez2004digital}.
In our analysis, we investigate the effect of three different upsampling techniques:
\begin{itemize}[topsep=0pt, itemsep=0pt, partopsep=4pt, parsep=4pt]
    \item \textbf{\textit{Nearest Neighbor}}: The missing pixels in the upsampled image are approximated by copying the nearest~pixel~\citep[nearest-neighbor upsampling; for a visual representation see the work of][]{odena2016deconvolution}.
    \item \textbf{\textit{Bilinear}}: Similar to nearest neighbor, but \emph{after} copying the pixel values, the upsampled image is convolved with an anti-aliasing kernel (the filter \,$[1,2,1]$\,)\footnote{Note that for brevity, we list the 1D-variant of the anti-aliasing kernel; in practice, we generate the 2D-variant as the outer product: $m\,m^T$, where $m$ is the corresponding kernel.}. 
    This strategy is employed in the original \styleG{}.
    \item \textbf{\textit{Binomial}}: We follow \citet{zhang2019shiftinvar} and test the Binomial-5 kernel (i.\,e., the application of the filter \,$[1,4,6,4,1]$\,) as a replacement for the bilinear kernel. 
\end{itemize}
We trained three different versions of \styleG{} on the LSUN bedrooms~\cite{yu15lsun} dataset: in one, we kept the standard bilinear upsampling strategy, and in two, we replaced it by nearest-neighbor upsampling and binomial upsampling, respectively.
We train at a resolution of $256 \times 256$, using the standard model and settings provided in the \styleG{} repository~\cite{karras2019style}.

The results are shown in Figure~\ref{fig:upsample:comparison}.
As expected, with more elaborated upsampling techniques and with a larger size of the employed kernel, the spectral images become smoother and the artifacts less severe.
These findings are in line with our hypothesis that the artifacts are caused by upsampling operations.

\section{Frequency-Based Deep-Fake Recognition}
\label{chap:detection}

In the following, we describe our experiments to demonstrate the effectiveness of investigating the frequency domain for differentiating GAN-generated images.
In a first experiment, we show that DCT-transformed images are fully linearly separable, while classification on raw pixels requires non-linear models.
Further, we verify that our model utilizes the artifacts discovered in Section~\ref{chap:stats:experiments}.
Then, we recreate the experiments by~\citet{yu2019attributing} and show how we can utilize the frequency domain to match generated images to their underlying architecture, demonstrating that we can achieve higher accuracy while utilizing fewer parameters in our models.
Finally, we investigate how our classifier behaves when confronted with common image~perturbations.

All experiments in this chapter were performed on a server running Ubuntu 18.04, with 192 GB RAM, an Intel Xeon Gold 6230, and four Nvidia Quadro RTX 5000.

\subsection{Detecting Fake Images}
\label{chap:detection:detection}

First, we want to demonstrate that using frequency information allows to efficiently separate real from fake images. We consider our introductory example, i.\,e., aim at distinguishing real images from the FFHQ data set and fake images generated by \styleG{}.
As discussed in Section~\ref{chap:stats}, in the frequency domain, the images show severe artifacts.
These artifacts makes it easy to use a simple linear classifier.
To demonstrate this, we perform a ridge regression on real and generated images, after applying a DCT. For comparison, we also perform ridge regression on the original data representation. 

\begin{table}[t]
    \centering
    \caption{\textbf{Ridge regression performed on FFHQ data set.} We report the accuracy on the test set. We also report the gain in accuracy when training in the frequency domain instead of using raw pixels. Best score is highlighted in \textbf{bold}.}
    \resizebox{0.85\columnwidth}{!}{
    \begin{tabular}{lrr}
        \toprule
        Method & Accuracy  & Gain \\ \midrule
        Ridge-Regression-Pixel & 75.78\,\%  & \\ 
        Ridge-Regression-DCT & \textbf{100.00}\,\% & + 24.22\,\%  \\ \bottomrule
    \end{tabular}}
    \label{tab:detection:style}
\end{table}

\begin{table*}[ht]
    \centering
    \caption{\textbf{Ridge regression performed on data samples generated by \styleG{} for different upsampling techniques}. More elaborate upsampling techniques seem to remove artifacts in the image domain. }
    \resizebox{0.85\textwidth}{!}{
    \begin{tabular}{lrrrrrr}
        \toprule
        Method & Nearest Neighbor  & Gain & Bilinear & Gain & Binomial & Gain \\ \midrule
        Ridge-Regression-Pixel & 74.77\,\%  & & 62.13\,\%  & & 52.64\,\%  &\\ 
        Ridge-Regression-DCT & \textbf{98.24}\,\% & + 23.47\,\%  &\textbf{85.96}\,\% & + 23.83\,\% & \textbf{84.20}\,\% & + 31.56\,\% \\ \bottomrule
    \end{tabular}}
    \label{tab:detection:up}
\end{table*}

\begin{figure*}
    \centering
    \resizebox{0.95\textwidth}{!}{
    \begin{minipage}{0.31\textwidth}
        \includegraphics[width=\textwidth]{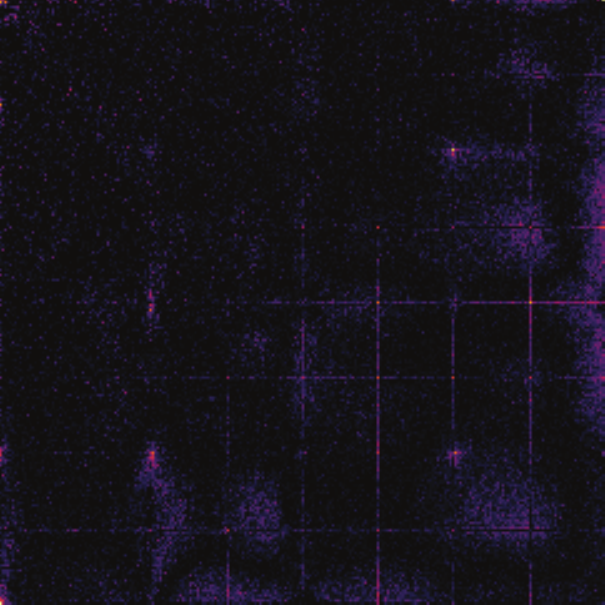}
        \caption*{Nearest Neighbor}
    \end{minipage}
    \begin{minipage}{0.31\textwidth}
        \includegraphics[width=\textwidth]{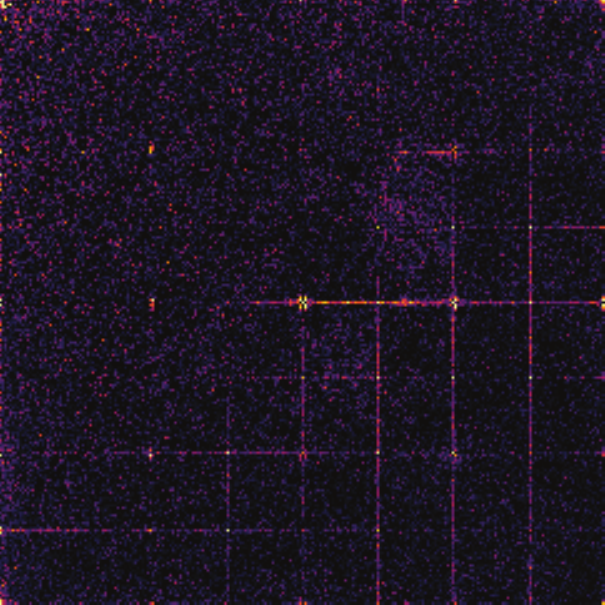}
        \caption*{Bilinear}
    \end{minipage}
    \begin{minipage}{0.31\textwidth}
        \includegraphics[width=\textwidth]{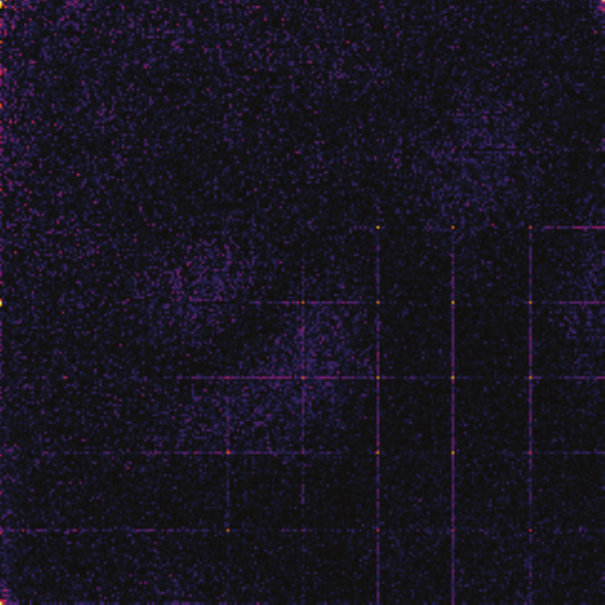}
        \caption*{Binomial}
    \end{minipage}
        \begin{minipage}{0.04\textwidth}
        \includegraphics[scale=0.55]{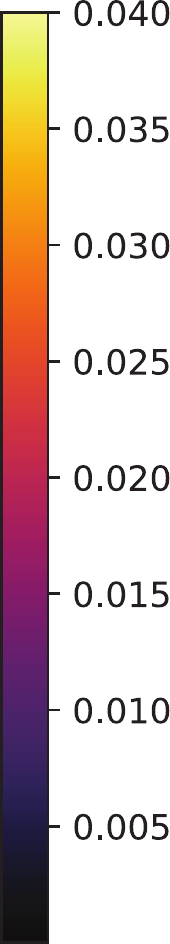}
        \caption*{}
    \end{minipage}}
    \caption{
        \textbf{A heatmap of which frequencies are used for classification.} We extracted the weight vector of the regression classifier trained on the different upsampling techniques. Then, we mapped it back to the corresponding frequencies. We plot the absolute value of the individual weights and clip their maximum value to 0.04 for better visibility. 
    }
    \label{fig:upsample:lasso}
\end{figure*}

\paragraph{Experiment Setup}
We sample 16,000 images from both the training data and the generator of the \styleG{}, respectively.
We split each set into 10,000  training, 1,000 validation and 5,000 test images, resulting in a training set of 20,000, a validation set of 2,000, and a test set of 10,000 samples.
For training a model on samples in the image domain, we normalize the pixel values to the range $[-1, 1]$.
In the frequency domain, we first convert the images using DCT, then log-scale the coefficients and, finally, normalize them by removing the mean and scaling to unit variance.
We optimize the linear regression models using the Adam optimizer~\cite{kingma2014adam} with an initial learning rate of $0.001$, minimizing the binary cross-entropy with $l_2$ regularization.
We select the regularization factor $\lambda$ via grid search from $\lambda \in \{10^{-1}, 10^{-2}, 10^{-3}, 10^{-4} \}$ on the validation data, picking the one with the best score.

\paragraph{Results}
The results of our experiments are listed in Table~\ref{tab:detection:style}, where we report the classification accuracy on the test set.
As depicted in Figure~\ref{fig:stats:flickr}, \styleG{} generates convincing face images, which are able to fool humans.
However, they seem to exhibit consistent patterns, since even the simple linear classifier reaches a non-trivial accuracy in the image domain.
In the frequency domain, however, we can perfectly separate the data set, with a  classification accuracy of 100$\,\%$ on the test set.

\subsection{Different Upsampling Techniques}
\label{chap:detection:upsampling}

We want to investigate if more elaborate upsampling techniques can thwart the detection.
To this end, we examine the different \styleG{} instances described in Section~\ref{chap:upsampling} (\ie, those who utilize nearest neighbour/bilinear/binomial upsampling).
Again, we train a ridge-regression on both the raw pixel values as well as DCT coefficients.
The experiment setup is the same as in the experiment described in Section~\ref{chap:detection:detection}

\paragraph{Results}
In Table~\ref{tab:detection:up}, we report the classification accuracy of the test set.
As one would intuitively suspect, heavier anti-aliasing reduces the accuracy of the classifier. 
Interestingly, this also seems to remove the consistent patterns exploited in the image domain.
Note that these samples are generated at a much lower resolution ($256 \times 256$) than the ones examined in Section~\ref{chap:detection:detection} ($1024 \times 1024$).
Thus, they require less upsampling operations and exhibit less artifacts, making a full linear separation impossible.
However, we can still reach $100\,\%$ test accuracy on the different data sets when we utilize a non-linear method like the CNN used in Section~\ref{chap:detection:identification}.

The small accuracy drop for the images generated by StyleGAN using binomial upsampling in comparison to the one using bilinear upsampling is surprising.
When we examine the corresponding mean spectrum (Figure~\ref{fig:upsample:comparison}), it contains less artifacts obvious to humans.
To verify that the classifier indeed utilizes the upsampling artifacts for its decision,
we perform an additional experiment: 
we train a logistic regression classifier with $l_1$ penalty (LASSO) on the data sets.
We then extract the corresponding weight vector and map it back to the corresponding frequencies.
Since the $l_1$ penalty forces weights to zero which play a minor role
in the classification, the remaining weights show that the high coefficients correspond to the frequencies which impact the decision the most.
The results are depicted in Figure~\ref{fig:upsample:lasso} and reveal that the binomial classifier still utilizes a grid-like structure to separate these images.
Additionally, we present the absolute difference of the mean spectra with respect to the training data in the supplementary material.

\subsection{Source Identification}
\label{chap:detection:identification}

The experiments described in this section are based on \citet{yu2019attributing}'s approach, and investigate how precisely a generated image's underlying architecture can be classified:
Yu~\etal{} trained four different GANs (\proG{}~\cite{karras2018progressive}, \sndG{}~\cite{miyato2018spectral}, \cramerG{}~\cite{bellemare2017cramer}, and  \mmdG{}~\cite{binkowski2018demystifying}) on the CelebA~\cite{liu2015faceattributes} and LSUN bedrooms dataset~\cite{yu15lsun} and generated images from all models. Then, based on these images, a classifier was trained that assigned an image to the corresponding subset class it belongs to, \ie, either being real, or being generated by \proG{}, \sndG{}, \cramerG{}, or \mmdG{}.

\paragraph{Experimental Setup}
The experiments are conducted on images of resolution $128 \times 128$.
We converted both data sets (i.\,e., celebA and LSUN bedrooms) specified by~\citet{yu2019attributingGithub}.
For each data set, we utilize their pre-trained models to sample 150,000 images from each GAN, and take another 150,000 real images randomly sampled from the underlying training data set.
We then partition these samples into 100,000 training, 20,000 validation and 30,000 test images, resulting in a combined set of 500,000 training, 100,000 validation and 150,000 test images.
We analyze the performance of different classifiers trained both on the images in their original representation (i.\,e., raw pixels) and after applying DCT: K-nearest-neighbor, Eigenfaces~\cite{sirovich1987low}), a CNN-based classifier developed by~\citet{yu2019attributing}, and a steganalysis method based on photo-response non-uniformity (PRNU) patterns by~\citet{marra2019gans}. These patterns also build on frequency information and utilizes high-pass filtered images to extract residuals information common to specific generators.

Moreover, we trained a shallow CNN\footnote{
    Using a CNN enabled a direct comparison to the performance of an analogous model on the raw-pixel input, where a CNN is the model of choice. 
    In preliminary experiments, we also tried utilizing simple feed forwad NNs, however, only a CNN was able to fully separate the data.
}, with only four convolution layers, to demonstrate that frequency information can significantly reduce the needed computation resources.
Details on the architecture can be found in the supplementary material.
Yu~\etal{} used a very large CNN with roughly 9 million parameters. In contrast, our CNN only utilizes around 170,000 parameters ($\sim$ 1.9\,\%). 
During training, we utilize the validation set for hyperparameter tuning and employ early stopping
We trained on log-scaled and normalized DCT coefficients.
For raw pixel, we scaled the values to the range $[\,-1, 1\,]$, except for the PRNU-based method, which operates directly on image data.

For training our CNN, we use the Adam optimizer with an initial learning rate of $0.001$ and a batch size of $1024$, minimizing the cross-entropy loss of the model.
For Yu~\etal{}'s CNN, we used the parameters specified in their repository~\cite{yu2019attributingGithub}.
We train their network for 8,000 mini-batch steps.
During first tests, we discovered their network usually converges at around 4,000 steps. Hence, we doubled the number of steps for the final evaluation and picked the best performing instance (measured on the validation set).
We also trained a variant of their classifier on DCT-transformed images.
This version usually converges around step 300, however, we are conservative and train for 1,000 steps.

For the PRNU classifier, we utilize the implementation of~\citet{luca_bondi_2019_2554965}. We perform a grid search for the wavelet decomposition level~$l \in \{1, 2, 3, 4\}$ and the estimated noise power~$\sigma \in \{0.05, \dots, 1\}$ with step size $0.05$. For the Eigenfaces-based classifier, we use Principal Component Analysis~(PCA) to reduce the dimensionality to a variance threshold~$v$ and train a linear Support Vector Machine~(SVM) on the transformed training data. 
We select the variance threshold~$v \in \{0.25, 0.5, 0.95\}$, the SVMs regularization parameter~$C \in \{0.0001, 0.001, 0.01, 0.1\}$, and the number of considered neighbors~$k \in \{1, 2^{\kappa}+1\}$ with $\kappa \in \{1,\dots, 10\}$, for the kNN classifier, via grid search over the validation set.

As the PRNU fingerprint algorithm does not require a large amount of training     
data~\cite{marra2019gans} and scaling more traditional methods to such large data sets is notoriously hard (i.\,e., kNN and Eigenfaces), we use a subset of 100,000 training samples and report the accuracy on 25,000 test samples.

\paragraph{Results}

\begin{table}[t!]
    \centering
    \caption{\textbf{The results of the source identification.}
    We report the test set accuracy, the gain in the frequency domain and highlight the best score in \textbf{bold}.
    }
      \resizebox{\linewidth}{!}{
    \begin{tabular}{lrrrr}
        \toprule
        Method                                  & LSUN      & Gain              & CelebA    & Gain              \\ \midrule
        
        kNN                                     & 39.96\,\%   &                 & 29.22\,\%   &                   \\ 
        kNN-DCT                                 & 81.56\,\%   & + 41.60\,\%     & 71.58\,\%   & + 42.36\,\%         \\ \midrule
        
        Eigenfaces                              & 47.07\,\%   &                 & 57.58\,\%   &                   \\ 
        Eigenfaces-DCT                          & 94.31\,\%   & + 47.24\,\%     & 88.39\,\%   & + 30.81\,\%         \\ \midrule
        
        PRNU~\citeauthor{marra2019gans}         & 64.28\,\%   &                 & 80.09\,\%   &                   \\ \midrule
        
        CNN~\citeauthor{yu2019attributing}      & 98.33\,\%   &                 & 99.70\,\%   &                   \\
        CNN~\citeauthor{yu2019attributing}-DCT  & 99.61\,\%   & + 1.28\,\%      & \textbf{99.91}\,\% & + 0.21\,\%   \\ \midrule
        
        CNN-Pixel                               & 98.95\,\%   &                   &  97.80\,\%  &                   \\
        CNN-DCT                                 & \textbf{99.64\,\%} & + 0.69\,\%   & 99.07\,\%   & + 1.27\,\%          \\\bottomrule
    \end{tabular}}
    \label{tab:detection:fingerprinting}
\end{table}

\begin{table}[t]
    \centering
    \caption{\textbf{The results of using only 20\,\% of the original data.}. We report the accuracy on the test set and the accuracy loss compared to the corresponding network trained on the full data set.}
      \resizebox{\linewidth}{!}{
    \begin{tabular}{lrrrr}
        \toprule
        Method                                  & LSUN      & Loss              & CelebA        & Loss  \\ \midrule
        
        CNN~\citeauthor{yu2019attributing}      & 92.30\,\% &   -6.03\,\%   &   98.57\,\%   &   -1.13\,\%   \\
        CNN~\citeauthor{yu2019attributing}-DCT  & 99.39\,\% &   -0.22\,\%   &   99.58\,\%   &   -0.33\,\%   \\  \midrule
            
        CNN-Pixel                               & 85.82\,\% &   -13.13\,\%  &   96.33\,\%   &   -1.47\,\%   \\
        CNN-DCT                                 & 99.14\,\% &   -0.50\,\%   &   98.47\,\%   &   -0.60\,\%   \\ \bottomrule
    \end{tabular}}
    \label{tab:detection:trainingdata}
\end{table}

\begin{table*}[t!]
    \centering
    \caption{\textbf{Results of common image perturbations on LSUN bedrooms.} 
    We report the accuracy on the test set.
    Best score is highlighted in \textbf{bold}.
    The column \textit{CD} refers to the performance of the corresponding classifier trained on clean data and evaluated on perturbed test data.
    In the column \textit{PD}, we depict the performance when trained on training data which is also perturbed.
    }
    \resizebox{\linewidth}{!}{
    \begin{tabular}{lcccccccccc}
        \toprule
        & \multicolumn{2}{c}{Blur} & \multicolumn{2}{c}{Cropped} & \multicolumn{2}{c}{Compression} & \multicolumn{2}{c}{Noise} & \multicolumn{2}{c}{Combined} \\
         & CD & PD & CD & PD & CD & PD & CD & PD & CD & PD  \\ \midrule
        CNN-Pixel & 60.56\,\% & 88.23\,\% & 74.49\,\% & 97.82\,\% & 68.66\,\% & 78.67\,\% & \textbf{59.51}\,\% & 78.18\,\% & 65.98\,\% & 83.54\,\%  \\ 
        CNN-DCT & \textbf{61.42}\,\% & \textbf{93.61}\,\% & \textbf{83.52}\,\% & \textbf{98.83}\,\% & \textbf{71.86}\,\% & \textbf{94.83}\,\% & 48.99\,\% & \textbf{89.56}\,\% & \textbf{67.76}\,\% & \textbf{92.17}\,\%  \\ \bottomrule 
    \end{tabular}}
    \label{tab:detection:image_pertubations}
    \begin{flushleft}
    \small \vspace{-0.25em}
    CD: Clean Data; PD: Perturbed Data\\
    \end{flushleft}
\end{table*}

The results of our experiments are presented in Table~\ref{tab:detection:fingerprinting}.
We report the accuracy computed over the test set.
For each method, we additionally report the gain in accuracy when trained on DCT coefficients.

The use of the frequency domain significantly improves the performance of all tested classifiers, which is in line with our findings from the previous sections.
The simpler techniques improve the most, with kNN gaining a performance boost of roughly 42\,\% and Eigenfaces improving by 47.24\,\% and 30.81\,\%, respectively.
Our shallow CNN already achieves high accuracy when it is trained on raw pixels (CNN-Pixel), but it still gains a performance boost (+0.69\,\% and 1.27\,\%, respectively). 
The CNN employed by~\citet{yu2019attributing} mirrors the result of our classifier.
Additionally, it seems to be important to utilize the entire frequency spectrum, since the PRNU-based classifier by~\citet{marra2019gans} achieves much lower accuracy.

The performance gains of classifiers trained on the frequency domain are difficult to examine by solely looking at the accuracy.
Thus, we consider the error rates instead:
Examining our shallow CNN, we decrease the error rate from $1.05\,\%$ and $2.20\,\%$ (without the use of the frequency domain) to $0.36\,\%$ and $0.93\,\%$, which corresponds to a reduction by $66\,\%$ and $50\,\%$, respectively. 
These results are mirrored for the CNN by~\citeauthor{yu2019attributing}, where the rates drop from $1.76\,\%$ and $0.3\,\%$ to $0.39\,\%$ and $0.09\,\%$, \ie, a reduction by $75\,\%$ and $66\,\%$.

\begin{figure}[t!]
    \centering
        \includegraphics[width=\linewidth]{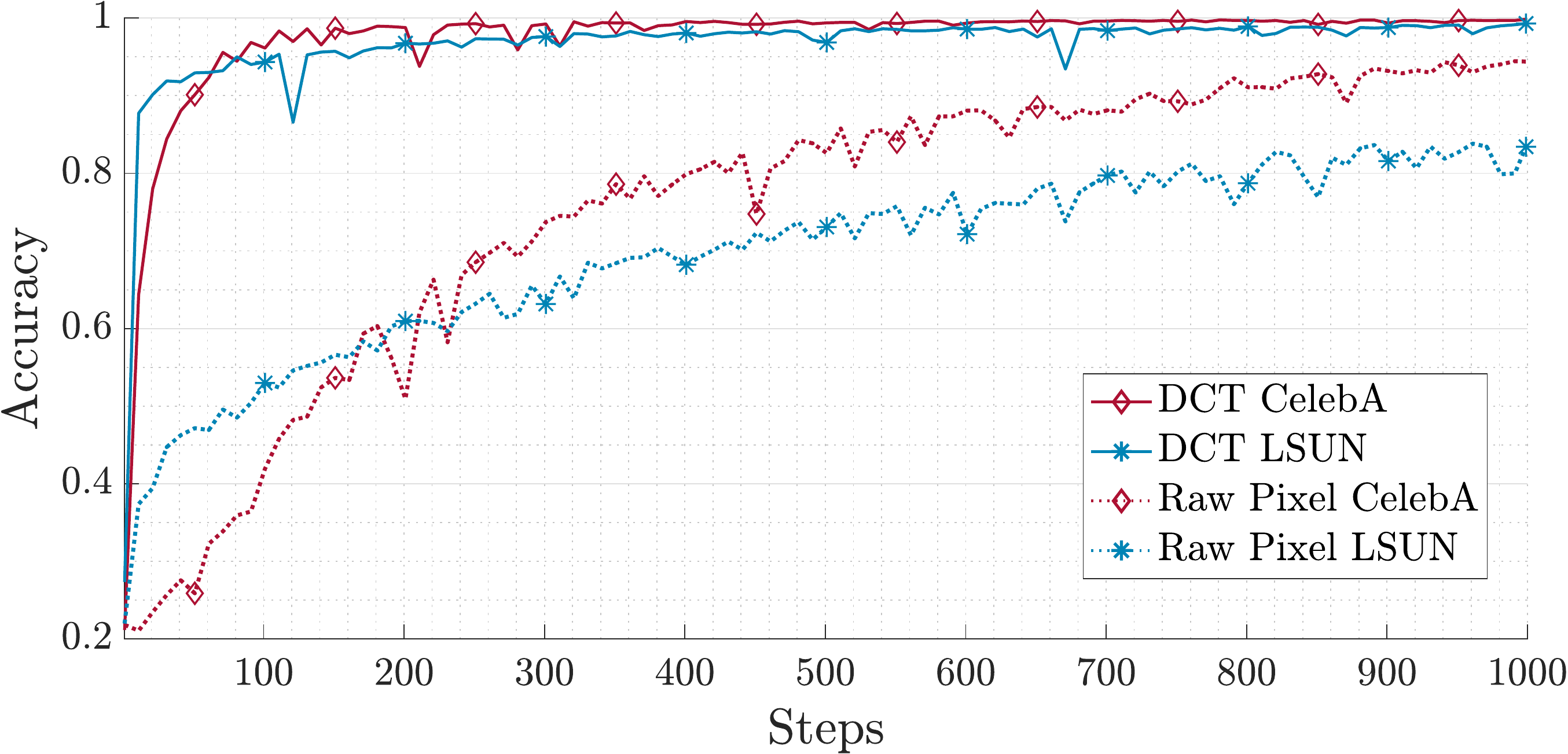}
    \caption{\textbf{Validation accuracy for the CCN-classifier by~\citeauthor{yu2019attributing}}. We report the validation accuracy during training for the first 1,000 gradient steps, while dropping off in the pixel domain.}
    \label{fig:eval:theirs}
\end{figure}

\begin{figure}[t!]
    \centering
            \includegraphics[width=\linewidth]{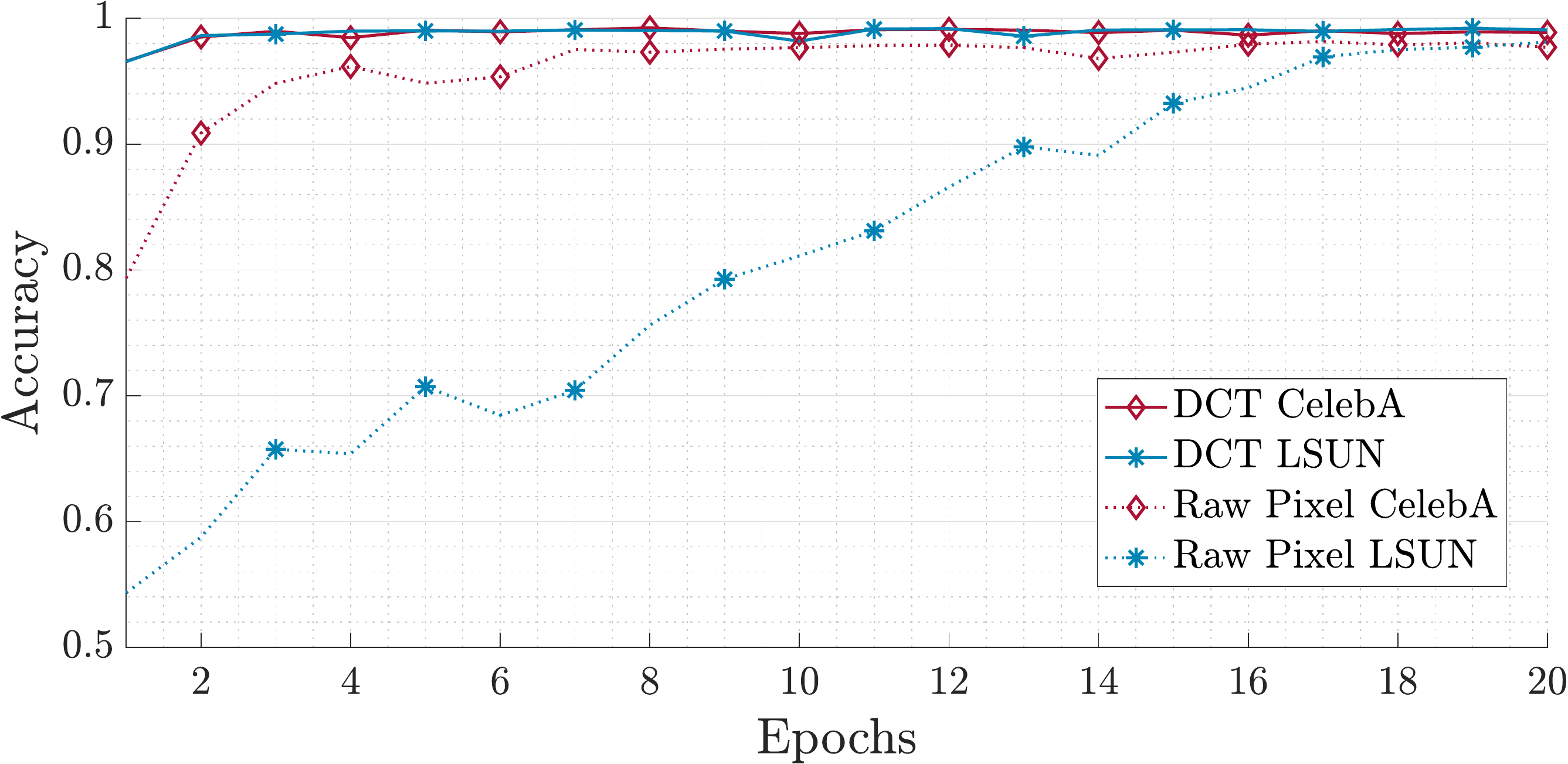}
    \caption{\textbf{Validation accuracy for our CNN-classifier}. We report the validation accuracy during training for the first 20 epochs.}
    \label{fig:eval:ours}
\end{figure}

\subsection{Training in the Frequency Domain}
\label{chap:detection:training}

During our experiments we noticed two additional benefits when performing classification in the frequency domain.
First, the models trained in the frequency domain require significantly less training data to achieve a high accuracy.
Second, training in the frequency domain is substantially easier.

We retrained the network by~\citeauthor{yu2019attributing} and our classifier with only 20\,\% of the original data and reevaluate them.
The results are presented in Table~\ref{tab:detection:trainingdata}.
In the frequency domain, both classifiers retain a high accuracy.
However, both drop significantly when trained on raw pixels. 
Note that the bigger classifier is able to compensate better for the lack of training data.

We have plotted the validation accuracy during training for both the CNN classifier by~\citeauthor{yu2019attributing}, as well as for our own in Figure~\ref{fig:eval:theirs} and Figure~\ref{fig:eval:ours}, respectively.
For both classifiers and both datasets (CelebA and LSUN), the DCT variant of the classifiers converges significantly faster.

\subsection{Resistance Against Common Image Perturbations}
\label{chap:detection:pertubations}

We want to examine if the artifacts persist through post-processing operations when the images get uploaded to websites like Facebook or Twitter.
Thus, we evaluate the resistance of our classifier against common image perturbations, namely: blurring, cropping, compression, adding random noise, and a combination of all of them.

\paragraph{Experimental Setup}
When creating the perturbed data with one kind of perturbation, for each data set we iterate through all images and apply the perturbation with a probability of 50\,\%.
This creates new data sets with about half the images perturbed, these again get divided into train/validation/tests sets, accuracy is reported for the test set.
For generating a data set with a combination of different perturbations, we cycle through the different corruptions in the order: blurring, cropping, compression, noise; and apply these with a probability of 50\,\%.
The corruptions are described in the following:
\begin{itemize}[topsep=0pt, itemsep=0pt, partopsep=4pt, parsep=4pt]
    \item \textbf{Blurring} applies Gaussian filtering with a kernel size randomly sampled from $(3,\,5,\,7,\,9)$.
    \item \textbf{Cropping} randomly crops the image along both axes. The percentage to crop is sampled from $U(5, 20)$. The cropped image is upsampled to its original resolution.
    \item \textbf{Compression} applies JPEG compression, the remaining quality factor is sampled from $U(10, 75)$.
    \item \textbf{Noise} adds i.i.d.~Gaussian Noise to the image. The variance of the Gaussian distribution is randomly sampled from $U(5.0, 20.0)$.
\end{itemize}
We additionally evaluate how well we can defend against common image perturbations by augmenting the training data with  perturbed data, \ie, we train a network with a training set that has also been altered by image perturbations.

\paragraph{Results}
The results are presented in Table~\ref{tab:detection:image_pertubations}.
Overall, our results show that the DCT variants of the classifiers
are more robust w.r.t.~all perturbations except of noise.

While the robustness to perturbations is increased, it is still possible to attack the classifier with adversarial examples.
In fact, subsequent work has demonstrated that it is possible to evade our classifier with specifically crafted perturbations~\cite{carlini2020evading}.
 \section{Discussion and Conclusion}
\label{chap:conclusion}

In this paper, we have provided a comprehensive analysis on the frequency spectrum exhibited by images generated from different GAN architectures.
Our main finding is that \emph{all} spectra contain artifacts common across architectures, data sets, and resolutions.
Speculating that these artifacts stem from upsampling operations, we experimented with different upsampling techniques which confirm our hypothesis. 
We than demonstrated that the frequency spectrum can be used to efficiently and accurately separate real from deep fake images.
We have found that the frequency domain both helps in enhancing simple (linear) models, as well as, more complex CNN-based methods, while simultaneously yielding better resistance against image perturbations.
Compared to hand-crafted, frequency-based methods, we discovered that the entire frequency spectrum can be utilized to achieve much higher performance.
We argue that our method will remain usable in the future because it relies on a fundamental property of today's GANs, \ie, the mapping from low dimensional latent space to a higher dimensional data space.

One suitable approach to mitigate this problem could be to remove upsampling methods entirely. However, this implies two problems.
First, this would remove the advantages of having a compact latent space altogether. 
Second, an instance of \styleG{} used to generate the pictures depicted in Figure~\ref{fig:stats:flickr} already needs 26.2M~\cite{karras2019style} parameters.
Removing the low-dimensional layers and only training at full resolution seems infeasible, at least for the foreseeable future.

Another approach could be to train GANs to generate consistent image spectra.
We experimented with both introducing a second DCT-based discriminator, as well as, regularizing the generator's loss function with a DCT-based penalty.
Unfortunately, neither approach led to better results.
Either the penalty or the second discriminator were weighted too weak and had no effect, or the DCT-based methods dominated and led to training collapse.
We leave the exploration of these methods as an interesting question for future work.

\section*{Acknowledgements}
We would like to thank our colleagues Cornelius Aschermann, Steffen Zeiler, Sina D{\"a}ubener, Philipp G{\"o}rz, Erwin Quiring, Konrad Rieck, and our anonymous reviewers for their valuable feedback and fruitful discussions. 
This work was supported by the Deutsche Forschungsgemeinschaft (DFG, German Research Foundation) under Germany's Excellence Strategy -- EXC-2092  \textsc{CaSa} -- 390781972.

\bibliographystyle{icml2020.bst}

\begin{thebibliography}{60}
\providecommand{\natexlab}[1]{#1}
\providecommand{\url}[1]{\texttt{#1}}
\expandafter\ifx\csname urlstyle\endcsname\relax
  \providecommand{\doi}[1]{doi: #1}\else
  \providecommand{\doi}{doi: \begingroup \urlstyle{rm}\Url}\fi

\bibitem[Arjovsky et~al.(2017)Arjovsky, Chintala, and
  Bottou]{arjovsky2017wasserstein}
Arjovsky, M., Chintala, S., and Bottou, L.
\newblock Wasserstein gan.
\newblock In \emph{International Conference on Machine Learning (ICML)}, 2017.

\bibitem[Azulay \& Weiss(2018)Azulay and Weiss]{azulay2018deep}
Azulay, A. and Weiss, Y.
\newblock Why do deep convolutional networks generalize so poorly to small
  image transformations?
\newblock \emph{arXiv preprint arXiv:1805.12177}, 2018.

\bibitem[Bappy et~al.(2017)Bappy, Roy-Chowdhury, Bunk, Nataraj, and
  Manjunath]{bappy2017exploiting}
Bappy, J.~H., Roy-Chowdhury, A.~K., Bunk, J., Nataraj, L., and Manjunath, B.
\newblock Exploiting spatial structure for localizing manipulated image
  regions.
\newblock In \emph{IEEE International Conference on Computer Vision (ICCV)},
  2017.

\bibitem[Bayar \& Stamm(2016)Bayar and Stamm]{bayar2016deep}
Bayar, B. and Stamm, M.~C.
\newblock A deep learning approach to universal image manipulation detection
  using a new convolutional layer.
\newblock In \emph{ACM Workshop on Information Hiding and Multimedia Security},
  2016.

\bibitem[Bellemare et~al.(2017)Bellemare, Danihelka, Dabney, Mohamed,
  Lakshminarayanan, Hoyer, and Munos]{bellemare2017cramer}
Bellemare, M.~G., Danihelka, I., Dabney, W., Mohamed, S., Lakshminarayanan, B.,
  Hoyer, S., and Munos, R.
\newblock The cramer distance as a solution to biased wasserstein gradients.
\newblock \emph{arXiv preprint arXiv:1705.10743}, 2017.

\bibitem[Bestagini et~al.(2013)Bestagini, Milani, Tagliasacchi, and
  Tubaro]{bestagini2013local}
Bestagini, P., Milani, S., Tagliasacchi, M., and Tubaro, S.
\newblock Local tampering detection in video sequences.
\newblock In \emph{IEEE International Workshop on Multimedia Signal Processing
  (MMSP)}, 2013.

\bibitem[Bi{\'n}kowski et~al.(2018)Bi{\'n}kowski, Sutherland, Arbel, and
  Gretton]{binkowski2018demystifying}
Bi{\'n}kowski, M., Sutherland, D.~J., Arbel, M., and Gretton, A.
\newblock Demystifying mmd gans.
\newblock In \emph{International Conference on Learning Representations
  (ICLR)}, 2018.

\bibitem[Bondi \& Bonettini(2019)Bondi and Bonettini]{luca_bondi_2019_2554965}
Bondi, L. and Bonettini, N.
\newblock polimi-ispl/prnu-python: v.1.2.
\newblock \emph{Zenodo}, 2019.

\bibitem[Brock et~al.(2019)Brock, Donahue, and Simonyan]{brock2018large}
Brock, A., Donahue, J., and Simonyan, K.
\newblock Large scale {GAN} training for high fidelity natural image synthesis.
\newblock In \emph{International Conference on Learning Representations
  (ICLR)}, 2019.

\bibitem[Burton \& Moorhead(1987)Burton and Moorhead]{burton1987color}
Burton, G.~J. and Moorhead, I.~R.
\newblock Color and spatial structure in natural scenes.
\newblock \emph{Applied optics}, 1987.

\bibitem[Carlini \& Farid(2020)Carlini and Farid]{carlini2020evading}
Carlini, N. and Farid, H.
\newblock Evading deepfake-image detectors with white-and black-box attacks.
\newblock In \emph{IEEE Conference on Computer Vision and Pattern Recognition
  (CVPR) Workshops}, 2020.

\bibitem[Cozzolino et~al.(2017)Cozzolino, Poggi, and
  Verdoliva]{cozzolino2017recasting}
Cozzolino, D., Poggi, G., and Verdoliva, L.
\newblock Recasting residual-based local descriptors as convolutional neural
  networks: an application to image forgery detection.
\newblock In \emph{ACM Workshop on Information Hiding and Multimedia Security},
  2017.

\bibitem[Durall et~al.(2020)Durall, Keuper, and Keuper]{durall2020watch}
Durall, R., Keuper, M., and Keuper, J.
\newblock Watch your up-convolution: Cnn based generative deep neural networks
  are failing to reproduce spectral distributions.
\newblock In \emph{IEEE Conference on Computer Vision and Pattern Recognition
  (CVPR)}, 2020.

\bibitem[Field(1987)]{field1987relations}
Field, D.~J.
\newblock Relations between the statistics of natural images and the response
  properties of cortical cells.
\newblock \emph{Journal of the Optical Society of America. A, Optics and image
  science}, 1987.

\bibitem[Field(1999)]{field1999wavelets}
Field, D.~J.
\newblock Wavelets, vision and the statistics of natural scenes.
\newblock \emph{Philosophical Transactions of the Royal Society of London.
  Series A: Mathematical, Physical and Engineering Sciences}, 1999.

\bibitem[Fridrich(2009)]{fridrich2009digital}
Fridrich, J.
\newblock Digital image forensics.
\newblock \emph{IEEE Signal Processing Magazine}, 2009.

\bibitem[Fried et~al.(2019)Fried, Tewari, Zollh{\"o}fer, Finkelstein,
  Shechtman, Goldman, Genova, Jin, Theobalt, and Agrawala]{fried2019text}
Fried, O., Tewari, A., Zollh{\"o}fer, M., Finkelstein, A., Shechtman, E.,
  Goldman, D.~B., Genova, K., Jin, Z., Theobalt, C., and Agrawala, M.
\newblock Text-based editing of talking-head video.
\newblock \emph{ACM Transactions on Graphics (TOG)}, 2019.

\bibitem[Gonzalez \& Woods(1992)Gonzalez and Woods]{gonzalez2004digital}
Gonzalez, R.~C. and Woods, R.~E.
\newblock \emph{Digital Image Processing}.
\newblock Pearson, 1992.

\bibitem[Goodfellow et~al.(2014)Goodfellow, Pouget-Abadie, Mirza, Xu,
  Warde-Farley, Ozair, Courville, and Bengio]{goodfellow2014generative}
Goodfellow, I., Pouget-Abadie, J., Mirza, M., Xu, B., Warde-Farley, D., Ozair,
  S., Courville, A., and Bengio, Y.
\newblock Generative adversarial nets.
\newblock In \emph{Advances in Neural Information Processing Systems
  (NeurIPS)}, 2014.

\bibitem[Gulrajani et~al.(2017)Gulrajani, Ahmed, Arjovsky, Dumoulin, and
  Courville]{gulrajani2017improved}
Gulrajani, I., Ahmed, F., Arjovsky, M., Dumoulin, V., and Courville, A.~C.
\newblock Improved training of wasserstein gans.
\newblock In \emph{Advances in Neural Information Processing Systems
  (NeurIPS)}, 2017.

\bibitem[Hao(2019)]{hao2019deepfake}
Hao, K.
\newblock The biggest threat of deepfakes isn’t the deepfakes themselves.
\newblock \emph{MIT Technology Review}, 2019.

\bibitem[Kaggle(2019)]{kaggle2019comp}
Kaggle.
\newblock Generative dog images - experiment with creating puppy pics.
\newblock
  \href{https://www.kaggle.com/c/generative-dog-images}{https://www.kaggle.com/c/generative-dog-images},
  2019.

\bibitem[Karras et~al.(2018)Karras, Aila, Laine, and
  Lehtinen]{karras2018progressive}
Karras, T., Aila, T., Laine, S., and Lehtinen, J.
\newblock Progressive growing of {GAN}s for improved quality, stability, and
  variation.
\newblock In \emph{International Conference on Learning Representations
  (ICLR)}, 2018.

\bibitem[Karras et~al.(2019)Karras, Laine, and Aila]{karras2019style}
Karras, T., Laine, S., and Aila, T.
\newblock A style-based generator architecture for generative adversarial
  networks.
\newblock In \emph{IEEE Conference on Computer Vision and Pattern Recognition
  (CVPR)}, 2019.

\bibitem[Khosla et~al.(2011)Khosla, Jayadevaprakash, Yao, and
  Fei-Fei]{standford2011dog}
Khosla, A., Jayadevaprakash, N., Yao, B., and Fei-Fei, L.
\newblock Novel dataset for fine-grained image categorization.
\newblock In \emph{First Workshop on Fine-Grained Visual Categorization, IEEE
  Conference on Computer Vision and Pattern Recognition (CVPR)}, 2011.

\bibitem[Kingma \& Ba(2015)Kingma and Ba]{kingma2014adam}
Kingma, D.~P. and Ba, J.
\newblock Adam: A method for stochastic optimization.
\newblock In \emph{International Conference on Learning Representations
  (ICLR)}, 2015.

\bibitem[Kumar et~al.(2019)Kumar, Kumar, de~Boissiere, Gestin, Teoh, Sotelo,
  de~Br{\'e}bisson, Bengio, and Courville]{kumar2019melgan}
Kumar, K., Kumar, R., de~Boissiere, T., Gestin, L., Teoh, W.~Z., Sotelo, J.,
  de~Br{\'e}bisson, A., Bengio, Y., and Courville, A.~C.
\newblock Melgan: Generative adversarial networks for conditional waveform
  synthesis.
\newblock In \emph{Advances in Neural Information Processing Systems
  (NeurIPS)}, 2019.

\bibitem[Liu et~al.(2015)Liu, Luo, Wang, and Tang]{liu2015faceattributes}
Liu, Z., Luo, P., Wang, X., and Tang, X.
\newblock Deep learning face attributes in the wild.
\newblock In \emph{IEEE International Conference on Computer Vision (ICCV)},
  December 2015.

\bibitem[Luk{\'a}{\v{s}} et~al.(2006)Luk{\'a}{\v{s}}, Fridrich, and
  Goljan]{lukavs2006digital}
Luk{\'a}{\v{s}}, J., Fridrich, J., and Goljan, M.
\newblock Digital camera identification from sensor pattern noise.
\newblock \emph{IEEE Transactions on Information Forensics and Security}, 2006.

\bibitem[Lyu(2013)]{lyu2013natural}
Lyu, S.
\newblock Natural image statistics in digital image forensics.
\newblock In \emph{Digital Image Forensics}, pp.\  239--256. Springer, 2013.

\bibitem[Marra et~al.(2018)Marra, Gragnaniello, Cozzolino, and
  Verdoliva]{marra2018detection}
Marra, F., Gragnaniello, D., Cozzolino, D., and Verdoliva, L.
\newblock Detection of gan-generated fake images over social networks.
\newblock In \emph{IEEE Conference on Multimedia Information Processing and
  Retrieval (MIPR)}, 2018.

\bibitem[Marra et~al.(2019)Marra, Gragnaniello, Verdoliva, and
  Poggi]{marra2019gans}
Marra, F., Gragnaniello, D., Verdoliva, L., and Poggi, G.
\newblock Do gans leave artificial fingerprints?
\newblock In \emph{IEEE Conference on Multimedia Information Processing and
  Retrieval (MIPR)}, 2019.

\bibitem[McCloskey \& Albright(2018)McCloskey and
  Albright]{mccloskey2018detecting}
McCloskey, S. and Albright, M.
\newblock Detecting gan-generated imagery using color cues.
\newblock \emph{arXiv preprint arXiv:1812.08247}, 2018.

\bibitem[Mirza \& Osindero(2014)Mirza and Osindero]{mirza2014conditional}
Mirza, M. and Osindero, S.
\newblock Conditional generative adversarial nets.
\newblock \emph{arXiv preprint arXiv:1411.1784}, 2014.

\bibitem[Miyato et~al.(2018)Miyato, Kataoka, Koyama, and
  Yoshida]{miyato2018spectral}
Miyato, T., Kataoka, T., Koyama, M., and Yoshida, Y.
\newblock Spectral normalization for generative adversarial networks.
\newblock \emph{International Conference on Learning Representations (ICLR)},
  2018.

\bibitem[Mo et~al.(2018)Mo, Chen, and Luo]{mo2018fake}
Mo, H., Chen, B., and Luo, W.
\newblock Fake faces identification via convolutional neural network.
\newblock In \emph{ACM Workshop on Information Hiding and Multimedia Security},
  2018.

\bibitem[Nataraj et~al.(2019)Nataraj, Mohammed, Manjunath, Chandrasekaran,
  Flenner, Bappy, and Roy-Chowdhury]{nataraj2019detecting}
Nataraj, L., Mohammed, T.~M., Manjunath, B., Chandrasekaran, S., Flenner, A.,
  Bappy, J.~H., and Roy-Chowdhury, A.~K.
\newblock Detecting gan generated fake images using co-occurrence matrices.
\newblock \emph{Electronic Imaging}, 2019.

\bibitem[Odena et~al.(2016)Odena, Dumoulin, and Olah]{odena2016deconvolution}
Odena, A., Dumoulin, V., and Olah, C.
\newblock Deconvolution and checkerboard artifacts.
\newblock \emph{Distill}, 2016.

\bibitem[Petzka et~al.(2018)Petzka, Fischer, and
  Lukovnicov]{petzka2017regularization}
Petzka, H., Fischer, A., and Lukovnicov, D.
\newblock On the regularization of wasserstein gans.
\newblock In \emph{International Conference on Learning Representations
  (ICLR)}, 2018.

\bibitem[Radford et~al.(2016)Radford, Metz, and
  Chintala]{radford2015unsupervised}
Radford, A., Metz, L., and Chintala, S.
\newblock Unsupervised representation learning with deep convolutional
  generative adversarial networks.
\newblock In \emph{International Conference on Learning Representations
  (ICLR)}, 2016.

\bibitem[Razavi et~al.(2019)Razavi, van~den Oord, and
  Vinyals]{razavi2019generating}
Razavi, A., van~den Oord, A., and Vinyals, O.
\newblock Generating diverse high-fidelity images with vq-vae-2.
\newblock In \emph{Advances in Neural Information Processing Systems
  (NeurIPS)}, 2019.

\bibitem[Salimans et~al.(2016)Salimans, Goodfellow, Zaremba, Cheung, Radford,
  and Chen]{salimans2016improved}
Salimans, T., Goodfellow, I., Zaremba, W., Cheung, V., Radford, A., and Chen,
  X.
\newblock Improved techniques for training gans.
\newblock In \emph{Advances in Neural Information Processing Systems
  (NeurIPS)}, 2016.

\bibitem[Simonite(2019)]{simonite2019whichface}
Simonite, T.
\newblock Artificial intelligence is coming for our faces.
\newblock \emph{Wired}, 2019.

\bibitem[Sirovich \& Kirby(1987)Sirovich and Kirby]{sirovich1987low}
Sirovich, L. and Kirby, M.
\newblock Low-dimensional procedure for the characterization of human faces.
\newblock \emph{Journal of the Optical Society of America. A, Optics and image
  science}, 1987.

\bibitem[Song et~al.(2020)Song, Wu, Qian, He, and Loy]{song2020everybody}
Song, L., Wu, W., Qian, C., He, R., and Loy, C.~C.
\newblock Everybody's talkin': Let me talk as you want.
\newblock \emph{arXiv preprint arXiv:2001.05201}, 2020.

\bibitem[Tariq et~al.(2019)Tariq, Lee, Kim, Shin, and Woo]{tariq2019gan}
Tariq, S., Lee, S., Kim, H., Shin, Y., and Woo, S.~S.
\newblock Gan is a friend or foe? a framework to detect various fake face
  images.
\newblock In \emph{ACM/SIGAPP Symposium on Applied Computing}, 2019.

\bibitem[Thompson \& Lapowsky(2017)Thompson and
  Lapowsky]{thompson2017memewarfare}
Thompson, N. and Lapowsky, I.
\newblock How russian trolls used meme warfare to divide america.
\newblock \emph{Wired}, 2017.

\bibitem[Tolhurst et~al.(1992)Tolhurst, Tadmor, and
  Chao]{tolhurst1992amplitude}
Tolhurst, D., Tadmor, Y., and Chao, T.
\newblock Amplitude spectra of natural images.
\newblock \emph{Ophthalmic and Physiological Optics}, 1992.

\bibitem[Torralba \& Oliva(2003)Torralba and Oliva]{torralba2003statistics}
Torralba, A. and Oliva, A.
\newblock Statistics of natural image categories.
\newblock \emph{Network: computation in neural systems}, 2003.

\bibitem[Valle et~al.(2018)Valle, Cai, and Doshi]{valle2018tequilagan}
Valle, R., Cai, W., and Doshi, A.
\newblock Tequilagan: How to easily identify gan samples.
\newblock \emph{arXiv preprint arXiv:1807.04919}, 2018.

\bibitem[van~den Oord et~al.(2017)van~den Oord, Vinyals, et~al.]{van2017neural}
van~den Oord, A., Vinyals, O., et~al.
\newblock Neural discrete representation learning.
\newblock In \emph{Advances in Neural Information Processing Systems
  (NeurIPS)}, 2017.

\bibitem[Wang et~al.(2020)Wang, Wang, Zhang, Owens, and Efros]{wang2019cnn}
Wang, S.-Y., Wang, O., Zhang, R., Owens, A., and Efros, A.~A.
\newblock Cnn-generated images are surprisingly easy to spot... for now.
\newblock In \emph{IEEE Conference on Computer Vision and Pattern Recognition
  (CVPR)}, 2020.

\bibitem[{Wen-Hsiung Chen} et~al.(1977){Wen-Hsiung Chen}, {Smith}, and
  {Fralick}]{Chen1977}
{Wen-Hsiung Chen}, {Smith}, C., and {Fralick}, S.
\newblock A fast computational algorithm for the discrete cosine transform.
\newblock \emph{IEEE Transactions on Communications}, 1977.

\bibitem[West \& Bergstrom(2019)West and Bergstrom]{which2019website}
West, J. and Bergstrom, C.
\newblock Which face is real?
\newblock
  \href{http://www.whichfaceisreal.com}{http://www.whichfaceisreal.com}, 2019.

\bibitem[Yu et~al.(2015)Yu, Zhang, Song, Seff, and Xiao]{yu15lsun}
Yu, F., Zhang, Y., Song, S., Seff, A., and Xiao, J.
\newblock Lsun: Construction of a large-scale image dataset using deep learning
  with humans in the loop.
\newblock \emph{arXiv preprint arXiv:1506.03365}, 2015.

\bibitem[Yu et~al.(2019{\natexlab{a}})Yu, Davis, and Fritz]{yu2019attributing}
Yu, N., Davis, L.~S., and Fritz, M.
\newblock Attributing fake images to gans: Learning and analyzing gan
  fingerprints.
\newblock In \emph{IEEE International Conference on Computer Vision (ICCV)},
  2019{\natexlab{a}}.

\bibitem[Yu et~al.(2019{\natexlab{b}})Yu, Davis, and
  Fritz]{yu2019attributingGithub}
Yu, N., Davis, L.~S., and Fritz, M.
\newblock Attributing fake images to gans: Learning and analyzing gan
  fingerprints.
\newblock
  \href{https://github.com/ningyu1991/GANFingerprints}{https://github.com/ningyu1991/GANFingerprints},
  2019{\natexlab{b}}.

\bibitem[Zhang(2019)]{zhang2019shiftinvar}
Zhang, R.
\newblock Making convolutional networks shift-invariant again.
\newblock In \emph{International Conference on Machine Learning (ICML)}, 2019.

\bibitem[Zhou et~al.(2018)Zhou, Han, Morariu, and Davis]{zhou2018learning}
Zhou, P., Han, X., Morariu, V.~I., and Davis, L.~S.
\newblock Learning rich features for image manipulation detection.
\newblock In \emph{IEEE International Conference on Computer Vision (ICCV)},
  2018.

\bibitem[Zhu et~al.(2017)Zhu, Park, Isola, and Efros]{zhu2017unpaired}
Zhu, J.-Y., Park, T., Isola, P., and Efros, A.~A.
\newblock Unpaired image-to-image translation using cycle-consistent
  adversarial networks.
\newblock In \emph{IEEE International Conference on Computer Vision (ICCV)},
  2017.

\end{thebibliography}

\onecolumn
\renewcommand\thesection{\Alph{section}}
\setcounter{section}{0}

\section*{Supplementary Material}

In this supplementary material, we present all plots in full size, additional statistics, as well as details on our classifier architecture.
Note we depict statistics split into color channels only for the Kaggle dataset, since they are consistent with the ones computed over gray-scale images.

\section{FFHQ}

We plot the mean of the DCT spectrum of the Flicker-Faces-HQ (FFHQ) data set and an instance of \styleG{}.
We estimate $\E{\dctcom{I}}$ by averaging over 10,000 images.
Additionally, we plot the absolute difference between the two spectra, notice the additional artifacts scattered throughout the spectrum which are not on the grid.

\begin{figure}[h!]
    \centering
    \begin{subfigure}{.48\textwidth}
        \centering
        \includegraphics[width=\textwidth]{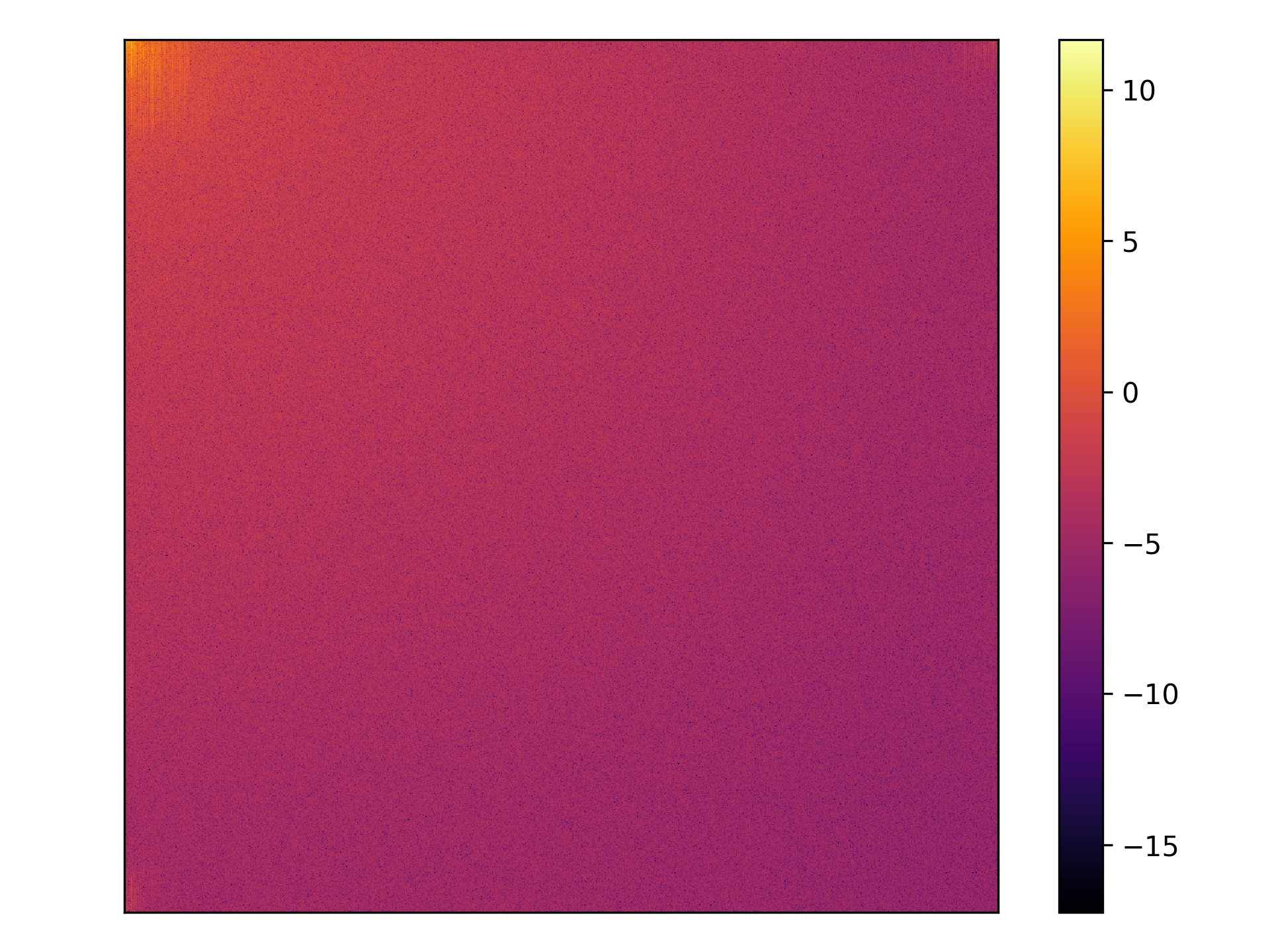}
        \caption*{FFHQ}
    \end{subfigure}
    \begin{subfigure}{.48\textwidth}
        \centering
        \includegraphics[width=\textwidth]{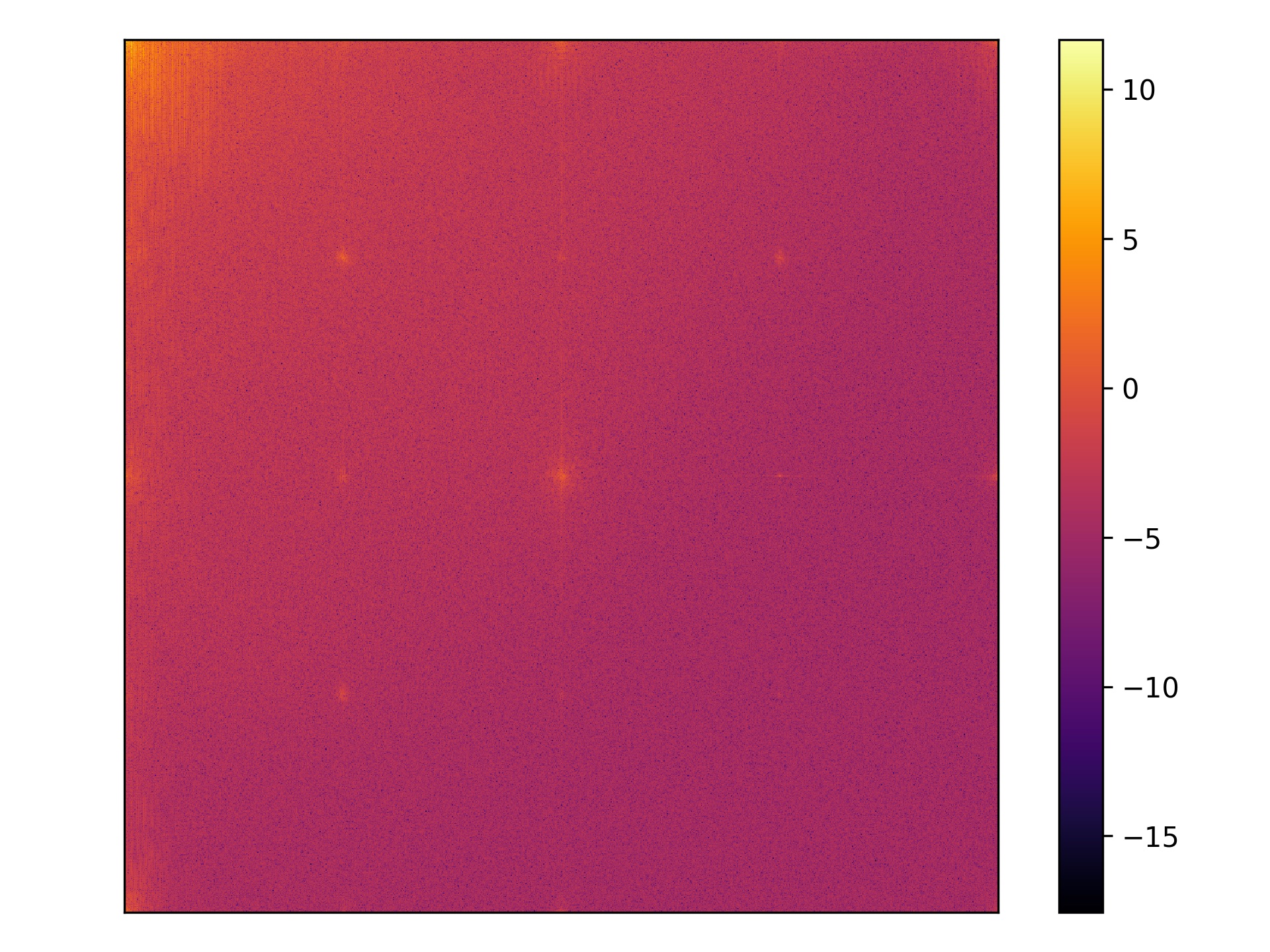}
        \caption*{StyleGAN}
    \end{subfigure}
    \caption{\textbf{The frequency spectrum for real and generated faces (grayscale)}}
\end{figure}

\begin{figure}[h!]
    \centering
    \includegraphics[width=.6\textwidth]{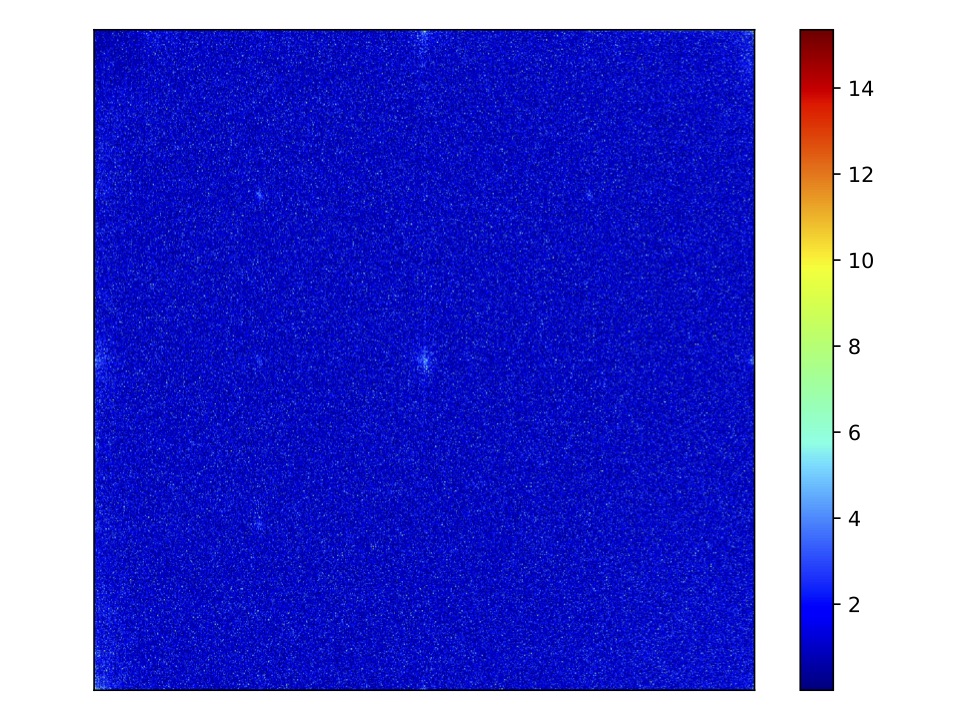}
    \caption*{$\mid \E{\dctcom{\text{FFHQ}}} - \E{\dctcom{\text{StyleGAN}}} \mid$}
    \caption{\textbf{The absolute difference between the spectra (grayscale)}}
\end{figure}

Here we also present a plot of a LASSO-regression trained on the FFHQ data set.
In context with Figure~\ref{fig:stats:flickr}, this makes sense, since compared to the real spectrum, the generated images diverge most in the higher frequencies (real images contain very little energy here).
Note that the high frequencies can also be attributed to upsampling operations~\citet{durall2020watch}.

\begin{figure}[ht]
    \centering
    \includegraphics[width=0.5\linewidth]{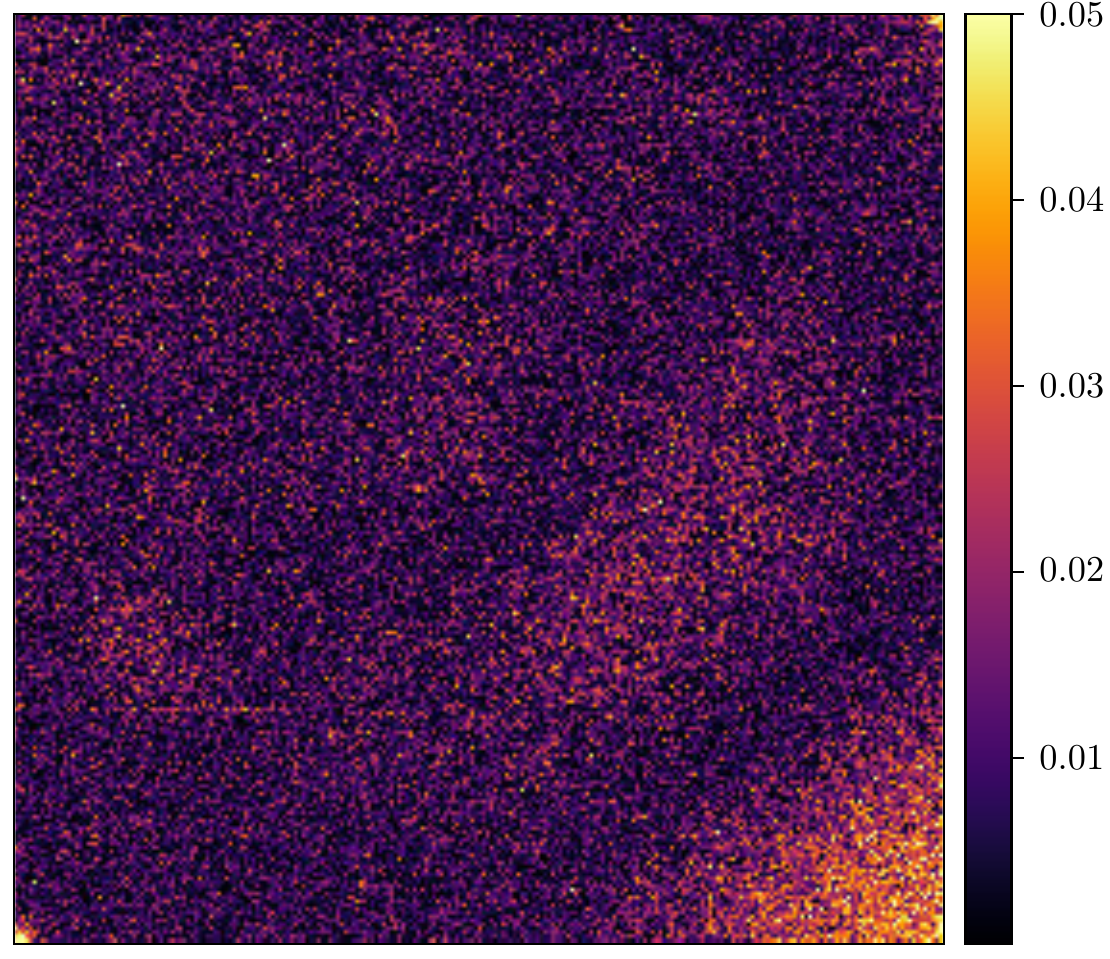}
    \caption{\textbf{A heatmap of which frequencies the LASSO-regression uses.} We extracted the weight vector of the regression classifier and mapped it back to the corresponding frequencies. We plot the absolute value of the individual weights and clip their maximum value to 0.05 for better visibility. Note the general focus towards higher frequencies, as well as the top right and lower left corner.}
    \label{fig:detection:weights}
\end{figure}
 
\clearpage 
\section{Kaggle}

We plot the mean of the DCT spectrum of the Standford dog data set and images generated by different instances of GANs (\bigG{}, \proG{}, \styleG{}, \sndG{}) trained upon it.
We estimate $\E{\dctcom{I}}$ by averaging over 10,000 images.

\begin{figure}[h!]
    \centering
    \begin{subfigure}{.25\textwidth}
        \centering
        \includegraphics[width=\textwidth]{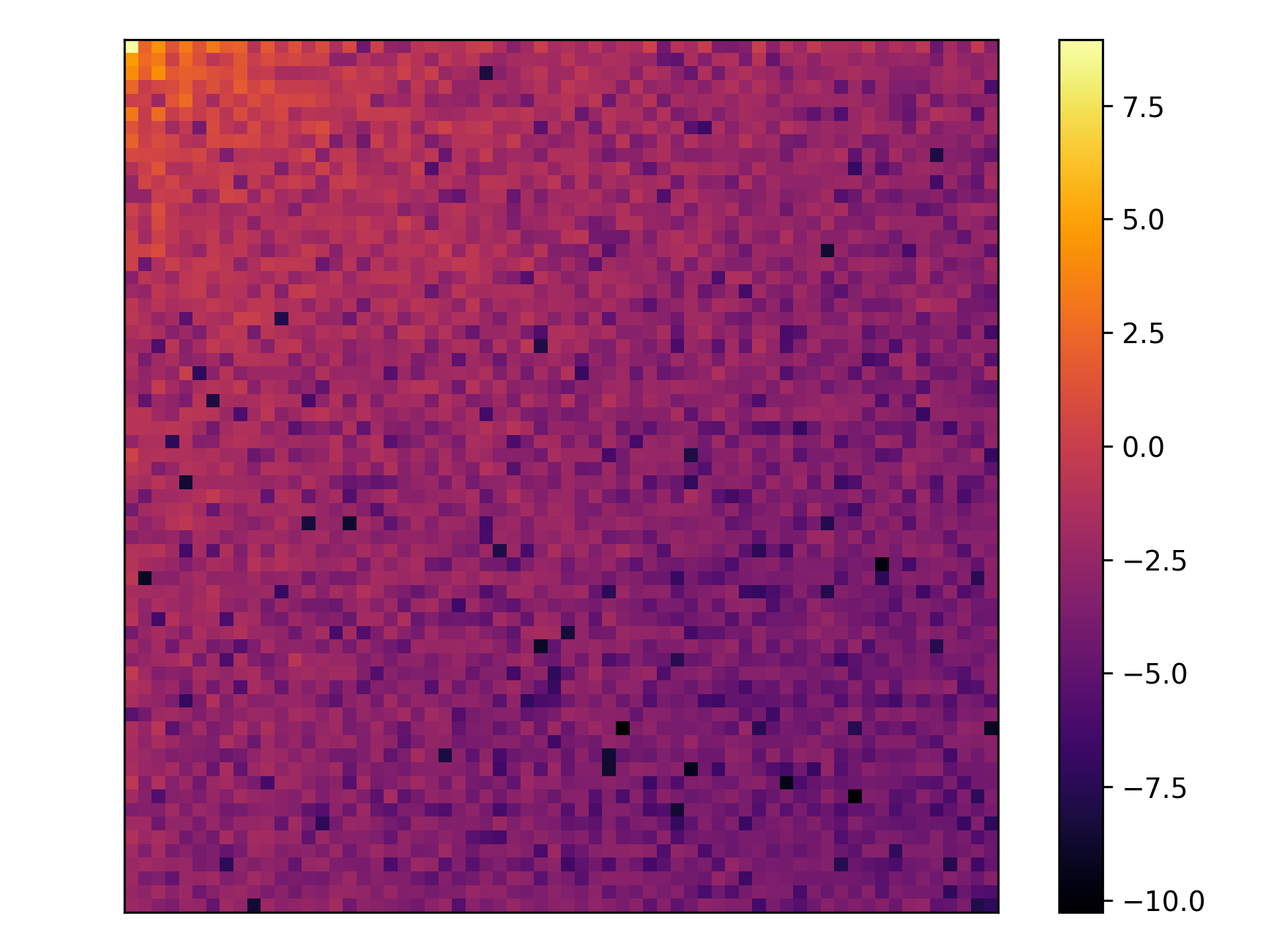}
        \caption*{Stanford dogs (red)}
    \end{subfigure}
    \begin{subfigure}{.25\textwidth}
        \centering
        \includegraphics[width=\textwidth]{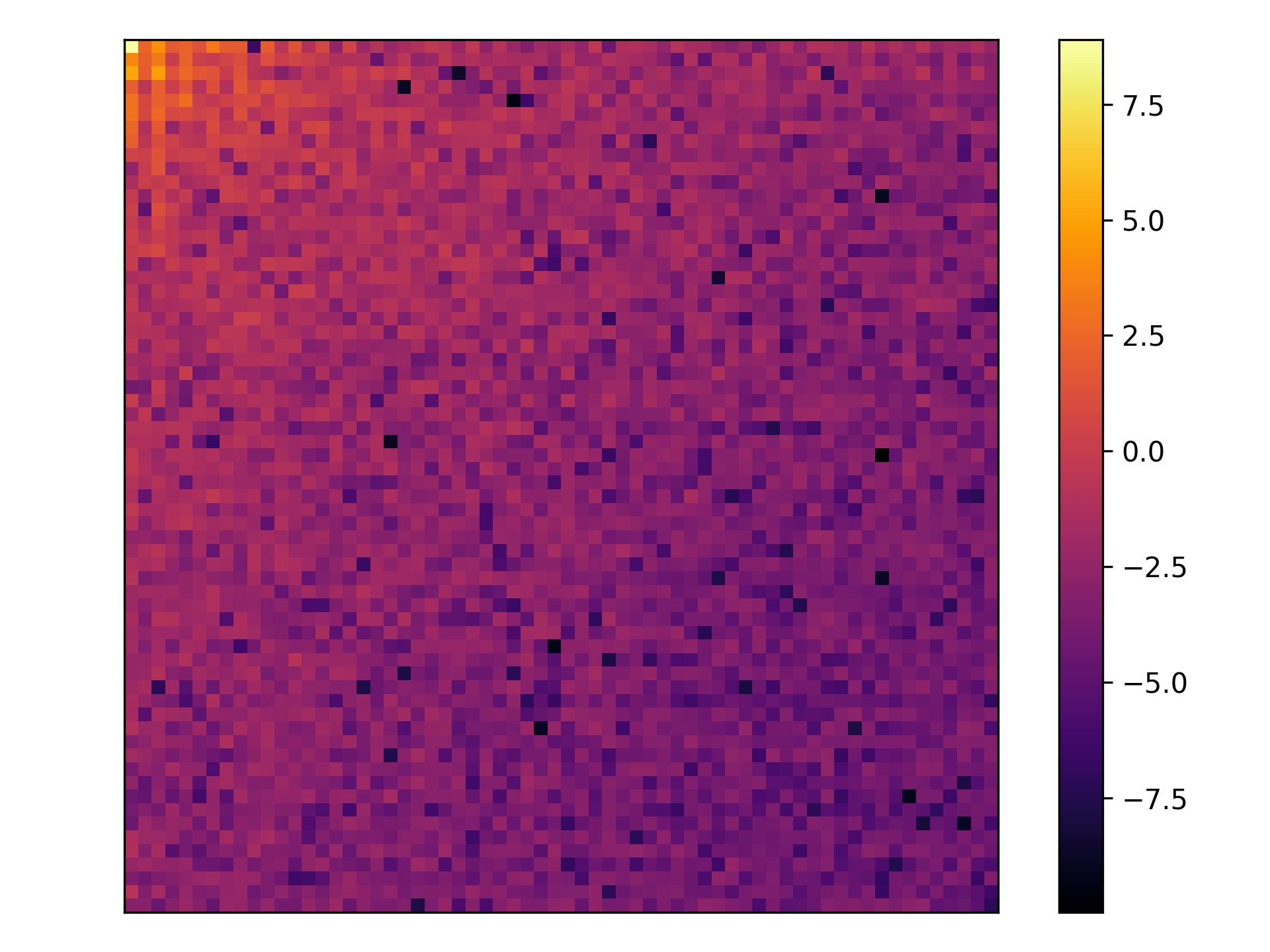}
        \caption*{Stanford dogs (green)}
    \end{subfigure}
    \begin{subfigure}{.25\textwidth}
        \centering
        \includegraphics[width=\textwidth]{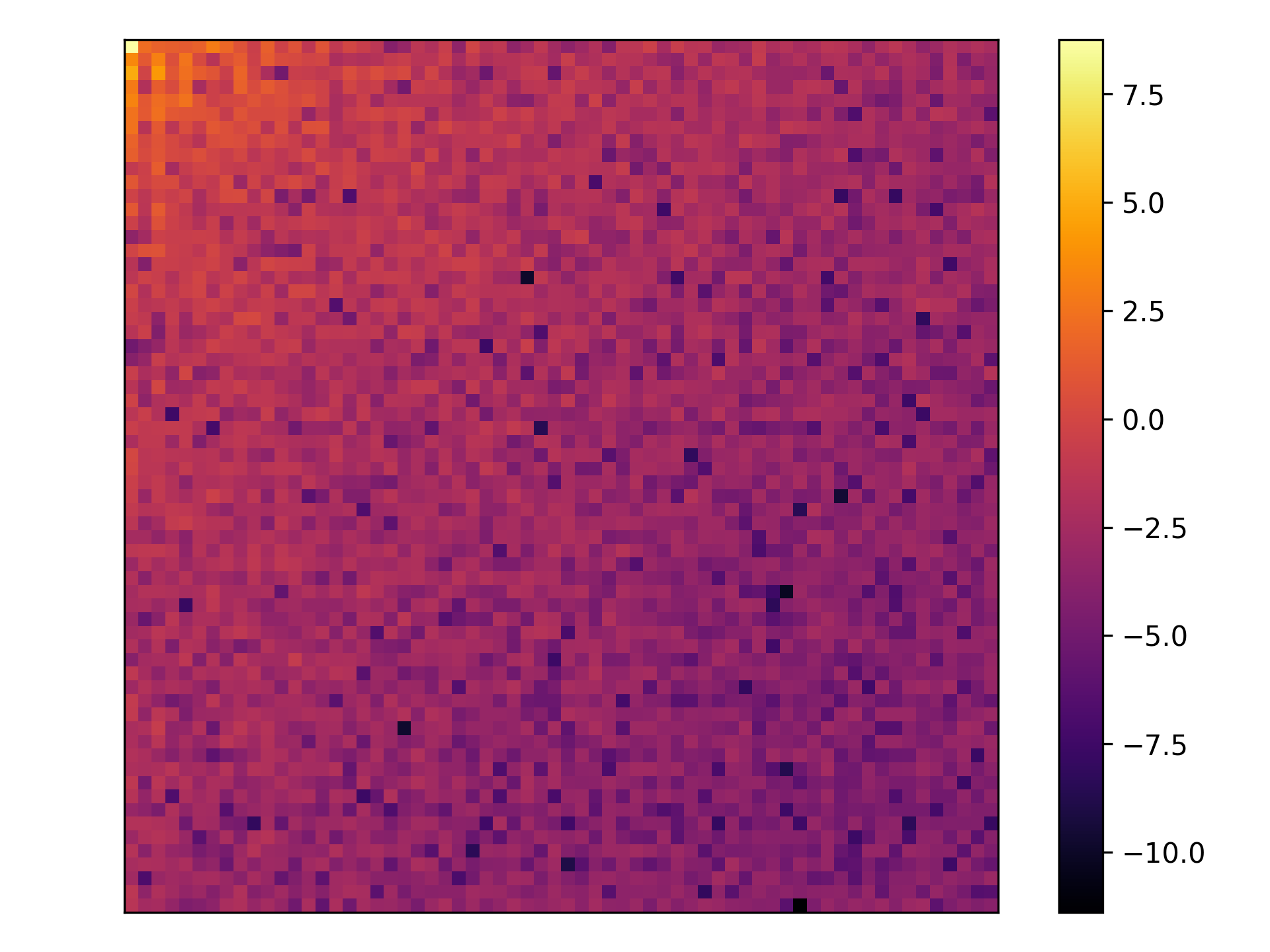}
        \caption*{Stanford dogs (blue)}
    \end{subfigure}

    \begin{subfigure}{.25\textwidth}
        \centering
        \includegraphics[width=\textwidth]{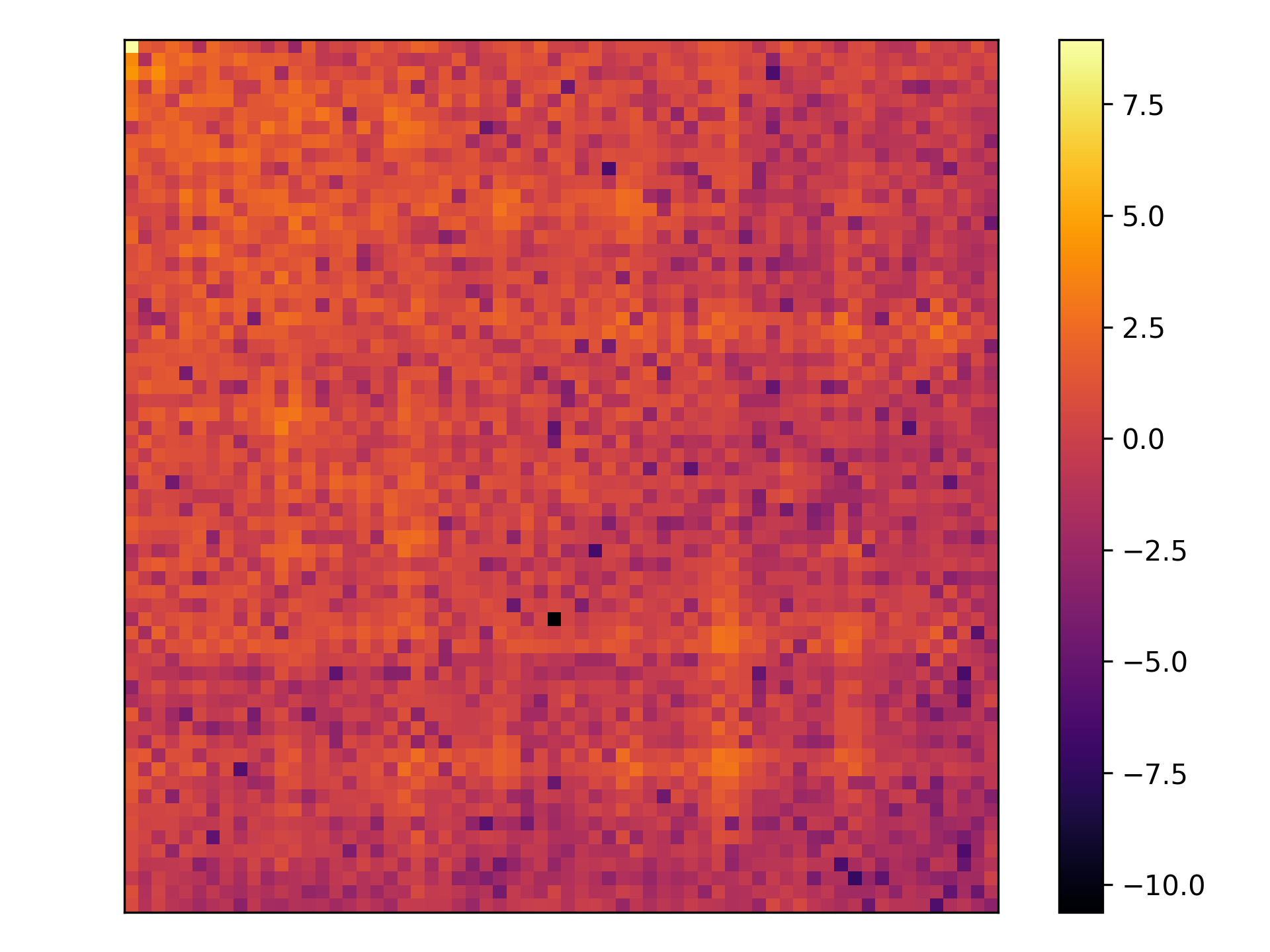}
        \caption*{BigGAN (red)}
    \end{subfigure}
    \begin{subfigure}{.25\textwidth}
        \centering
        \includegraphics[width=\textwidth]{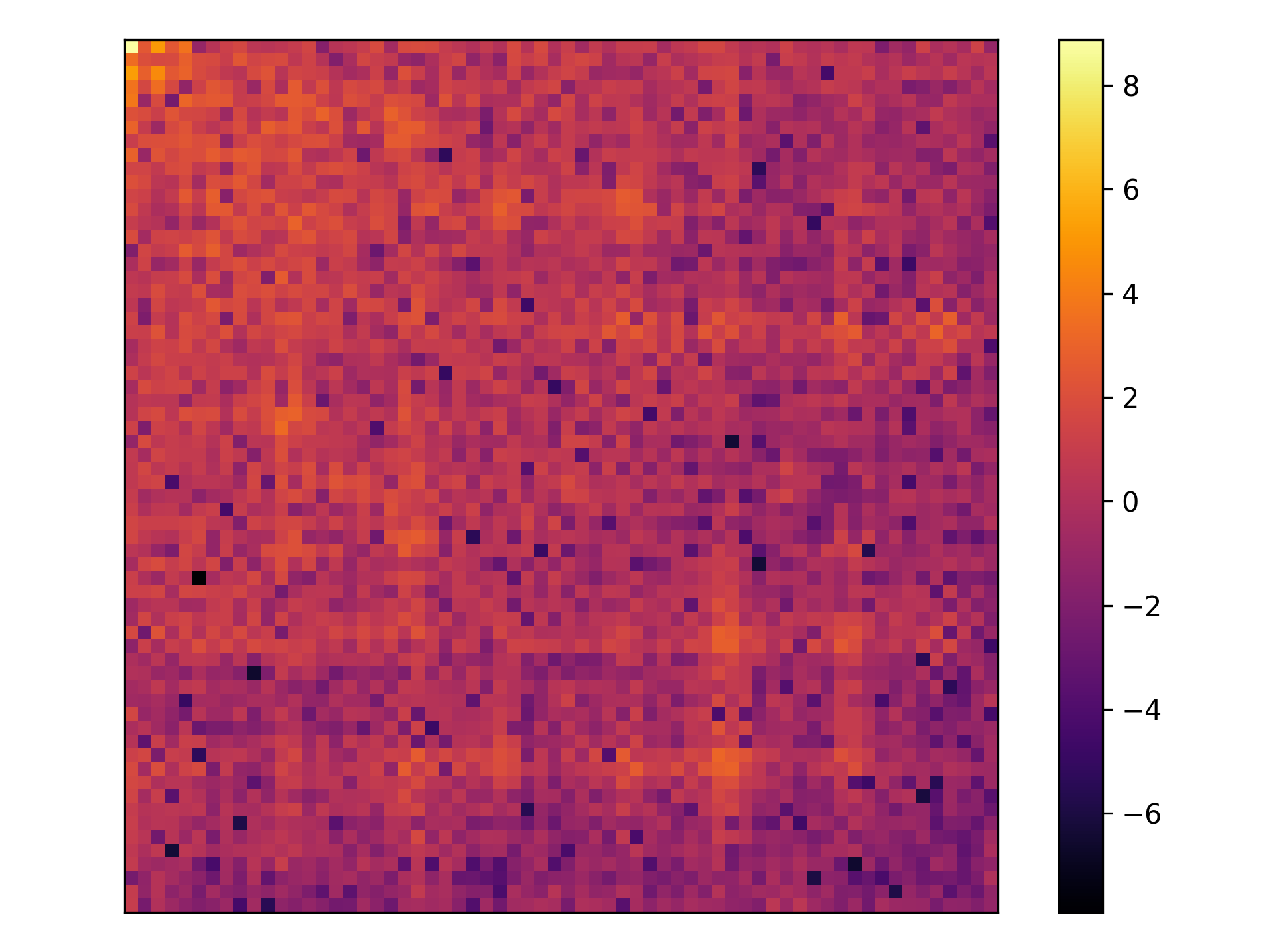}
        \caption*{BigGAN (green)}
    \end{subfigure}
    \begin{subfigure}{.25\textwidth}
        \centering
        \includegraphics[width=\textwidth]{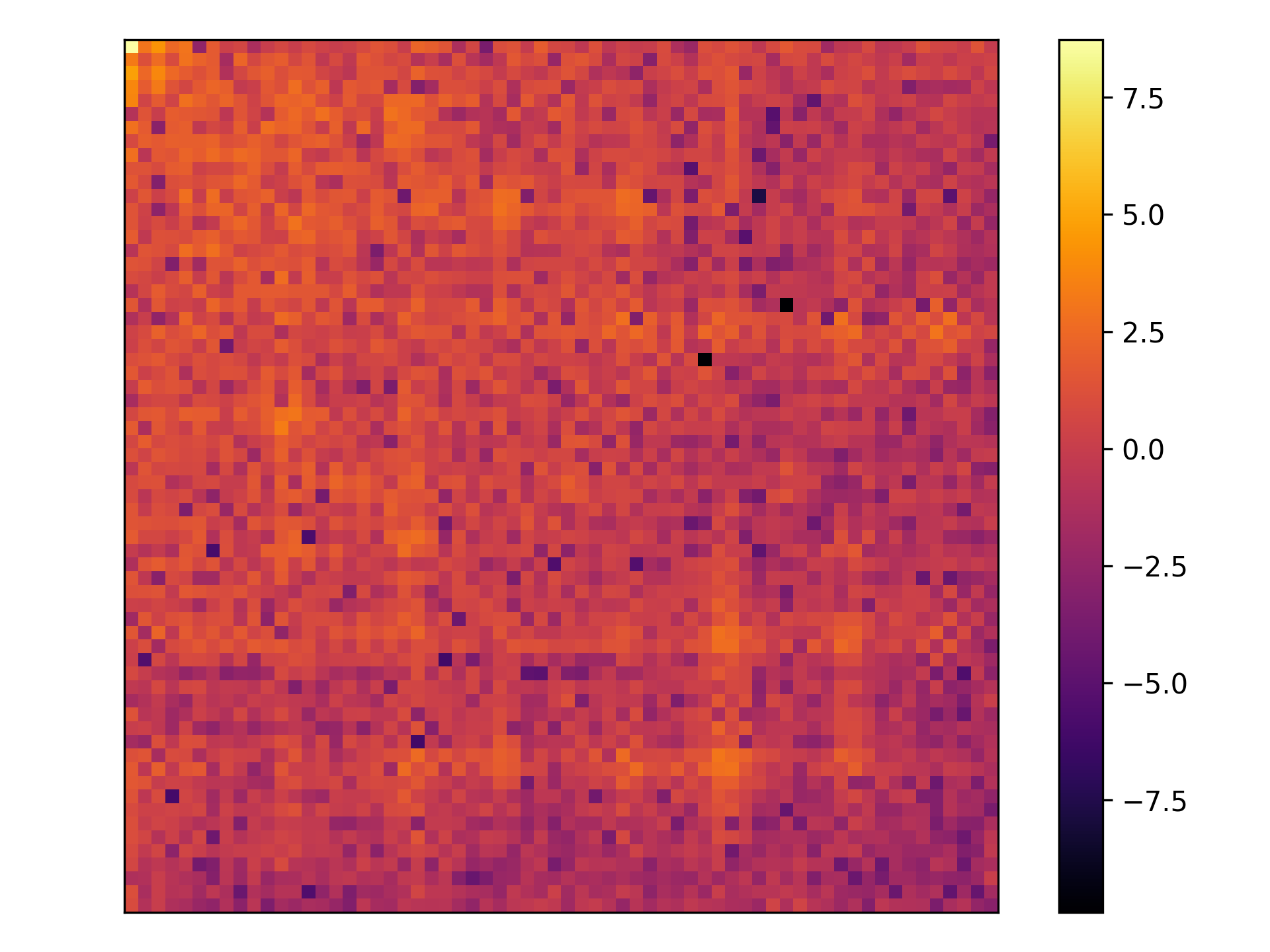}
        \caption*{BigGAN (blue)}
    \end{subfigure}

    \begin{subfigure}{.25\textwidth}
        \centering
        \includegraphics[width=\textwidth]{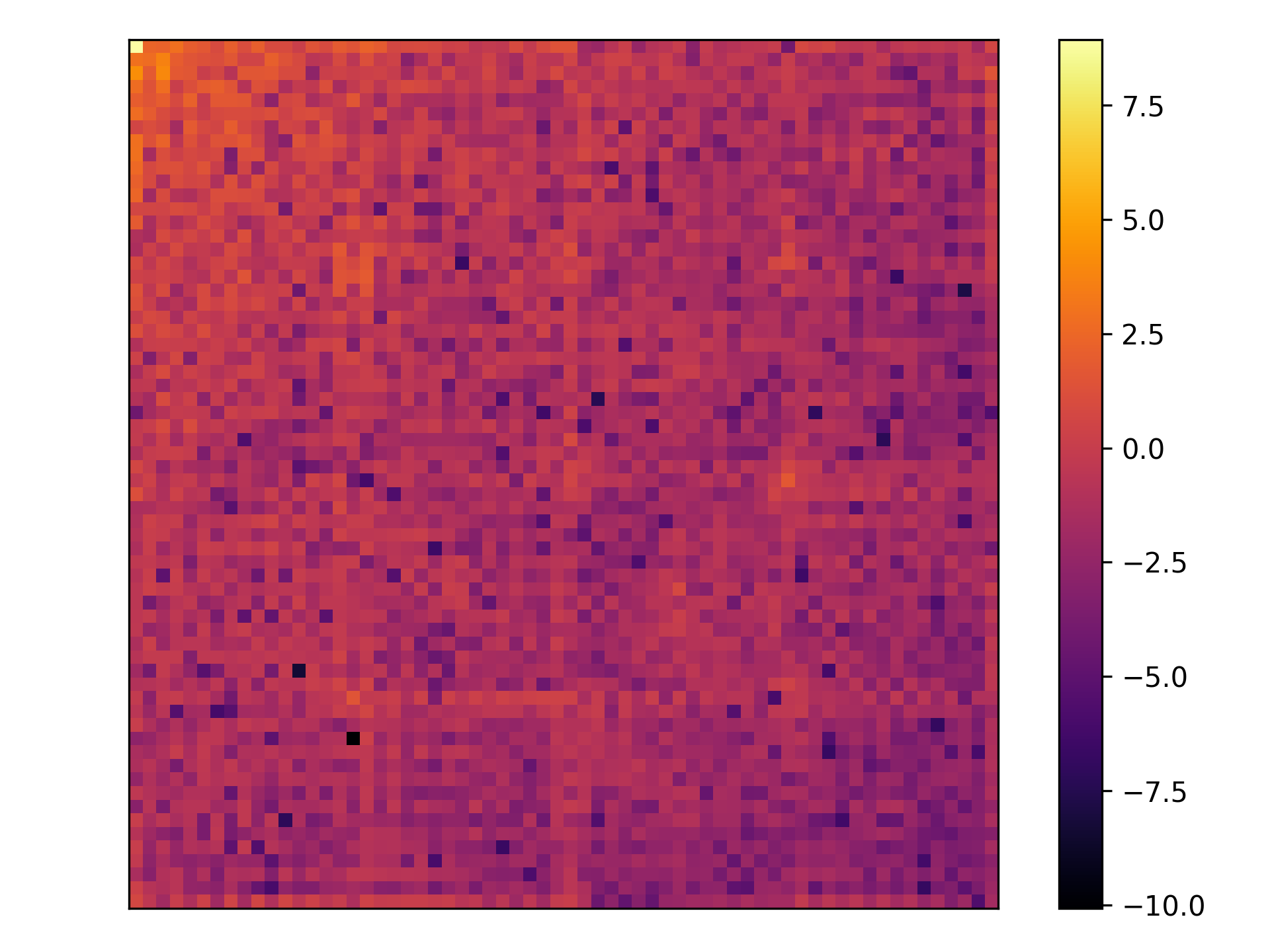}
        \caption*{ProGAN (red)}
    \end{subfigure}
    \begin{subfigure}{.25\textwidth}
        \centering
        \includegraphics[width=\textwidth]{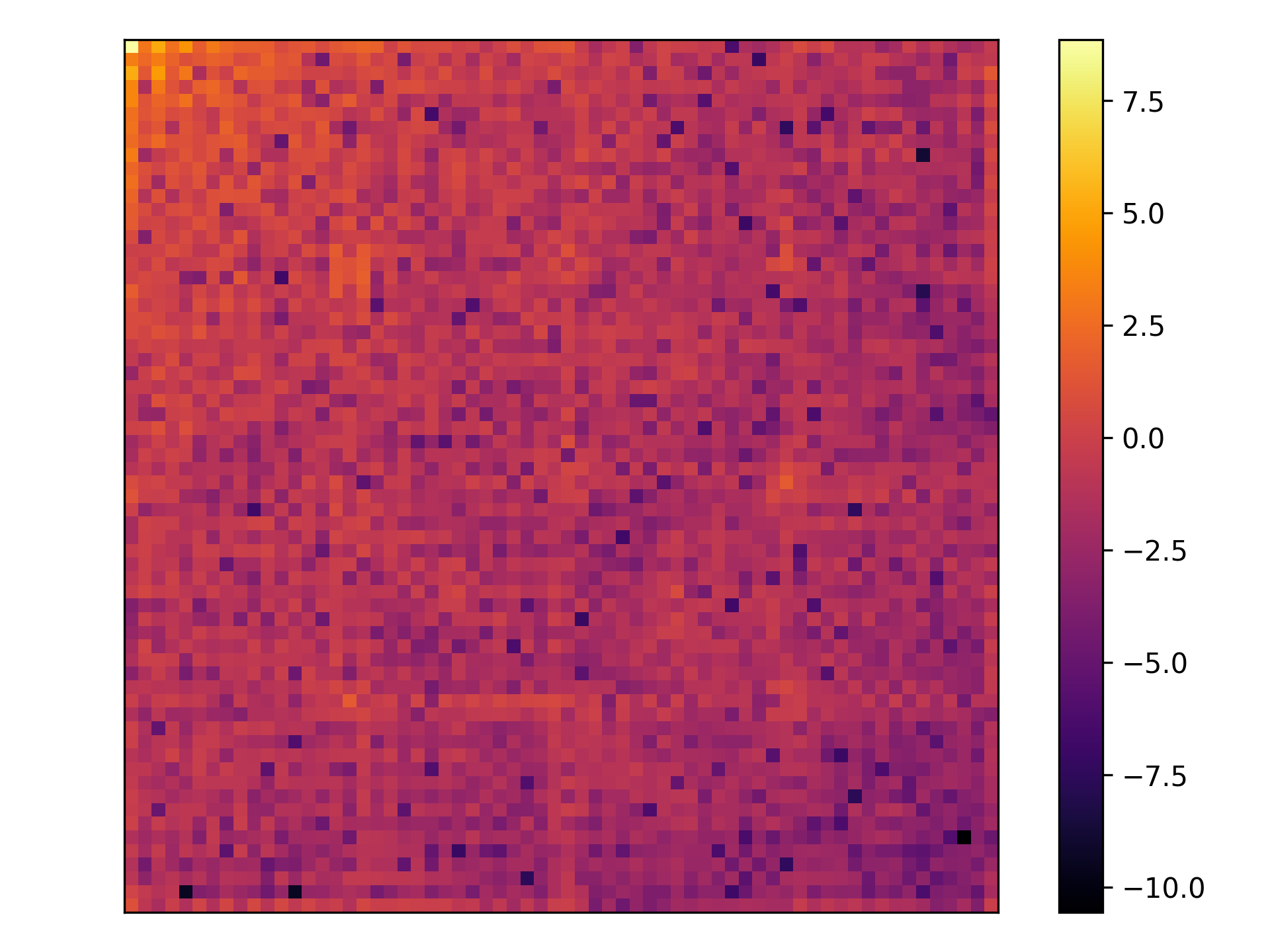}
        \caption*{ProGAN (green)}
    \end{subfigure}
    \begin{subfigure}{.25\textwidth}
        \centering
        \includegraphics[width=\textwidth]{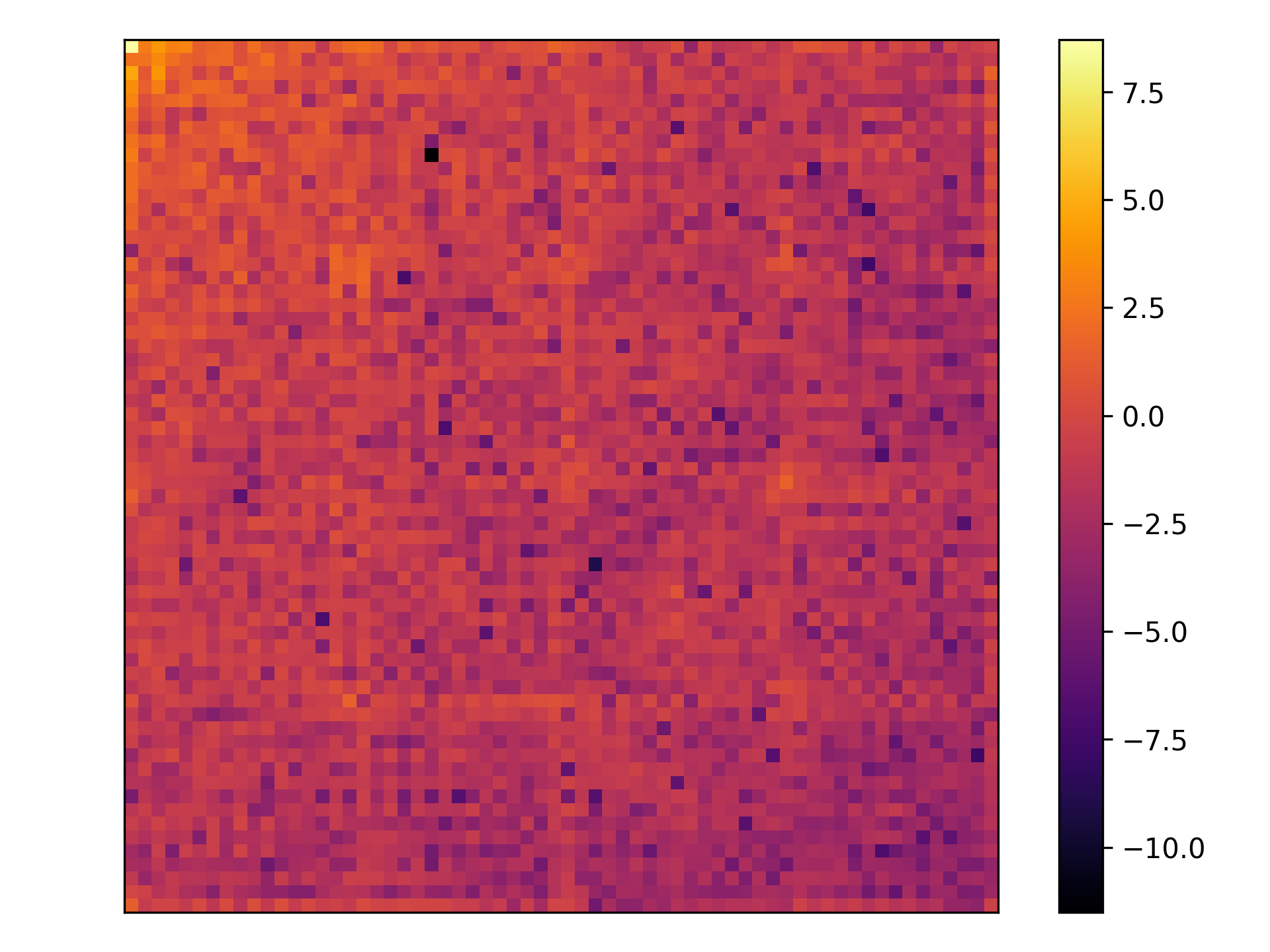}
        \caption*{ProGAN (blue)}
    \end{subfigure}

    \begin{subfigure}{.25\textwidth}
        \centering
        \includegraphics[width=\textwidth]{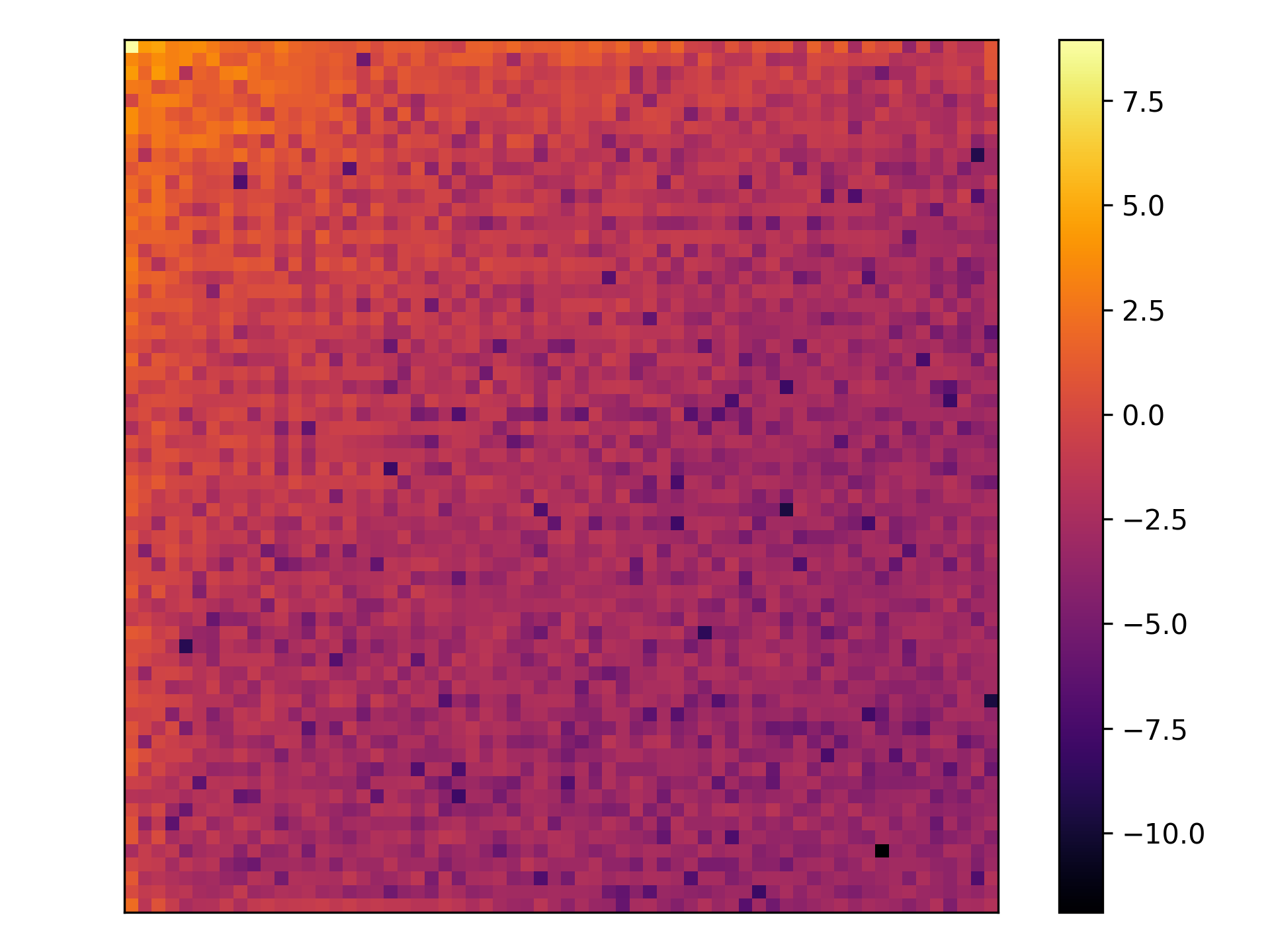}
        \caption*{StyleGAN (red)}
    \end{subfigure}
    \begin{subfigure}{.25\textwidth}
        \centering
        \includegraphics[width=\textwidth]{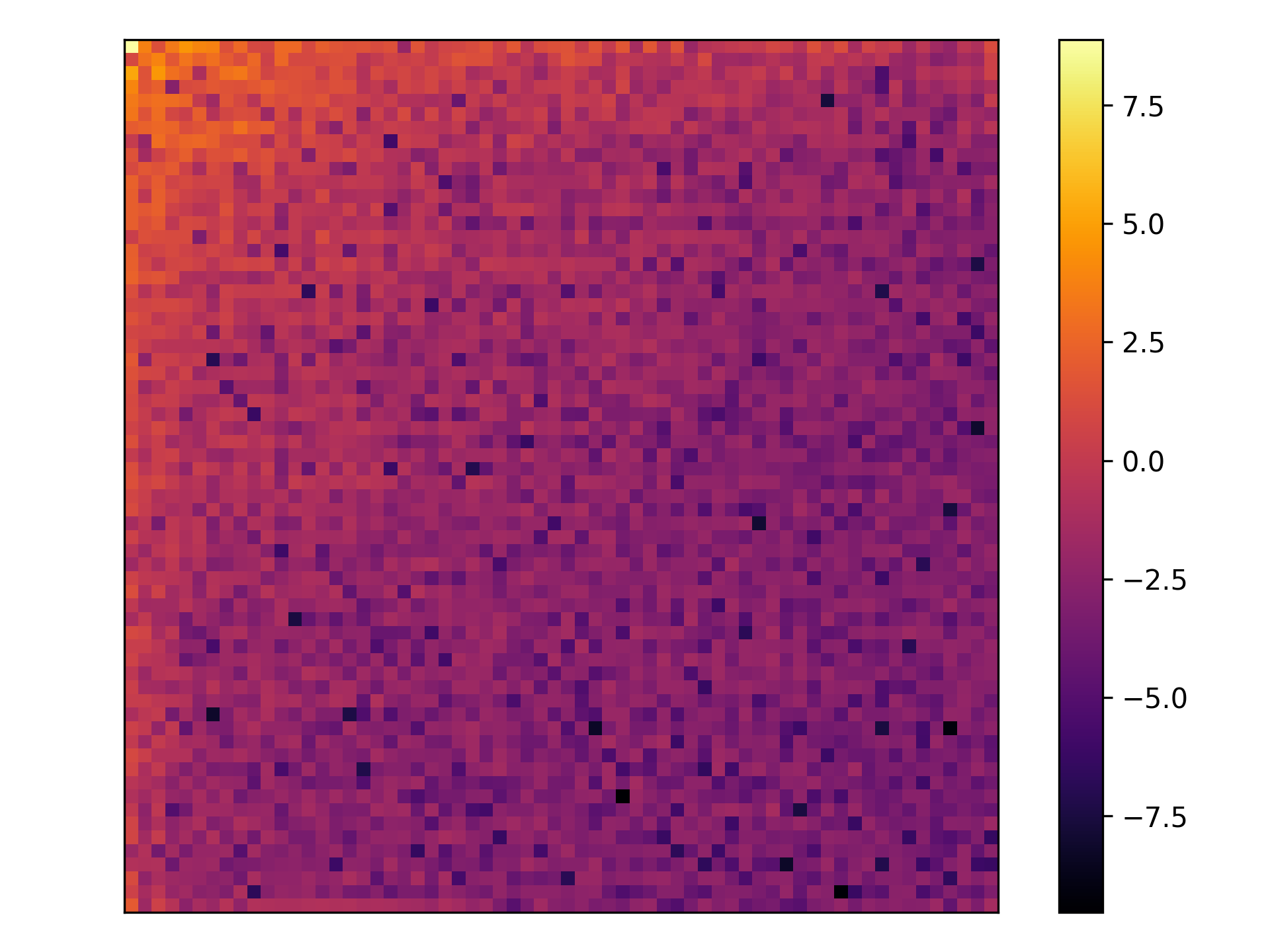}
        \caption*{StyleGAN (green)}
    \end{subfigure}
    \begin{subfigure}{.25\textwidth}
        \centering
        \includegraphics[width=\textwidth]{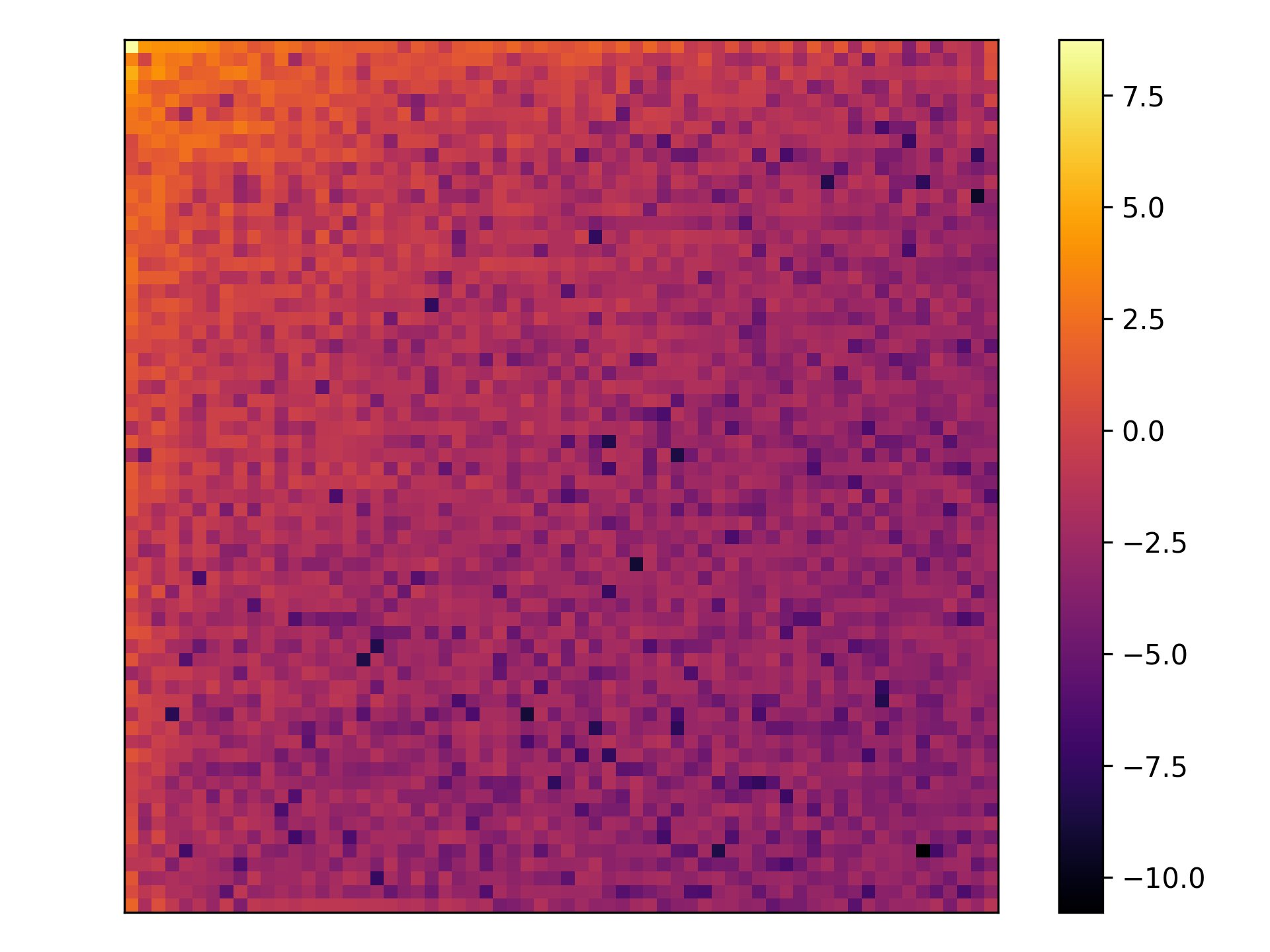}
        \caption*{StyleGAN (blue)}
    \end{subfigure}
    
    \begin{subfigure}{.25\textwidth}
        \centering
        \includegraphics[width=\textwidth]{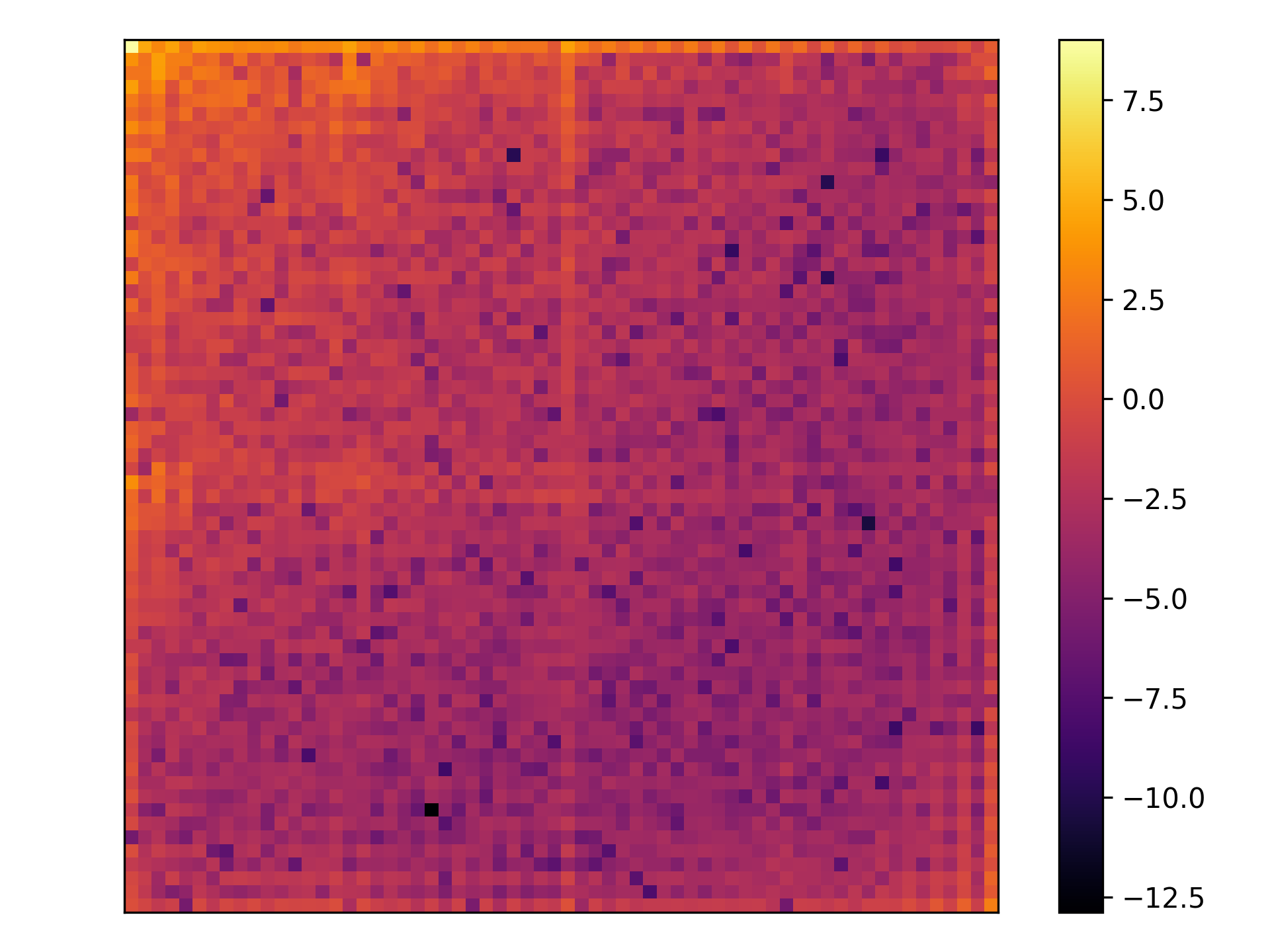}
        \caption*{SN-DCGAN (red)}
    \end{subfigure}
    \begin{subfigure}{.25\textwidth}
        \centering
        \includegraphics[width=\textwidth]{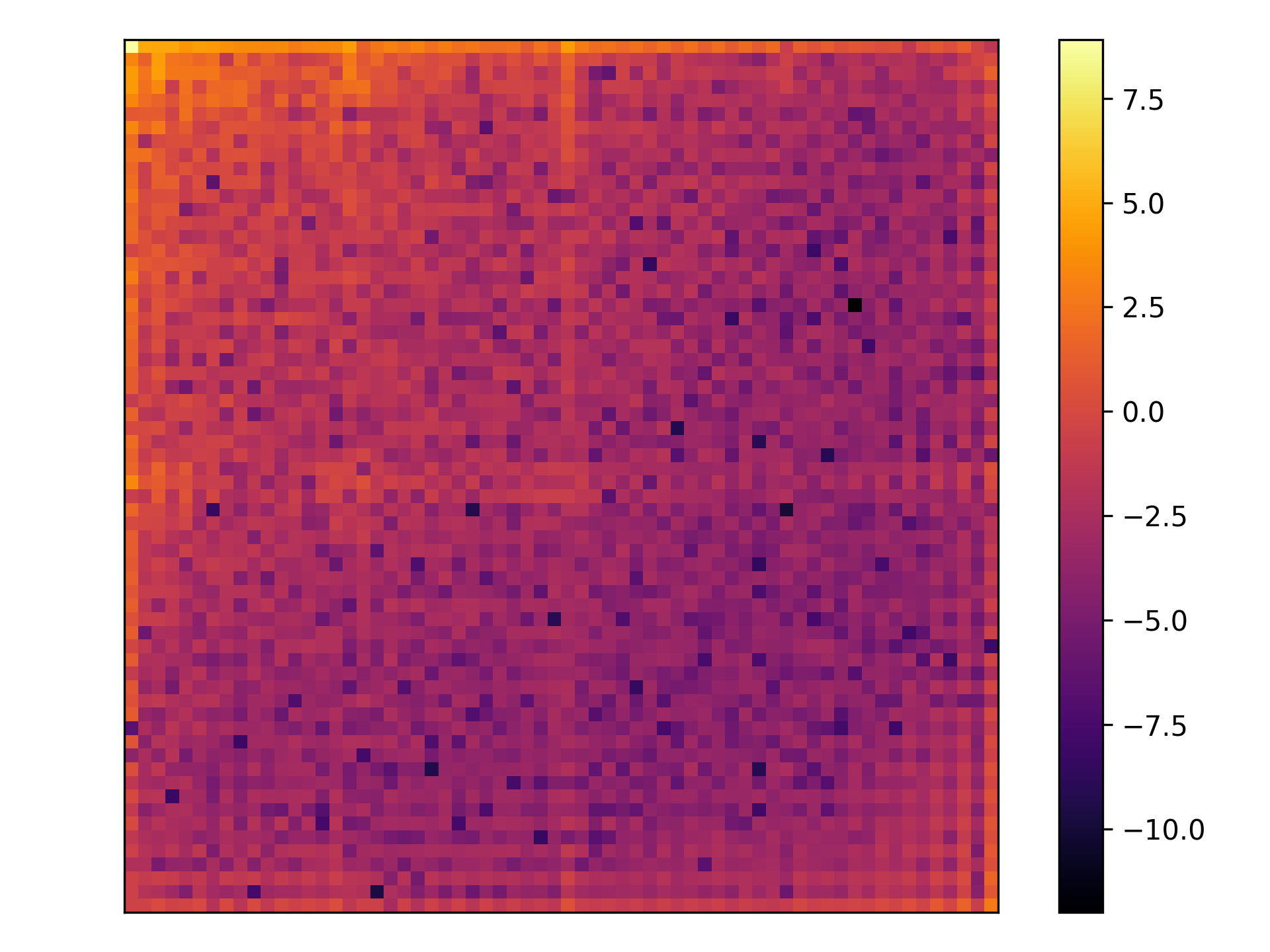}
        \caption*{SN-DCGAN (green)}
    \end{subfigure}
    \begin{subfigure}{.25\textwidth}
        \centering
        \includegraphics[width=\textwidth]{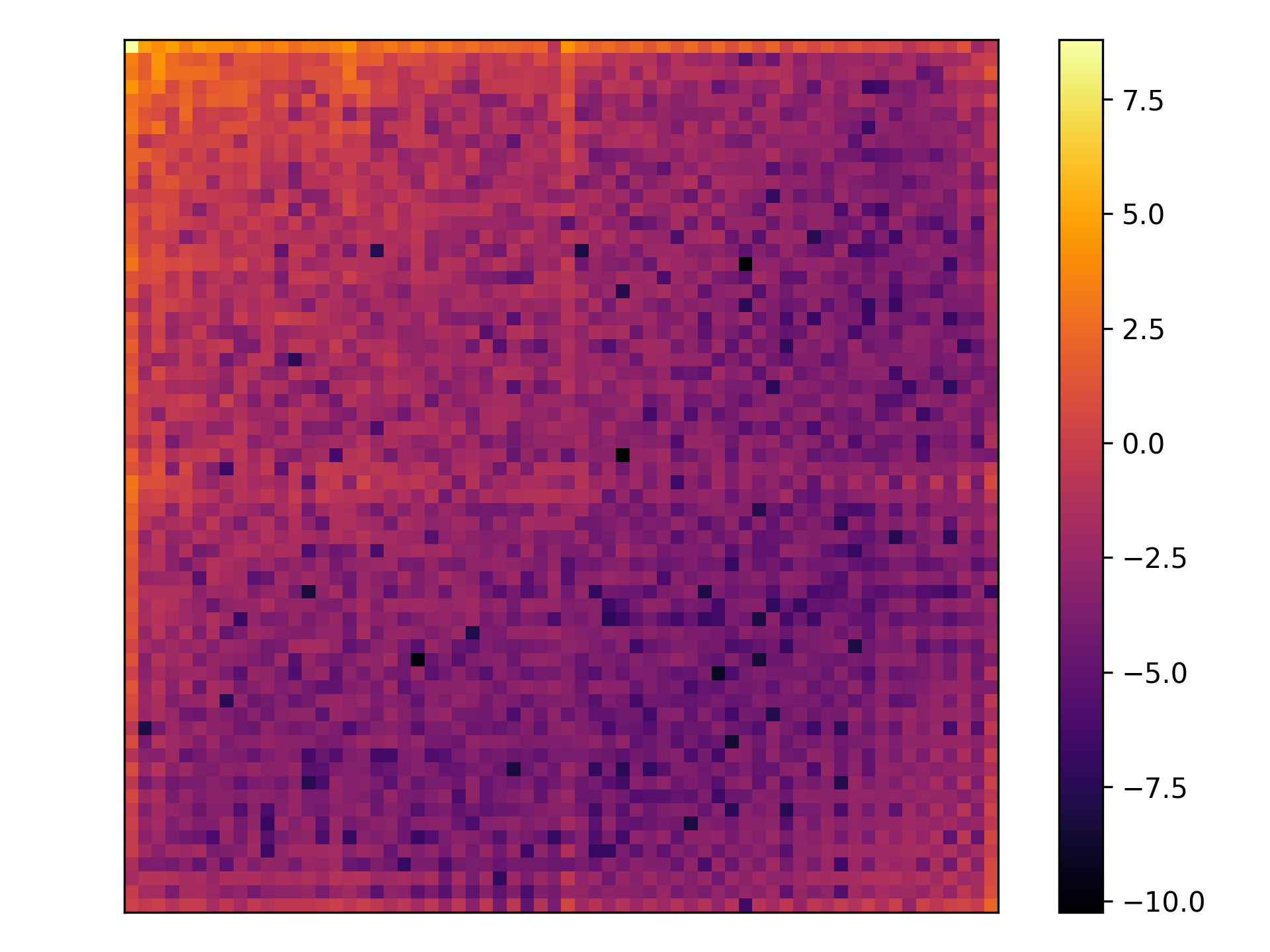}
        \caption*{SN-DCGAN (blue)}
    \end{subfigure}
    
    \caption{\textbf{The frequency spectrum of sample sets generated by different 
    types of GANs trained on the Stanford dog data set (split into color channel)}}
\end{figure}

\clearpage

\section{Upsampling}

The frequency spectrum resulting from different upsampling techniques.
We plot the mean of the DCT spectrum.
We estimate $\E{\dctcom{I}}$ by averaging over 10,000 images sampled from the corresponding network or the training data.
We additionally plot the absolute difference to the mean spectrum of the training images.
Note that, while there is less of a grid, the binomial upsampling still leaves artifacts scattered throughout the spectrum.

\begin{figure}[h!]
    \centering
    \begin{subfigure}{.43\textwidth}
        \centering
        \includegraphics[width=\textwidth]{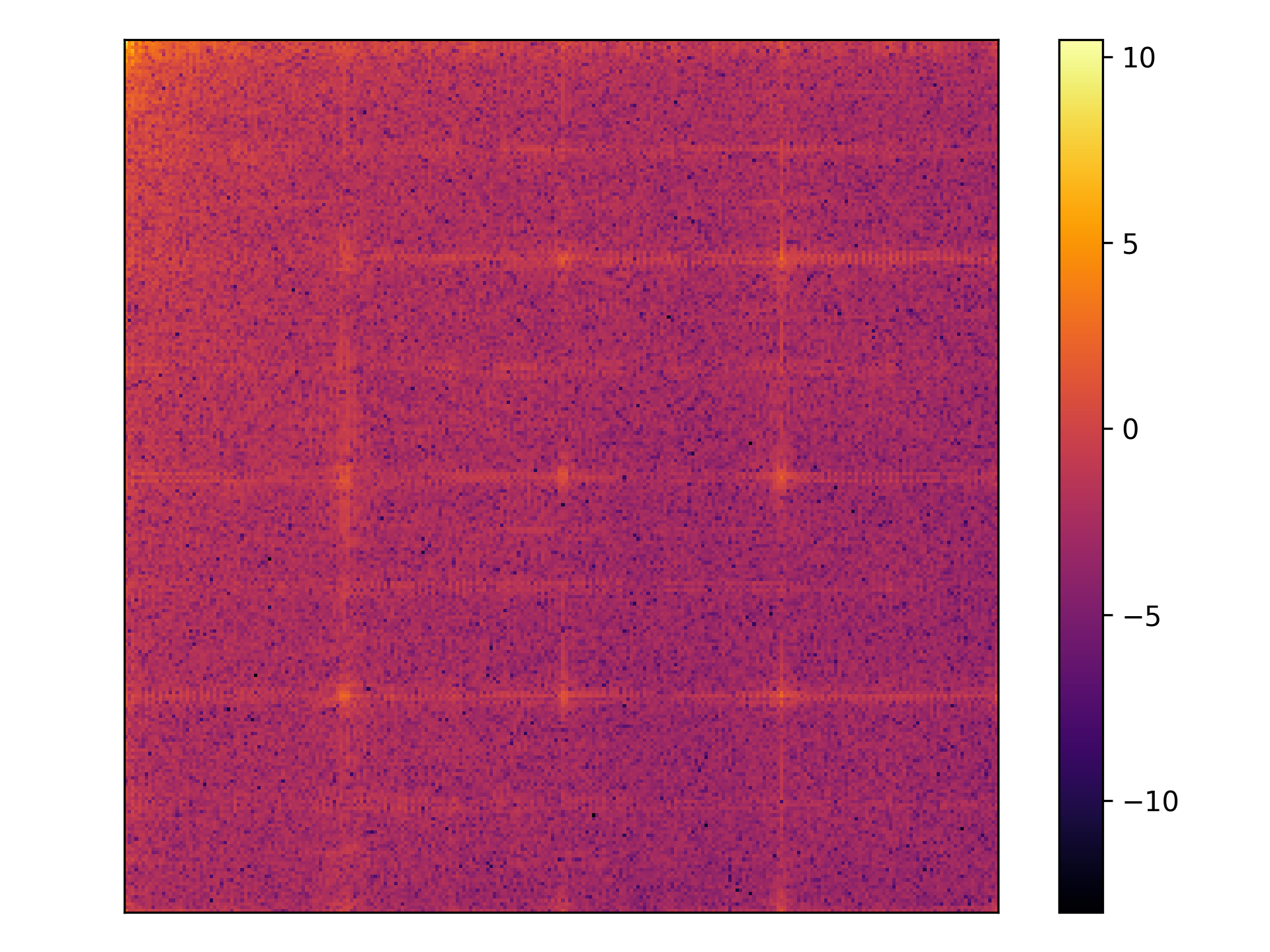}
        \caption*{Nearest Neighbor}
    \end{subfigure}
    \begin{subfigure}{.43\textwidth}
        \centering
        \includegraphics[width=\textwidth]{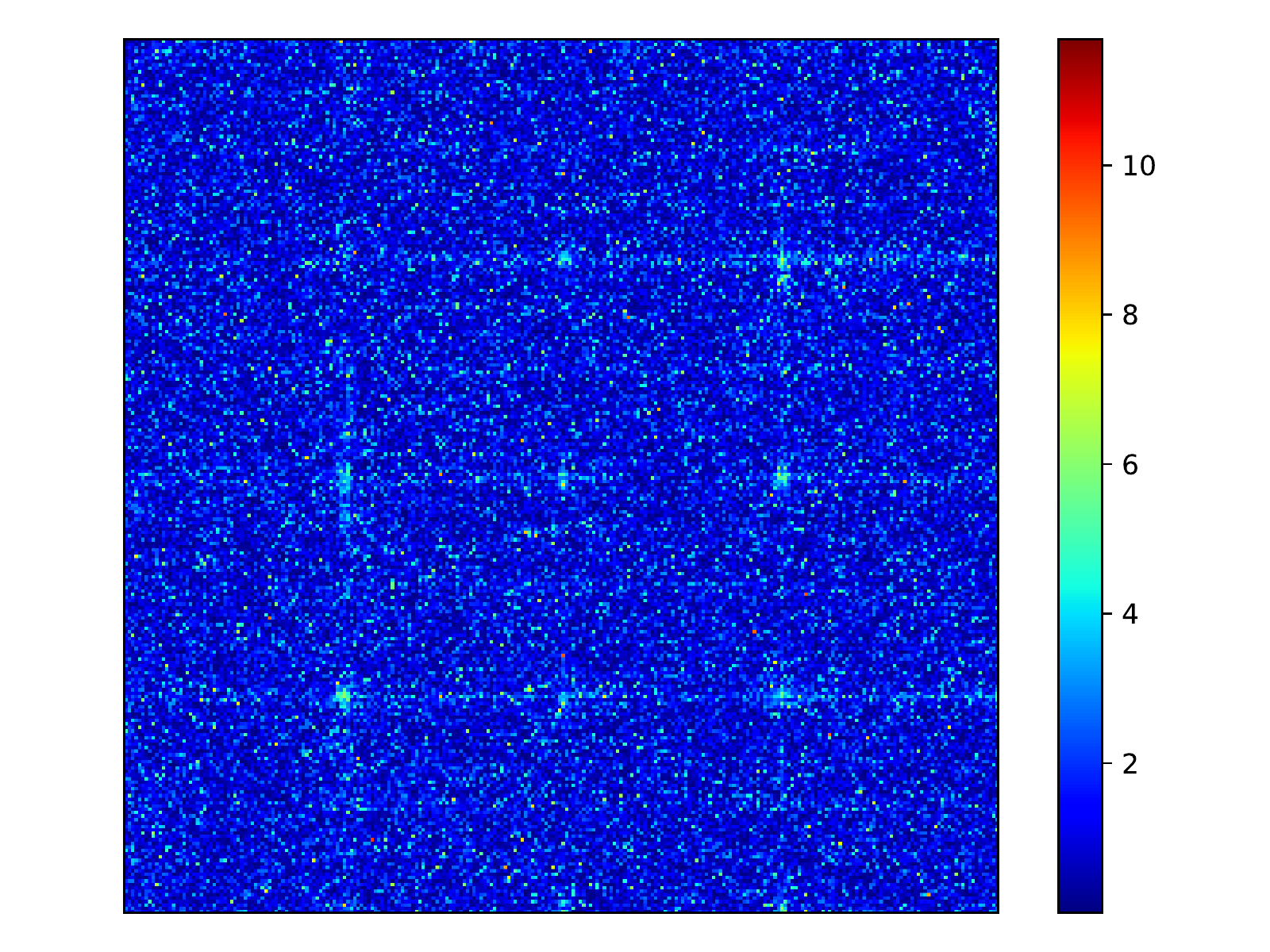}
        \caption*{$\mid \E{\dctcom{\text{LSUN}}} - \E{\dctcom{\text{Nearest Neighbor}}} \mid$}
    \end{subfigure}
    
    \begin{subfigure}{.43\textwidth}
        \centering
        \includegraphics[width=\textwidth]{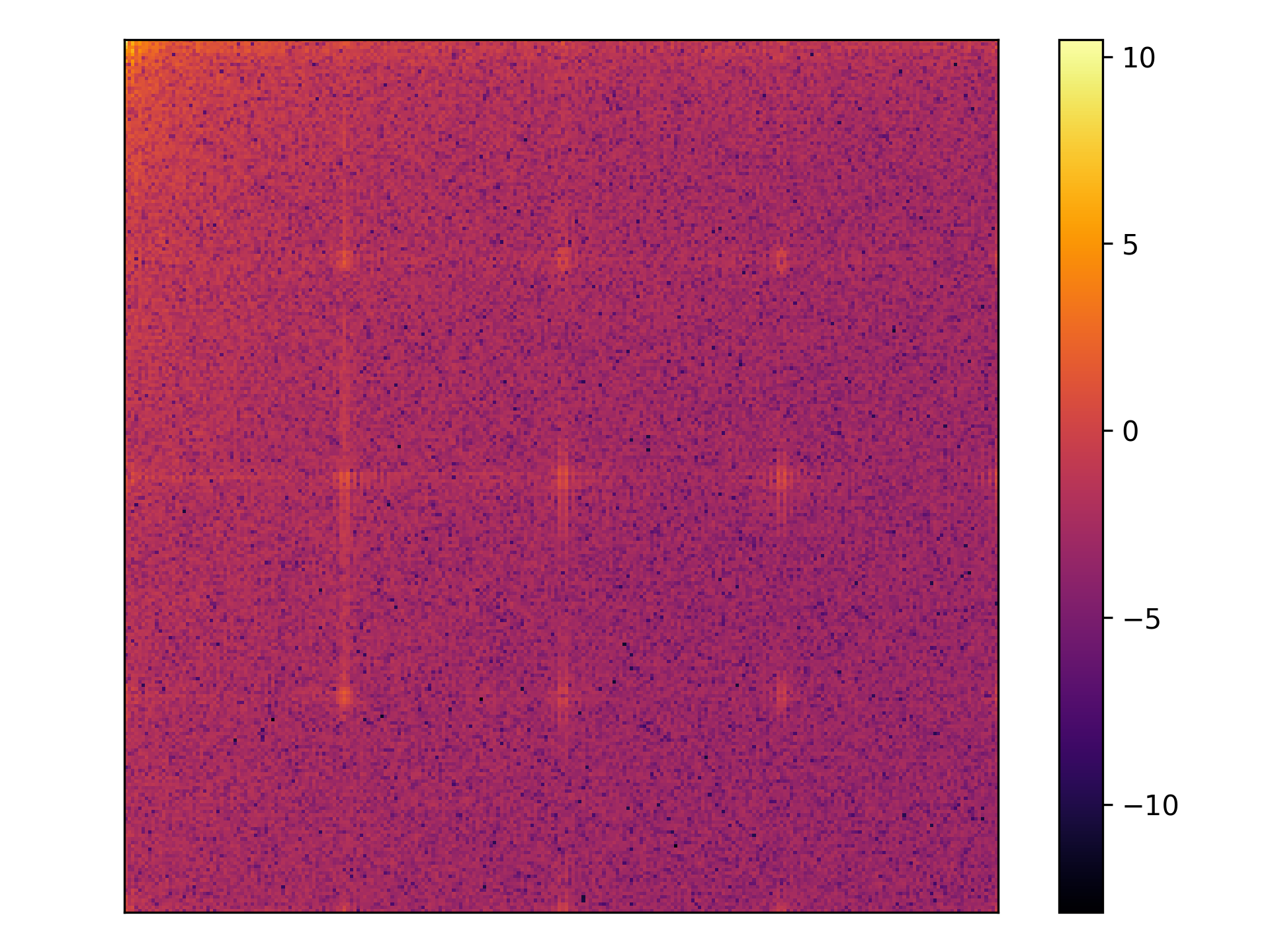}
        \caption*{Bilinear}
    \end{subfigure}
    \begin{subfigure}{.43\textwidth}
        \centering
        \includegraphics[width=\textwidth]{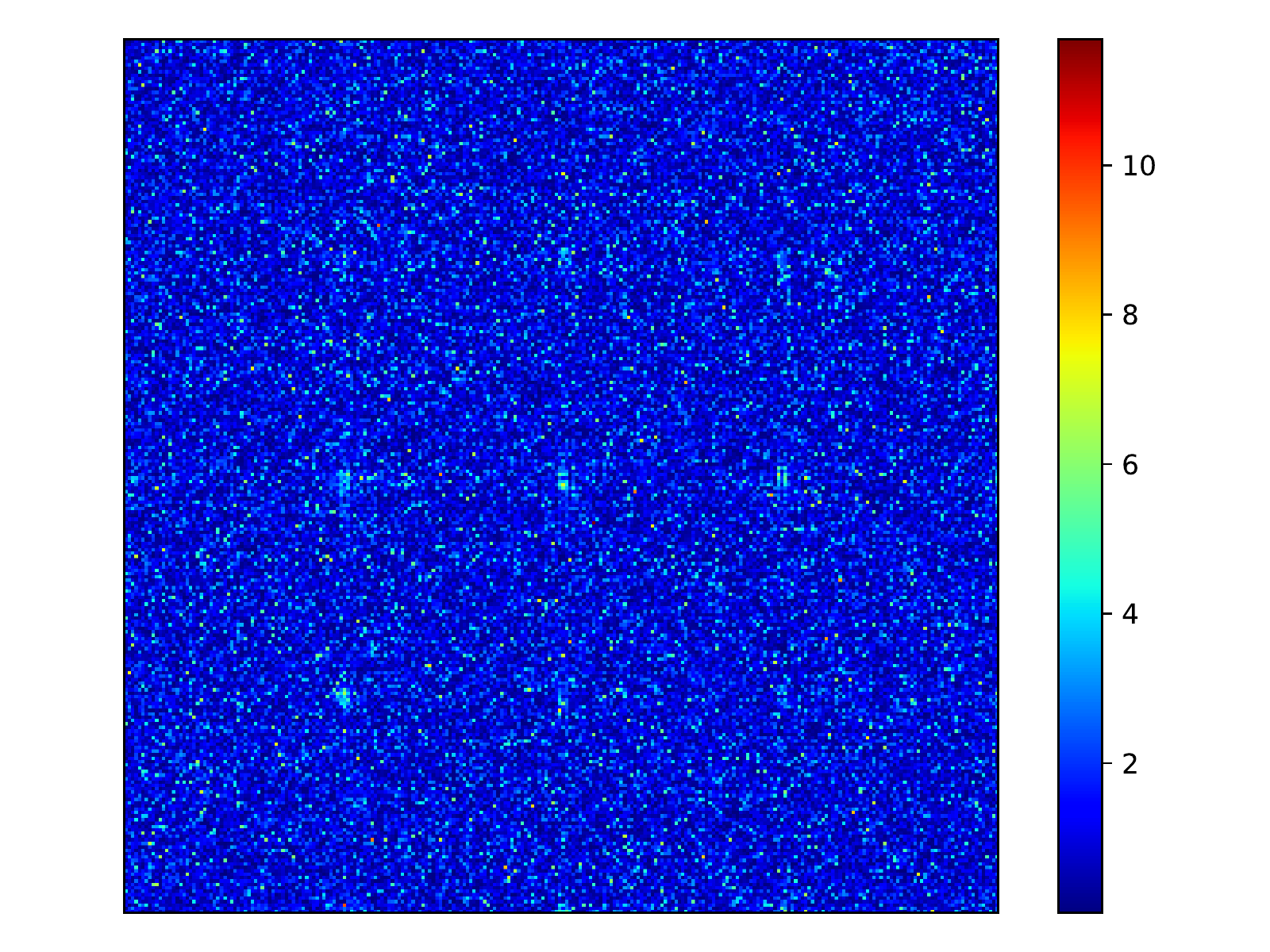}
        \caption*{$\mid \E{\dctcom{\text{LSUN}}} - \E{\dctcom{\text{Bilinear}}} \mid$}
    \end{subfigure}
    
    \begin{subfigure}{.43\textwidth}
        \centering
        \includegraphics[width=\textwidth]{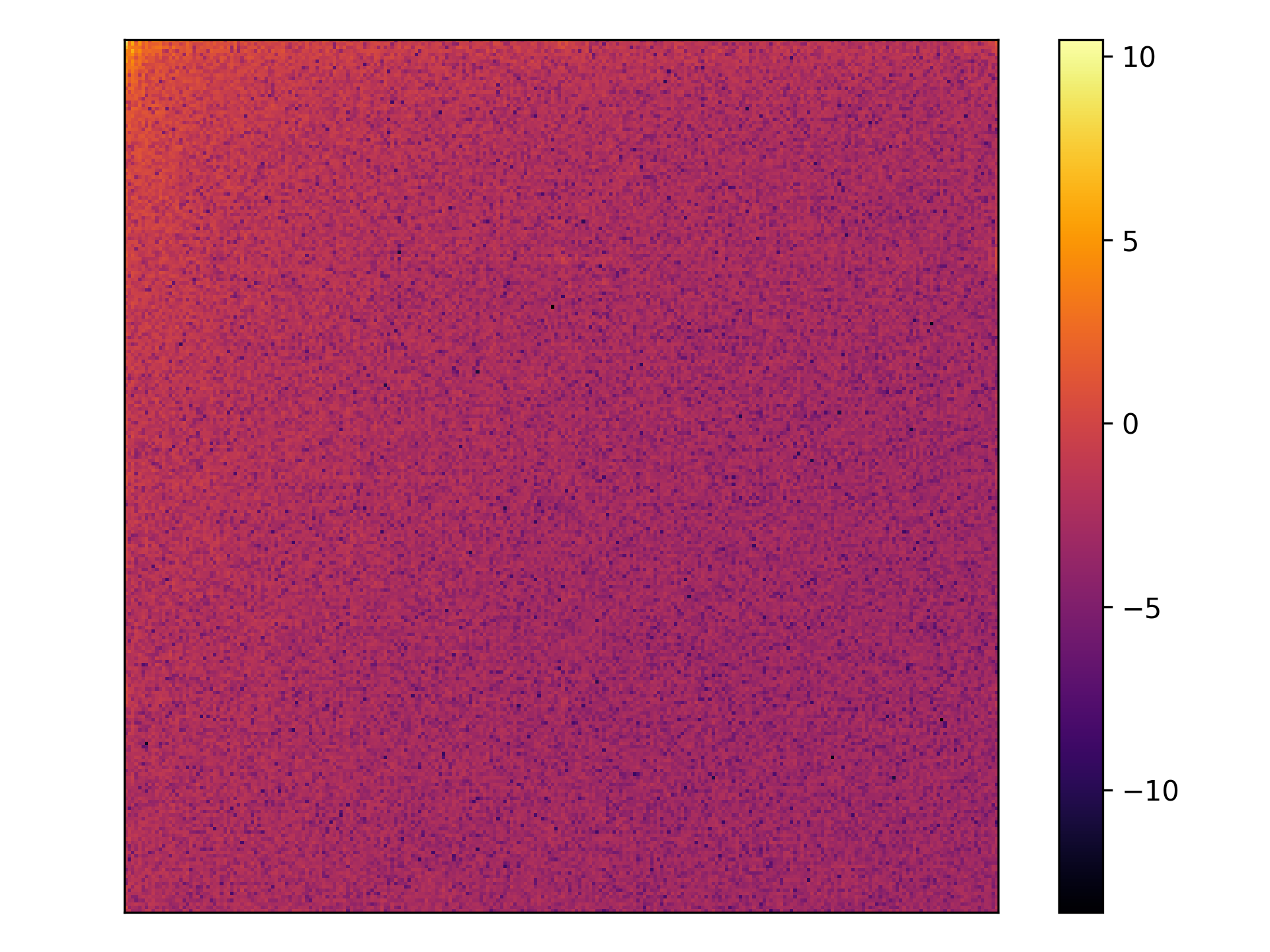}
        \caption*{Binomial}
    \end{subfigure}
    \begin{subfigure}{.43\textwidth}
        \centering
        \includegraphics[width=\textwidth]{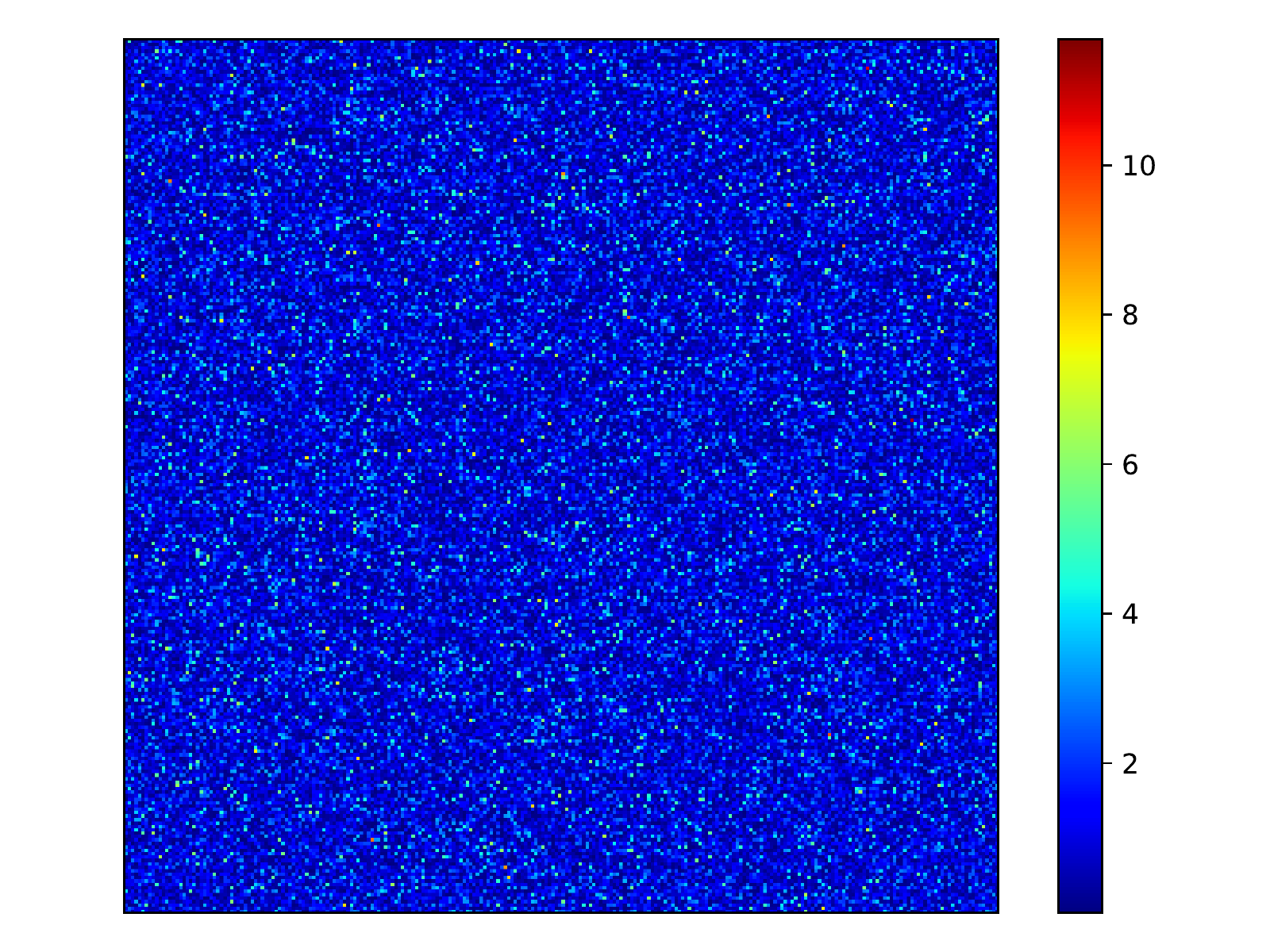}
        \caption*{$\mid \E{\dctcom{\text{LSUN}}} - \E{\dctcom{\text{Binomial}}} \mid$}
    \end{subfigure}
    
    \caption{\textbf{The frequency spectrum resulting from different upsampling techniques, as well as the absolute difference (grayscale)}}
\end{figure}

\clearpage

\section{Network Architecture}

For training our CNN we use the Adam optimizer, with an initial learning rate of $0.001$, $\beta_1 = 0.9$, $\beta_2 = 0.999$ and $\epsilon = 1^{-7}$, which are the standard parameters for a TensorFlow implementation.
We did some experiments with different settings, but none seem to influence the training substantially, so we kept the standard configuration.
We train with a batch size of 1024. Again, we experimented with lower batch sizes, which did not influence the training.
Thus, we simply picked the largest batch size our GPUs allowed for.

\begin{table}[h!]
    \centering
    \begin{tabular}{c}
        \toprule \midrule
        Input (128x128x3) \\\midrule
        Conv 3x3 (128x128x3) \\\midrule
        Conv 3x3 (128x128x8) \\\midrule
        Average-Pool 2x2 (64x64x8) \\\midrule
        Conv 3x3 (64x64x16) \\\midrule
        Average-Pool 2x2 (32x32x16) \\\midrule
        Conv 3x3 (32x32x32) \\\midrule
        Dense (5) \\\midrule
        \bottomrule
    \end{tabular}
    \caption{\textbf{The network architecture for our simply CNN.} We report the 
    size of each layer in (brackets).}
    \label{tab:my_label}
\end{table} 
\end{document}